%% file: bbf.tex
\documentclass{siamart1116}
\usepackage{graphicx,amsmath,amsfonts,amssymb,epstopdf,bm,bigints}
\usepackage{subcaption,aliascnt,todonotes}
\usepackage{caption}
\usepackage{booktabs,array,ragged2e}
\usepackage{color,enumitem,multirow}
\usepackage[nameinlink,capitalize]{cleveref}
\usepackage{adjustbox}
\usepackage{pdflscape}

\usepackage[english]{babel}
\usepackage[utf8x]{inputenc}
\usepackage[T1]{fontenc}
\usepackage{hyperref}
\usepackage{multirow}
\usepackage{hhline}
\SetKwProg{Fn}{Function}{}{}
\usepackage{algpseudocode}
\usepackage{lipsum}
\usepackage{amsopn}
\usepackage{float}
\newcolumntype{C}[1]{>{\centering\hspace{-10pt}}m{#1}}
\usepackage{diagbox} 

\usepackage[final]{revision}

\ifpdf
  \DeclareGraphicsExtensions{.eps,.pdf,.png,.jpg}
\else
  \DeclareGraphicsExtensions{.eps}
\fi

\numberwithin{theorem}{section}

\addto\extrasenglish{%

}

\newcommand{\TheTitle}{Block Basis Factorization} 
\newcommand{\TheAuthors}{R. Wang, Y. Li, M. Mahoney and E. Darve}

\headers{\TheTitle}{\TheAuthors}

\title{Block Basis Factorization for Scalable Kernel Evaluation}

\author{
    Ruoxi Wang \thanks{
        Institute for Computational and Mathematical Engineering,
        Stanford University,
        \email{ruoxi@stanford.edu}
    }
    \and
    Yingzhou Li \thanks{
        Department of Mathematics,
        Duke University,
        \email{yingzhou.li@duke.edu}
    }
    \and
    Michael W. Mahoney \thanks{
        International Computer Science Institute
        and Department of Statistics,
        University of California, Berkeley,
        \email{mmahoney@stat.berkeley.edu}
    }
    \and
    Eric Darve \thanks{
        Department of Mechanical Engineering,
        Stanford University,
        \email{darve@stanford.edu}
    }
}

\newcommand{\vecx}{{\bf x}}
\newcommand{\vecy}{{\bf y}}
\newcommand{\vecalpha}{{\bm \alpha}}
\newcommand{\norm}[1]{\lVert #1 \rVert}

\newcommand{\calK}{\mathcal{K}}
\newcommand{\calO}{\mathcal{O}}

\begin{document}
\date{}
\maketitle

\begin{abstract} 

Kernel methods are widespread in machine learning; however, they
are limited by the quadratic complexity of the construction,
application, and storage of kernel matrices. Low-rank matrix
approximation algorithms are widely used to address this problem and
reduce the arithmetic and storage cost.  However, we observed that
for some datasets with wide intra-class variability, the optimal
kernel parameter for smaller classes yields a matrix that is less
well approximated by low-rank methods.  In this paper, we propose
an efficient structured low-rank approximation method---the Block
Basis Factorization (BBF)---and its fast construction algorithm to
approximate radial basis function (RBF) kernel matrices. Our approach
has linear memory cost and floating point operations \added{R1}{5}{for many machine learning kernels}. BBF works for
a wide range of kernel bandwidth parameters and extends the domain
of applicability of low-rank approximation methods significantly.
Our empirical results demonstrate the stability and superiority over
the state-of-art kernel approximation algorithms.

\end{abstract} 

\section{Introduction}

Kernel methods are mathematically well-founded nonparametric methods
for learning. The essential part of kernel methods is a kernel function
$\mathcal{K}: \mathcal{X} \times \mathcal{X} \mapsto \mathbb{R}$. It
is associated with a feature map $\Psi$ from the original input
space $\mathcal{X} \in \mathbb{R}^d$ to a higher-dimensional Hilbert
space $\mathcal{H}$, such that $$\mathcal{K}(\vecx, \vecy) = \langle
\Psi(\vecx), \Psi(\vecy) \rangle_\mathcal{H}.$$ Presumably, the
underlying function for data in the feature space is linear. Therefore,
the kernel function enables us to build expressive nonlinear models
based on the machinery of linear models. In this paper, we consider
the radial basis function (RBF) kernel that is widely used in machine
learning.

The kernel matrix is an essential
part \replaced{R1}{3}{in most kernel methods}{of kernel methods in the training phase} and is defined in what follows. Given $n$
data points $\{\vecx_i\}_{i=1}^n$, the $(i,j)$-th entry in a kernel
matrix is $K_{ij} = \mathcal{K}(\vecx_i, \vecx_j)$. For example, the
solution to a kernel ridge regression is the same as the solution to
the linear system $$(K+\delta I) \vecalpha = \vecy.$$

Regrettably, any operations involving kernel matrices can be
computationally expensive.  Their construction, application and storage
complexities are quadratic in the number of data points $n$. Moreover,
for solving linear systems involving these matrices, the complexity is
even higher. It is $O(n^3)$ for direct solvers \cite{golub2012matrix}
and $O(n^2T)$ for iterative solvers \cite{hestenes1952methods,
paige1975solution}, where $T$ is the iteration number. This is
prohibitive in large-scale applications. One popular solution to
address this problem and reduce the arithmetic and storage cost
is using matrix approximation. If we are able to approximate the
matrix such that the number of entries that need to be stored is
reduced, then the timing for iterative solvers will be accelerated
(assuming memory is a close approximation of the running time for a
matrix-vector multiplication).

In machine learning, low-rank matrix approximations are
widely used \cite{golub2012matrix, mahoney2011randomized,
halko2011finding, liberty2007randomized, sarlos2006improved,
drineas2005nystrom,gittens2016revisiting,drineas2012fast,
zhang2010clustered}. When the kernel matrix has a large spectrum
gap, a high approximation accuracy can be guaranteed by theoretical
results. Even when the matrix does not have a large spectrum gap or
fast spectrum decay, these low-rank algorithms are still popular
practical choices to reduce the computational cost; however, the
approximation would be less accurate.

The motivation of our algorithm is that in many machine learning
applications, the RBF kernel matrices cannot be well-approximated by
low-rank matrices \cite{Stein2014}; nonetheless, they are not arbitrary high-rank
matrices and are often of certain structure. In the rest of
the introduction, we first discuss the importance of higher rank
matrices and then introduce the main idea of our algorithm that takes
advantage of those structures. The RBF kernel $f(\|\vecx-\vecy\|/h)$
has a \emph{bandwidth parameter} $h$ that controls the size of the
neighborhood, \emph{i.e.}, how many nearby points to be considered
for interactions. The numerical rank of a kernel matrix depends
strongly on this parameter. As $h$ decreases from large to small,
the corresponding kernel matrix can be approximated by a low-rank
matrix whose rank increases from $O(1)$ to $O(n)$. In the large-$h$
regime, traditional low-rank methods are efficient; however, in the
small-$h$ regime, these methods fall back to quadratic complexity.
The bandwidth parameter is often chosen to maximize the overall
performance of regression/classification tasks, and its value is
closely related to the smoothness of the underlying function. For
kernel regressions and kernelized classifiers, the hypothesis function
classes are $\sum_i \alpha_i \mathcal{K}_h(\vecx, \vecx_i)$ and $\sum_i
\alpha_i y_i\mathcal{K}_h(\vecx, \vecx_i)$, respectively. Both can
be viewed as interpolations on the training data points. Clearly,
the optimal value of $h$ should align with the smoothness of the
underlying function. Although many real-world applications have
found large $h$ to lead to good overall performances, in a lot of
cases a large $h$ will hurt the performance. For example, in kernel
regression, when the underlying function is non-smooth such as those
with sharp local changes, using a large bandwidth will smooth out the
local structures; in kernelized classifiers, when the true decision
surfaces that separate two classes are highly nonlinear, choosing
a large bandwidth imposes smooth decision surfaces on the model and
ignores local information near the decision surfaces.  In practice,
the previous situations where relatively small bandwidths are needed
are very common. One example is that for classification dataasets
with imbalanced classes, often the \replaced{R1}{4}{optiomal}{optimal} $h$ for smaller classes
is relatively small. Hence, if we are particularly interested in
the properties of smaller classes, a small $h$ is appropriate. As a
consequence, matrices of higher ranks occur frequently in practice.

Therefore, for certain machine learning problems, low-rank
approximations of dense kernel matrices are inefficient. This
motivates the development of approximation algorithms that extend the
applicability of low-rank algorithms to matrices of higher ranks,
\emph{i.e.}, that work efficiently for a wider range of kernel
bandwidth parameters.

In the field of scientific computing (which also considers
kernel matrices, but typically for very different ends),
hierarchical algorithms \cite{greengard1987fast, greengard1997new,
darve2000fast_error, darve2000fast_numerical, fong2009black,
ying2004kernel} efficiently approximate the forward application
of full-rank PDE kernel matrices in low dimensions. These
algorithms partition the data space recursively using a tree
structure and separate the interactions into near- and far-field
interactions, where the near-field interactions are calculated
hierarchically and the far-field interactions are approximated
using low-rank factorizations. Later, hierarchical matrices
($\mathcal{H}$-matrix, $\mathcal{H}^2$-matrix, HSS matrix,
HODLR matrix) \cite{hackbusch1999sparse, hackbusch2000sparse,
hackbusch2002data, chandrasekaran2006fast, xia2010fast,
aminfar2016fast} were proposed as algebraic variants of these
hierarchical algorithms. Based on the algebraic representation,
the application of the kernel matrix as well as its inverse, or
its factorization can be processed in quasi-linear ($O(n\log^kn)$)
operations. Due to the tree partitioning, the extension to high
dimensional kernel matrices is problematic. Both the computational
and storage costs grow exponentially with the data dimension, spoiling
the $O(n)$ or $O(n\log n)$ complexity of those algorithms.

In this paper, we adopt some ideas from hierarchical matrices and
butterfly factorizations~\cite{li2015butterfly, Li2018}, and propose
a Block Basis Factorization (BBF) structure that generalizes the
traditional low-rank matrix, along with its efficient construction
algorithm. We apply this scientific computing based method to a
range of problems, with an emphasis on machine learning problems.
We will show that the BBF structure achieves significantly higher
accuracy than plain low-rank matrices, given the same memory budget,
and the construction algorithm has a linear in $n$ complexity for
\replaced{R1}{5}{most}{many} machine learning \replaced{R1}{6}{kernels}{kernel learning tasks}.

The key of our structure is realizing that in most machine learning
applications, the sub-matrices representing the interactions from
one cluster to the entire data set are numerically low-rank. For
example, Wang et al.~\cite{wang2017numerical} mathematically proved that
if the diameter of a cluster $\mathcal{C}$ is smaller than that of
the entire dataset $\mathcal{X}$, then the rank of the sub-matrix
$\mathcal{K}(\mathcal{C}, \mathcal{X})$ is lower than the rank
of the entire matrix $\mathcal{K}(\mathcal{X}, \mathcal{X})$.
If we partition the data such that each cluster has a small
diameter, and the clusters are as far apart as possible from
each other, then we can take advantage of the low-rank property
of the sub-matrix $\mathcal{K}(\mathcal{C}, \mathcal{X})$ to
obtain a presentation that is more memory-efficient than low-rank
representations.

The application of our BBF structure is not limited to RBF kernels
or machine learning applications. There are many other types of
structured matrices for which the conventional low-rank approximations
may not be satisfactory. Examples include, but are not limited to,
covariance matrices from spatial data~\cite{stein2014limitations},
frontal matrices in the multi-frontal method for sparse matrix
factorizations~\cite{amestoy2015improving}, and kernel method in
dynamic systems~\cite{Bouvrie2017}.

\subsection{Main Contributions}
\label{main_results}

Our main contribution is three-fold. First, we showed that for
classification datasets whose decision surfaces have small radius
of curvature, a small kernel bandwidth parameter is needed for
high accuracy. Second, we proposed a novel matrix approximation
structure that extends the applicability of low-rank methods to
matrices whose ranks are higher. Third, we developed a corresponding
construction algorithm that produces errors with small variance (the
algorithm uses randomized steps), and that has linear, \emph{i.e.},
$O(n)$ complexity for most machine learning \replaced{R1}{6}{kernels}{kernel learning tasks}. Specifically,
our contributions are as follows.
\begin{itemize}
    \item For several datasets with imbalanced classes, we observed
    an improvement in accuracy for smaller classes when we set the
    kernel bandwidth parameter to be smaller than that selected from
    a cross-validation procedure. We attribute this to the nonlinear
    decision surfaces, which we quantify as the smallest radius of
    curvature of the decision boundary.

    \item We proposed a novel matrix structure called the Block Basis
    Factorization (BBF) for machine learning applications. {BBF}
    approximates the kernel matrix with linear memory and is efficient
    for a wide range of bandwidth parameters.

    \item We proposed a construction algorithm for the {BBF} structure
    that is accurate, stable and linear for \replaced{R1}{5}{most}{many} machine learning
    kernels. This is in contrast to most algorithms to calculate the
    singular value decomposition (SVD) which are more accurate but
    lead to a cubic complexity, or na\"ive random sampling algorithms
    (\emph{e.g.}, uniform sampling) which are linear but often
    inaccurate or unstable for incoherent matrices. We also provided
    a fast pre-computation algorithm to search for suggested input
    parameters for {BBF}.
\end{itemize}

Our algorithm involves three major steps. First, it divides the data
into $k$ distinct clusters, permutes the matrix according to these
clusters. The permuted matrix has $k^2$ blocks, each representing the
interactions between two clusters.  Second, it computes the column
basis for every row-submatrix (the interactions between one cluster
and the entire dataset) by first selecting representative columns
using a randomized sampling procedure and then compressing the columns
using a randomized {SVD}.  Last, it uses the corresponding column-
and row- basis to compress each of the $k^2$ blocks, also using a
randomized sub-sampling algorithm. Consequently, our method computes
an approximation for the $k^2$ blocks using a set of only $k$ bases.
The resulting framework yields a rank-$R$ approximation and achieves
a similar accuracy as the best rank-$R$ approximation\added{R1}{7}{, where $R$ refers to the approximation rank}. The memory
complexity for BBF is $\calO(nR/k+R^2)$, where $k$ is upper bounded
by $R$. This should be contrasted with a low-rank scheme that gives a
rank-$R$ approximation with $\calO(nR)$ memory complexity. BBF achieves
a similar approximation accuracy \added{R1}{8}{to the best rank-$R$ approximation} with a factor of $k$ saving on memory.

\subsection{Related Research}
\label{related_research}

There is a large body of research that aims to accelerate kernel
methods by low-rank approximations~\cite{golub2012matrix}.  Given a
matrix $K \in \mathbb{R}^{n \times n}$, a rank-$r$ approximation of
$K$ is given by $K \approx U V^\top$ where $U, V \in \mathbb{R}^{n
\times r}$, and $r$ is related to accuracy.  The SVD provides
the most accurate rank-$r$ approximation of a matrix in terms of
both 2-norm and Frobenius-norm; however, it has a cubic cost.
Recent work \cite{sarlos2006improved,liberty2007randomized,
halko2011finding,mahoney2011randomized} has reduced the cost to
$\mathcal{O}(n^2r)$ using randomized projections. These methods
require the construction of the entire matrix to proceed.  Another
line of the low-rank approximation research is the Nystr\"om
method \cite{drineas2005nystrom,gittens2016revisiting,Bac13},
which avoids constructing the entire matrix.  A na\"{\i}ve
Nystr\"om algorithm uniformly samples columns and reconstructs
the matrix with the sampled columns, which is computationally
inexpensive, but which works well only when the matrix has
uniform leverage scores, \emph{i.e.}, low coherence.  Improved
versions \cite{engquist2009fast,zhang2010clustered,drineas2012fast,
gittens2016revisiting,alaoui2014fast} of Nystr\"om have been proposed
to provide more sophisticated ways of column sampling.

There are several methods proposed to address the same
problem as in this paper.  The clustered low-rank approximation
(CLRA)~\cite{savas2011clustered} performs a block-wise low-rank
approximation of the kernel matrix from social network data with
quadratic construction complexity.  The memory efficient kernel
approximation (MEKA) \cite{si2014memory} successfully avoids the
quadratic complexity in CLRA.  Importantly, these previous methods
did not consider the class size and parameter size issues as we did
in detail.  Also, in our benchmark, we found that under multiple
trials, MEKA is not robust, \emph{i.e.}, it often failed to be
accurate and produced large errors. This is due to its inaccurate
structure and its simple construction algorithm. We briefly discuss
some significant differences between MEKA and our algorithm. In terms
of the structure, the basis in {MEKA} is computed from a smaller column
space and is inherently a less accurate representation, making it
more straightforward to achieve a linear complexity; in terms of the
algorithm, the uniform sampling method used in MEKA is less accurate
and less stable than the sophisticated sampling method used in {BBF}
that is strongly supported by theory.

There is also a strong connection between our algorithm and the
improved fast Gauss transform (IFGT)~\cite{yang2003improved},
which is an improved version of the fast Gauss
transform~\cite{Greengard1991}. Both BBF and IFGT use a clustering
approach for space partitioning. Differently, the IFGT approximates
the kernel function by performing an analytic expansion, while BBF
uses an algebraic approach based on sampling the kernel matrix. This
difference has made BBF more adaptive and achieve higher approximation
accuracy. Along the same line of adopting ideas from hierarchical
matrices, Chen et al.~\cite{Chen2017} combined the hierarchical
matrices and the Nystr\"om method to approximate kernel matrices
rising from machine learning applications.

The paper is organized as follows. \autoref{sec:bandwidth}
discusses the motivations behind extending low-rank structures and
designing efficient algorithms for higher-rank kernel matrices.
\autoref{sec:BBF} proposes a new structure that better approximates
higher-rank matrices and remains efficient for lower-rank
matrices, along with its efficient construction algorithm.  Finally,
\autoref{sec:results} presents our experimental results, which show the
advantages of our proposed BBF over the-state-of-arts in terms of the
structure, algorithm and applications to kernel regression problems.

\section{Motivation: kernel bandwidth and class size}
\label{sec:bandwidth}

In this section, we discuss the motivations behind designing an
algorithm that remains computationally efficient when the matrix
rank increases. Three main motivations are
as follows. First, the matrix rank depends strongly on the kernel
bandwidth parameters (chosen based on the particular problem), the
smaller the parameter, the higher the matrix rank. Second, a small
bandwidth parameter (higher-rank matrix) imposes high nonlinearity on
the model, hence, it is useful for regression problems with non-smooth
function surfaces and classification problems with complex decision
boundaries. Third, when the properties of smaller classes
are of particular interest, a smaller bandwidth parameter would be
appropriate and the resulting matrix would be of higher rank. In the
following we focus on the first two motivations.

\subsection{Dependence of matrix rank on kernel parameters}
\label{sec:rank_and_h}

\begin{figure}[htbp]
    \centering
    \begin{subfigure}[b]{0.45\textwidth}  
        \includegraphics[width=\textwidth]{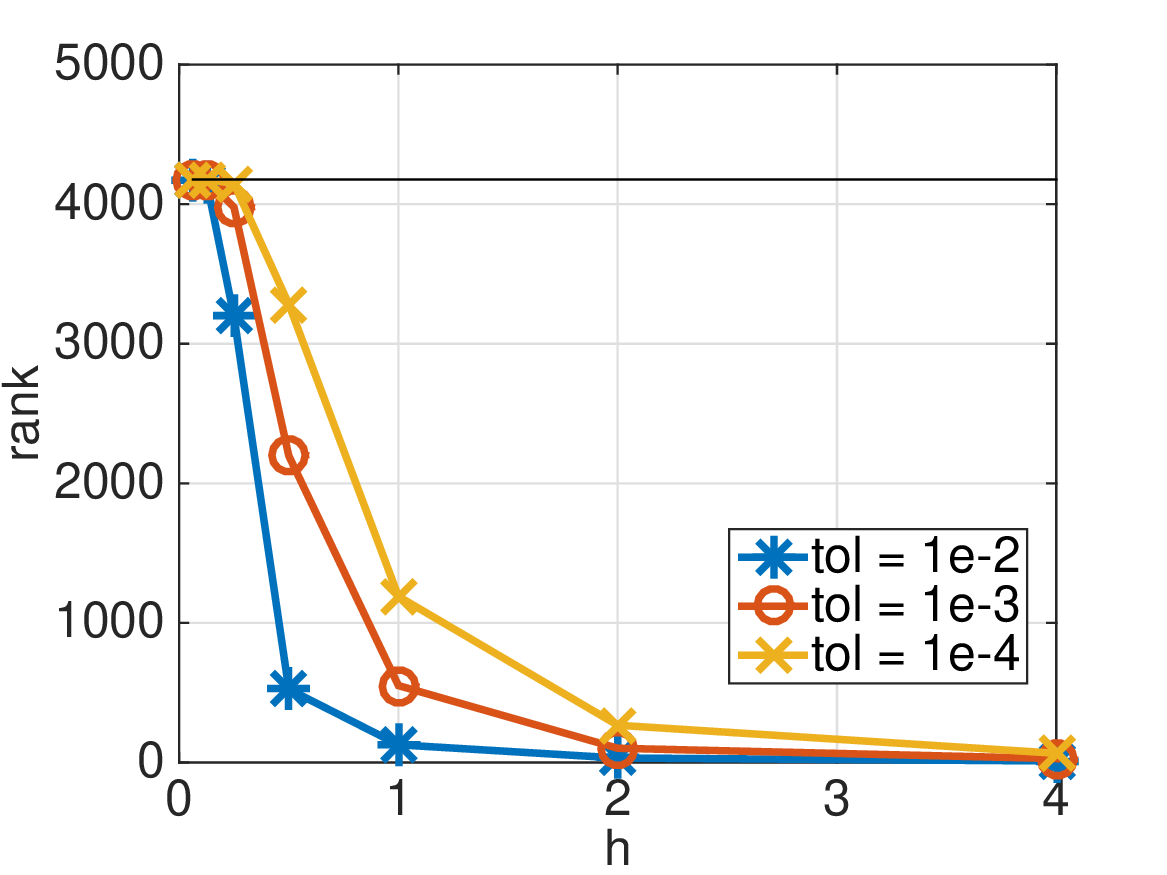}
        \caption{Numerical ranks}
    \end{subfigure}
    \begin{subfigure}[b]{0.45\textwidth}  
        \includegraphics[width=\textwidth]{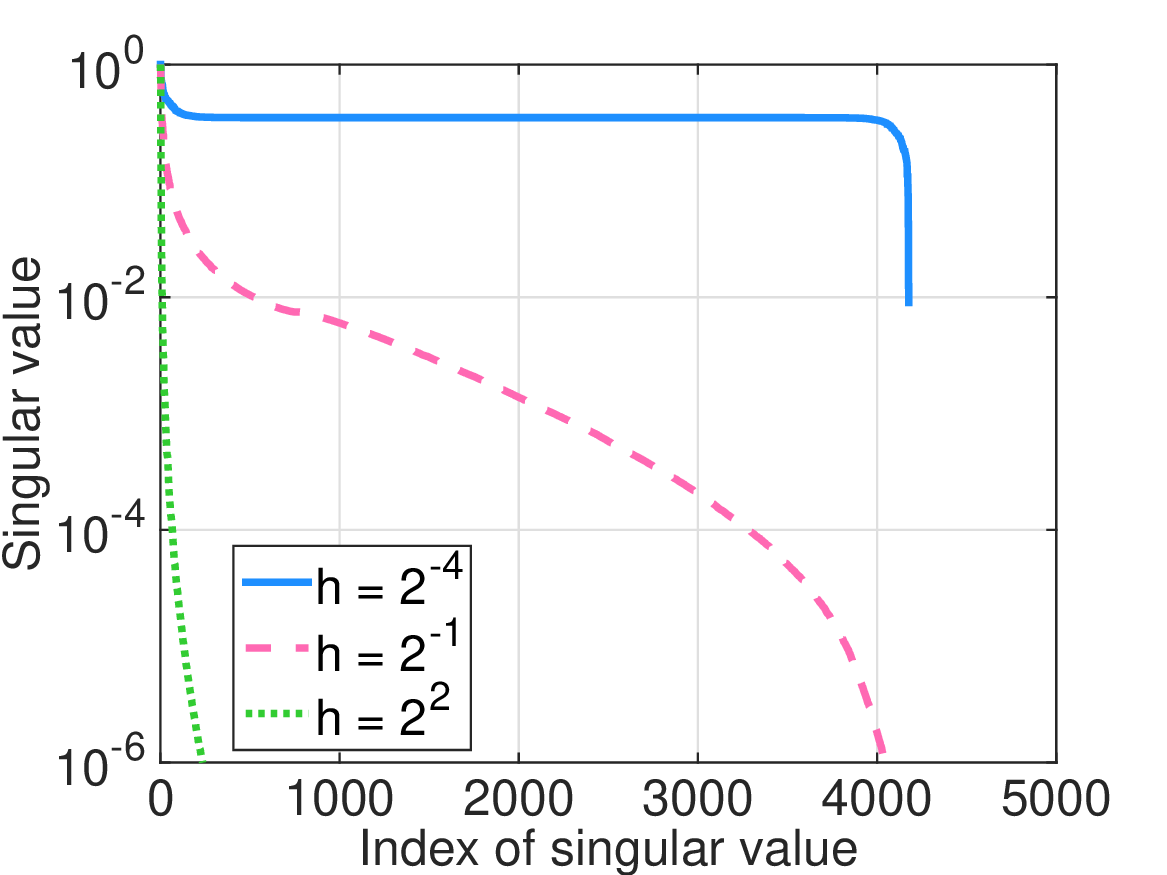}
        \caption{Singular value decay}
    \end{subfigure}

    \caption{Numerical ranks of the kernel matrix versus $h$. The
    data used is Abalone, and is normalized in each dimension.
    The numerical rank is computed as $\text{argmin}_k (\sigma_k <
    tol \cdot \sigma_1)$, where $\sigma_1 \ge \sigma_2 \ge, \ldots,
    \ge \sigma_n$ are the singular values, and $tol$ is the tolerance.
    The plot on the right shows the decay patterns of singular values
    for varying $h$.}
    \label{fig:rank_vs_h}
\end{figure}

We consider first the bandwidth parameters, and we will show that
the matrix rank depends strongly on the parameter.
Take the Gaussian kernel $\exp(-\|\vecx-\vecy\|^2/h^2)$
as an example. The bandwidth $h$ controls the function
smoothness. As $h$ increases, the function becomes more smooth,
and consequently, the matrix numerical rank decreases.
\autoref{fig:rank_vs_h} constructs a matrix from a real
dataset and shows the numerical rank versus $h$ with varying
tolerances $tol$.  As $h$ increases from $2^{-4}$ to $2^2$, the
numerical rank decreases from full (4177) to low (66 with tol =
$10^{-4}$, 28 with tol = $10^{-3}$, 11 with tol = $10^{-2}$).

Low-rank matrix approximations are efficient in the large-$h$
regime, and in such regime, the matrix rank is low. Unfortunately,
in the small-$h$ regime, they fall back to models with quadratic
complexity. One natural question is whether the situation where
a relatively small $h$ is useful occur in machine learning, or
whether low-rank methods are sufficient. We answer this question
in the following section, where we study kernel classifiers on real
datasets and investigate the influence of $h$ on accuracy.

\subsection{Optimal kernel bandwidth}
\label{sec:realdata}

We study the optimal bandwidth parameters used in practical situations,
and in particular, we consider kernel classifiers. In practice, the
parameter $h$ is selected by a cross-validation procedure combined with
a grid search, and we denote such parameter as $h_{CV}$. For datasets
with wide intra-variability, we observed that the optimal parameters
of some small classes turned out to be smaller than $h_{CV}$. By small
classes, we refer to those with fewer points or smaller diameters.

\begin{figure}[htbp]
    \centering
    \begin{subfigure}[b]{0.45\textwidth}  
        \includegraphics[width=\textwidth]{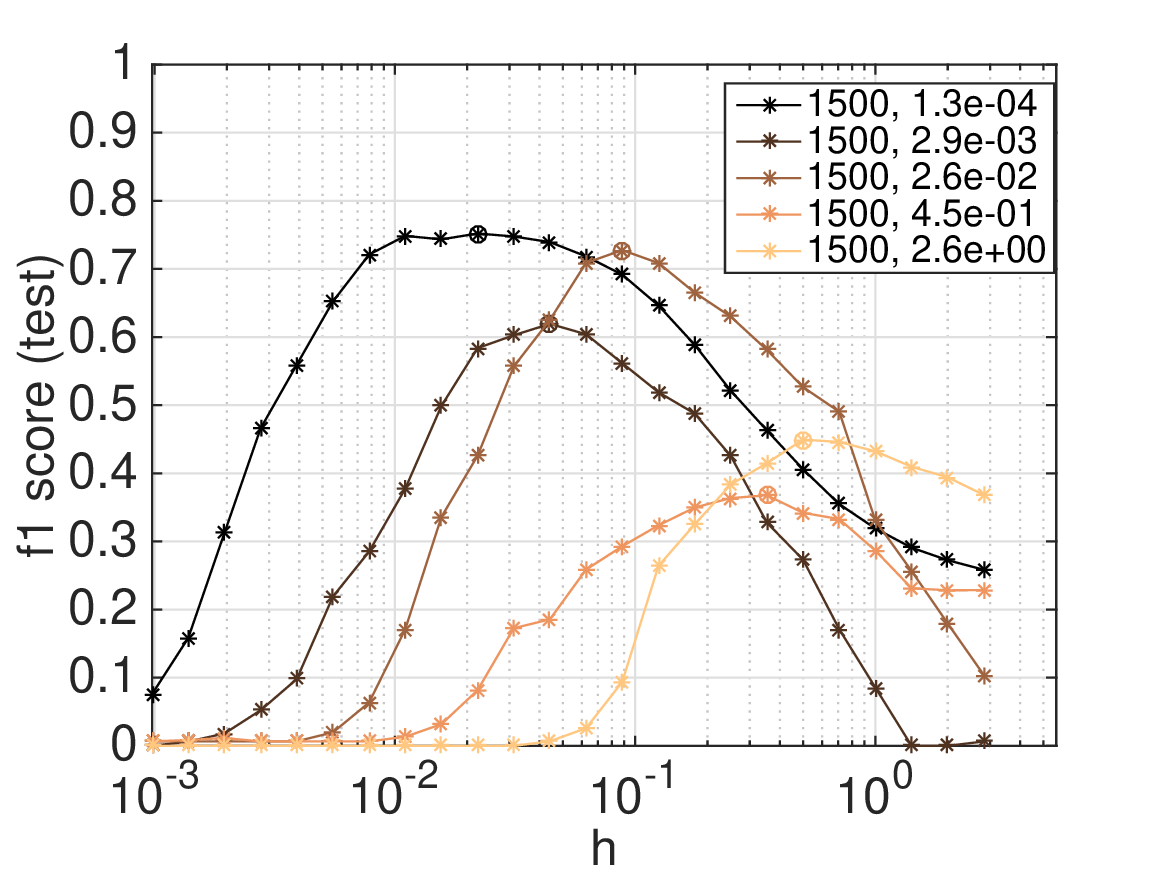}
        \caption{EMG, $h_{CV} = 0.17$}
    \end{subfigure}
    \begin{subfigure}[b]{0.45\textwidth}  
        \includegraphics[width=\textwidth]{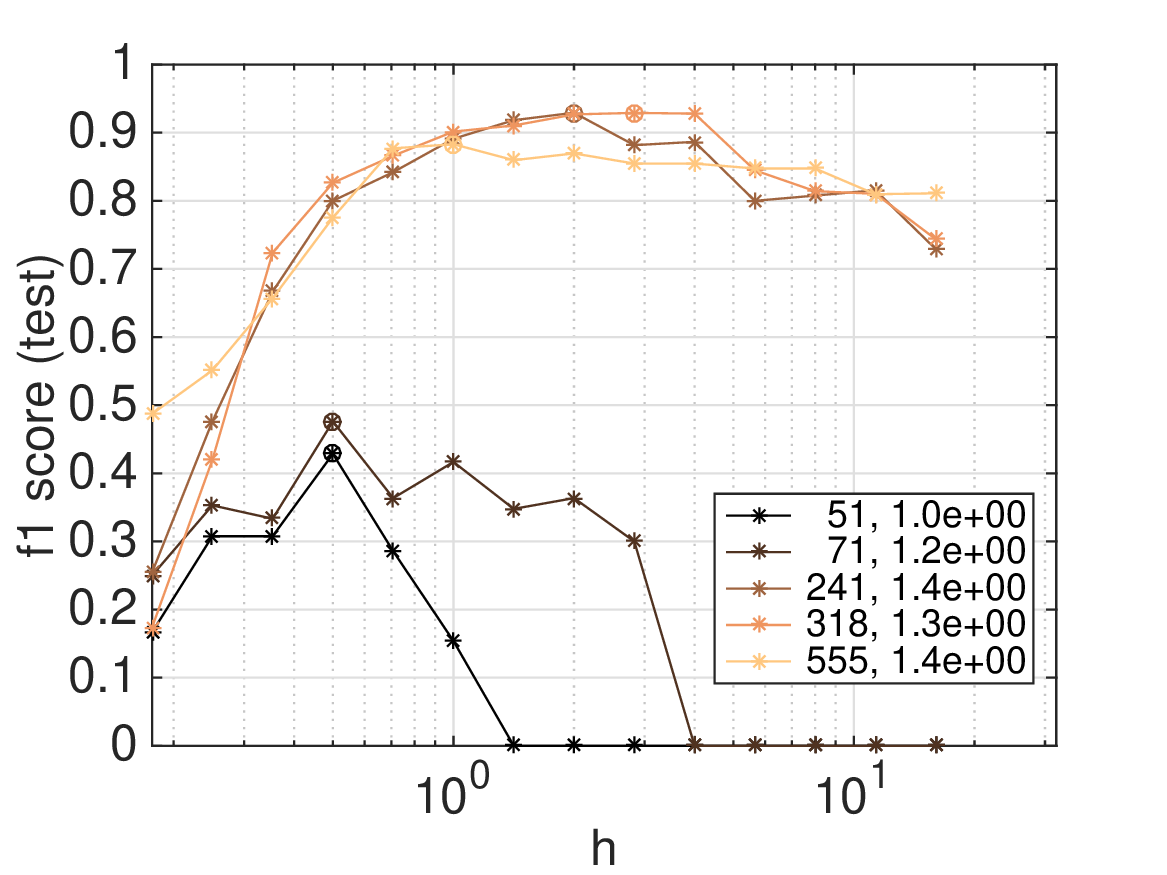}
        \caption{CTG, $h_{CV} = 1.42$}
    \end{subfigure}
    \begin{subfigure}[b]{0.45\textwidth}
        \includegraphics[width=\textwidth]{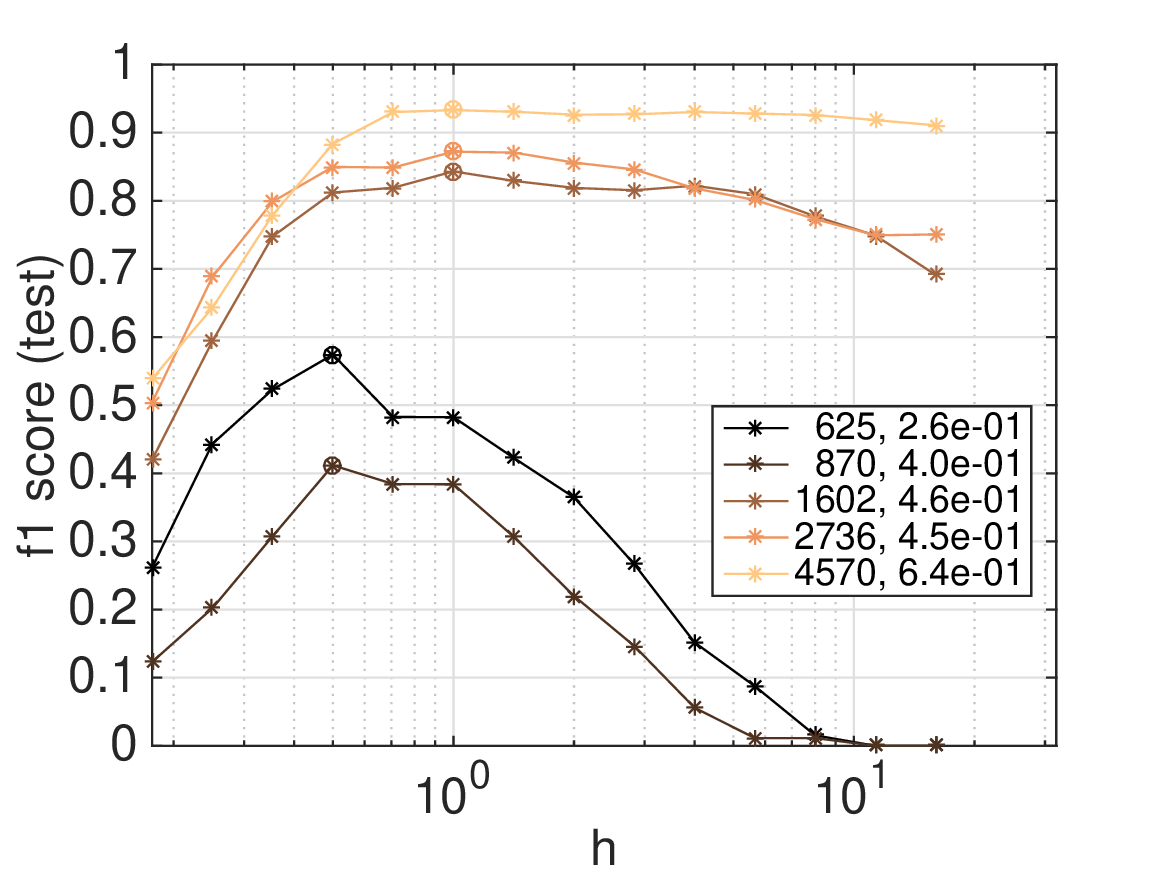}
        \caption{Otto, $h_{CV} = 1$}
    \end{subfigure} 
    \begin{subfigure}[b]{0.45\textwidth}
        \includegraphics[width=\textwidth]{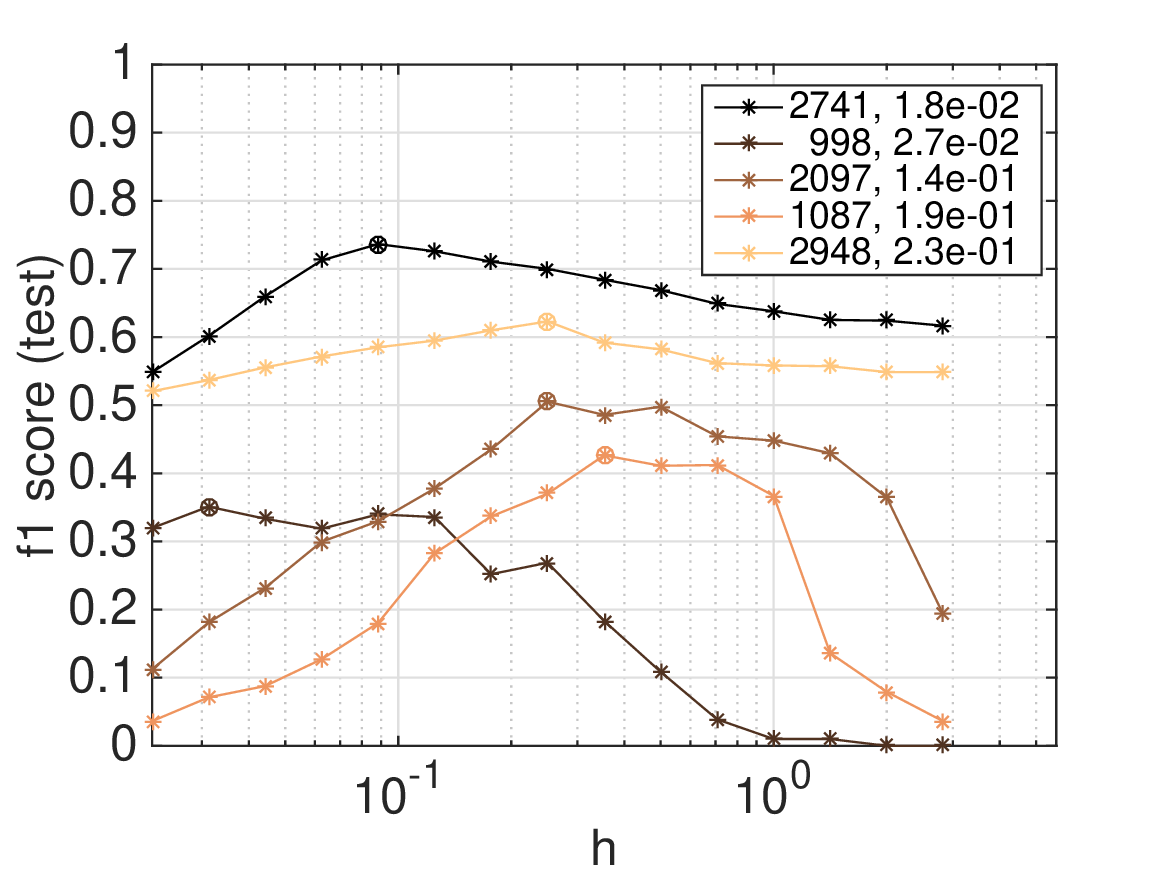}
        \caption{Gesture, $h_{CV} = 0.25$}
    \end{subfigure} 

    \caption{Test $F_1$ score of selected class for different datasets.
    Each curve represents one class, and the solid circle represents
    the maximum point along the curve.  The legend represents a pair
    ($n_i, r_i^2$), where $n_i$ is the number of point in each class,
    $r_i$ is the median distance to class center.  $h_{CV}$ is the
    parameter obtained from cross-validation. We see that for smaller
    classes (represented by darker colors), the $F_1$ score peaks at
    an $h$ that is smaller than $h_{CV}$.}
    \label{fig:different_h_behavior}
\end{figure}

\begin{figure}[htbp]
    \centering
    \begin{subfigure}[b]{0.45\textwidth}
        \includegraphics[width=\textwidth]{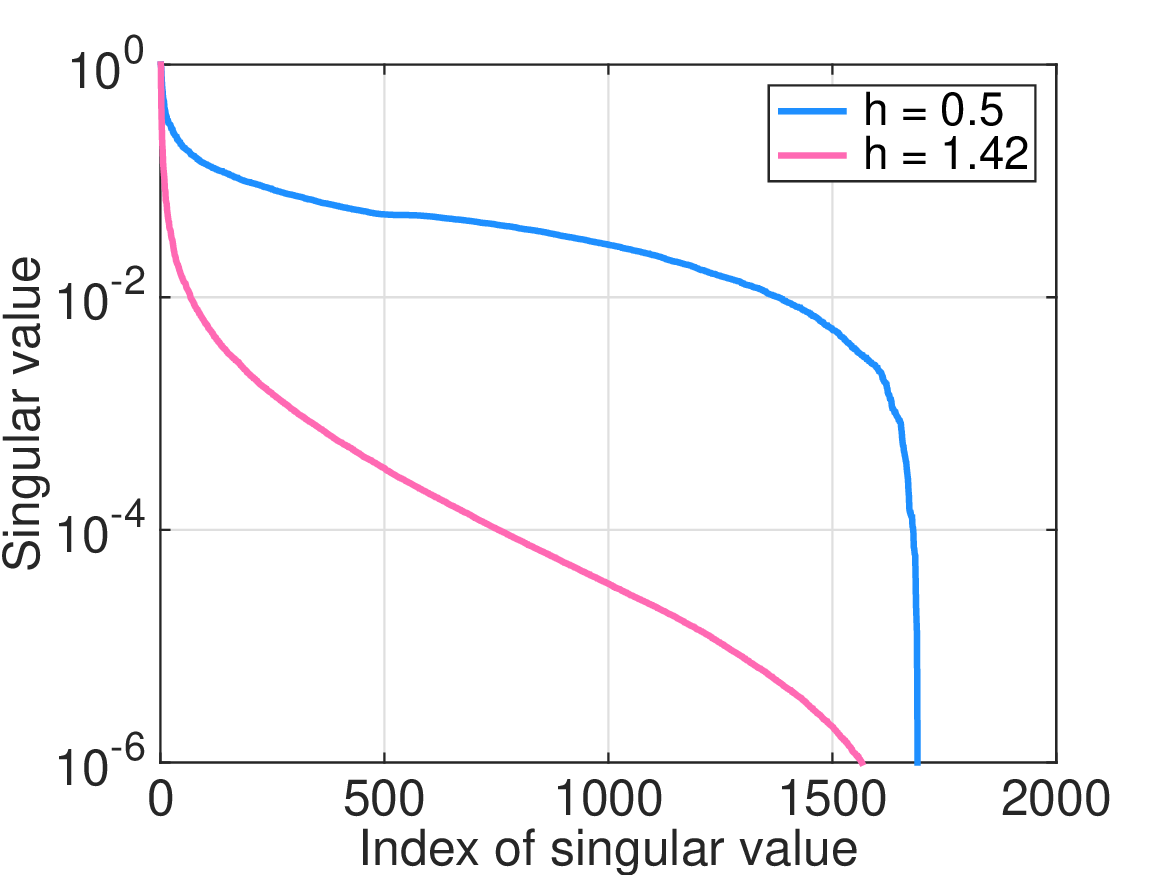}
        \caption{CTG}
    \end{subfigure} 
    \begin{subfigure}[b]{0.45\textwidth}
        \includegraphics[width=\textwidth]{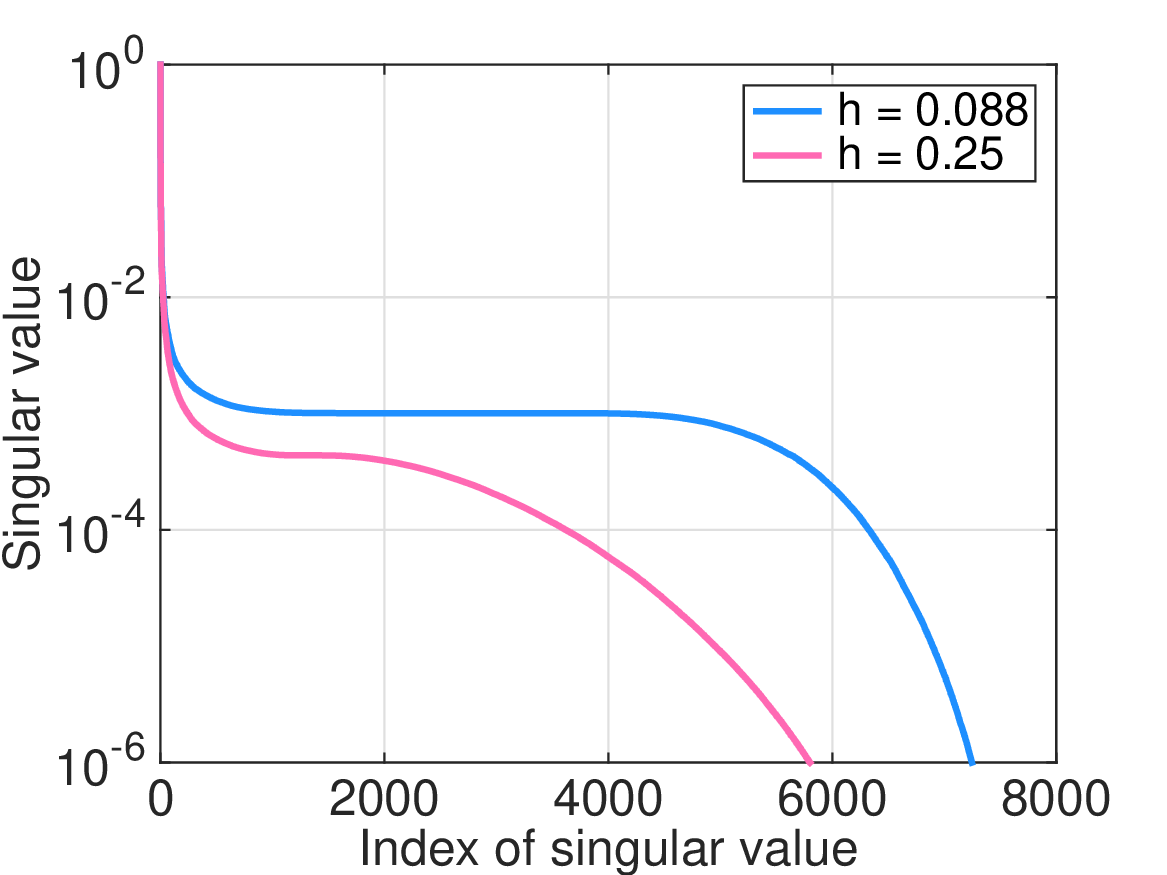}
        \caption{Gesture}
    \end{subfigure} 
    \caption{Singular value decay for CTG dataset and Gesture
    dataset on selected bandwidths. In subplot (a), $h_{CV} = 1.42$ and $h = 0.5$,
    respectively, is the optimal paraemter for the full dataset and the
    smallest dataset. In subplot (b), $h_{CV} = 0.25$ is the optimal
    parameter for the full dataset, and $h = 0.088$ achieves a good $F_1$
    score for clusters of small radii.} \label{fig:sval_ctg_twoh}
\end{figure}

Table~\ref{tab:datasets_classifcation} lists some classification
datasets with wide intra-variability. This class imbalance has
motivated us to study the individual performance of each class. We
found that there can be a significant discrepancy between $h_{CV}$
which is optimal overall for the entire dataset and the optimal
$h$ for a specific class. In \autoref{fig:different_h_behavior},
we used kernel SVM classifier under a wide range of $h$ and measure
the performance by $F_1$ score on the test data. The $F_1$ score is
the harmonic mean of the precision and recall, \emph{i.e.},
\[ \frac{2 \times \text{precision} \times
\text{recall}}{\text{precision} + \text{recall}}. \]
The data was randomly divided into 80\% training set and 20\% testing
set. \autoref{fig:different_h_behavior} shows the test $F_1$ score
versus $h$ for selected classes. We see that for some smaller classes
represented by darker colors, the $F_1$ score peaks at a value for
$h$ that is smaller than $h_{CV}$. Specifically, for the smallest
class (black curve) of each dataset, as $h$ increases from their own
optimal $h$ to $h_{CV}$, the test $F_1$ scores drop by 21\%, 100\%,
16\%, and 5\% for EMG, CTG, Otto and Gesture datasets, respectively.
To interpret the value of $h$ in terms of matrix rank, we plotted the
singular values for different values of $h$ for the CTG and Gesture
dataset in \autoref{fig:sval_ctg_twoh}. We see that when using
$h_{CV}$, the numerical rank is much lower than using a smaller $h$
which leads to a better performance on smaller classes.

The above observation suggests that the value of $h_{CV}$ is
mostly influenced by large classes and using $h_{CV}$ may degrade
the performance of smaller classes. Therefore, to improve the
prediction accuracy for smaller classes, one way is to reduce the
bandwidth $h$. Unfortunately, a decrease in $h$ increases the rank
of the corresponding kernel matrix, making low-rank algorithms
inefficient. Moreover, even if we create the model using $h_{CV}$,
as discussed previously, the rank of the kernel matrix will not be
very low in most cases. These altogether stress the importance of
developing an algorithm that extends the domain of applicability of
low-rank algorithms.
    
\begin{table}[htbp]
    \scriptsize

    \caption{Statistics for classification datasets and their
    \emph{selected} classes, $r_i$ is the median distance to the
    center for class $i$, $n_i$ is the number of points in class $i$.}
    \label{tab:datasets_classifcation}

    \centering
    \vspace{.4cm}

    \begin{tabular}{ c c c | c c c c c c}
        \toprule
        \multirow{2}{*}{\textbf{Data}} & \multirow{2}{*}{\textbf{n}}
        & \multirow{2}{*}{{\bf d}} & \multirow{2}{*}{} &
        \multicolumn{5}{c}{\bf Selected Classes} (other classes not
        shown) \\
        \hhline{~~~~-----}
        & & & & 1 & 2 & 3 & 4 & 5 \\
        \midrule
        \multirow{2}{*}{EMG} & \multirow{2}{*}{28,500} & \multirow{2}{*}{8}
        & $n_i$ & 1,500  & 1,500 & 1,500 & 1,500 & 1,500\\
        & & & $r_i^2$ & {$\bf 1.3 \times 10^{-4}$} & $2.9 \times 10^{-3}$
        & $2.6 \times 10^{-2}$ & $4.5 \times 10^{-1}$
        & { $\bf 2.6 \times 10^{0}$}\\
        \midrule
        \multirow{2}{*}{CTG} & \multirow{2}{*}{2,126}
        & \multirow{2}{*}{23} & $n_i$ & {\bf 51} & 71 & 241 & 318
        & {\bf 555}\\
        & & & $r_i^2$ & 1.0 & 1.2 & 1.4 & 1.3 & 1.4 \\
        \midrule
        \multirow{2}{*}{Gesture} & \multirow{2}{*}{9,873}
        & \multirow{2}{*}{32} & $n_i$  & 2,741 & 998 & 2,097 & 1,087
        & 2,948 \\
        & & & $r_i^2$ & {$\bf 1.8 \times 10^{-2}$}
        & $2.7 \times 10^{-2}$ & $1.4 \times 10^{-1}$
        & $1.9 \times 10^{-1}$ & {$\bf 2.3 \times 10^{-1}$} \\
        \midrule
        \multirow{2}{*}{Otto} & \multirow{2}{*}{20,000}
        & \multirow{2}{*}{93} & $n_i$ & {\bf 625} & 870 & 1602 & 2736
        & {\bf 4570}\\
        & & & $r_i^2$ & 0.26 & 0.40 & 0.46 & 0.45 & 0.64 \\
        \bottomrule
    \end{tabular}
\end{table}
            
\subsection{Factors affecting the optimal kernel bandwidth}
\label{sec:2ddata}

This section complements the previous section by investigating
some data properties that influence the optimal kernel bandwidth
parameter $h$.

\begin{figure}[htbp]
    \centering
    \includegraphics[width=0.4\textwidth]{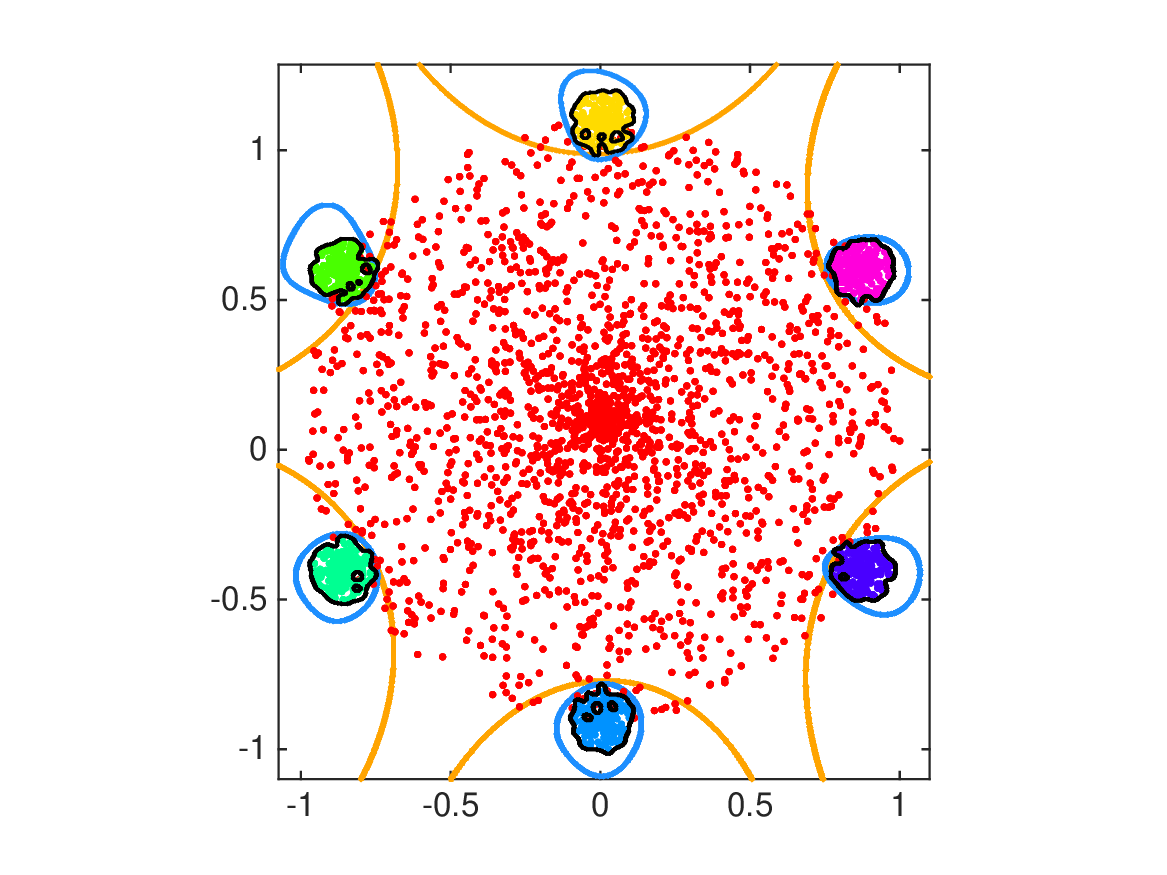}

    \caption{Decision boundary with varying smallest radius of
    curvatures. Dots represent data points and different classes
    are color coded. The curves represent the decision boundaries,
    of which the radii of curvature are large, median, and small for
    the orange, blue and black (those surrounding the small clusters)
    curves, respectively.}
    \label{fig:curvature}
\end{figure}

We studied synthetic two-dimensional data and our experiments
suggested that the optimal $h$ depends strongly on the \emph{smallest
radius of curvature} of the decision surface (depicted in
\autoref{fig:curvature}). By optimal we mean the parameter that
yields the highest accuracy, and if multiple such parameters exist,
we refer to the largest one to be optimal and denote it as $h^*$.
    
\begin{figure}[htbp]
    \centering
    \begin{subfigure}[b]{0.45\textwidth}  
        \includegraphics[width=\textwidth]{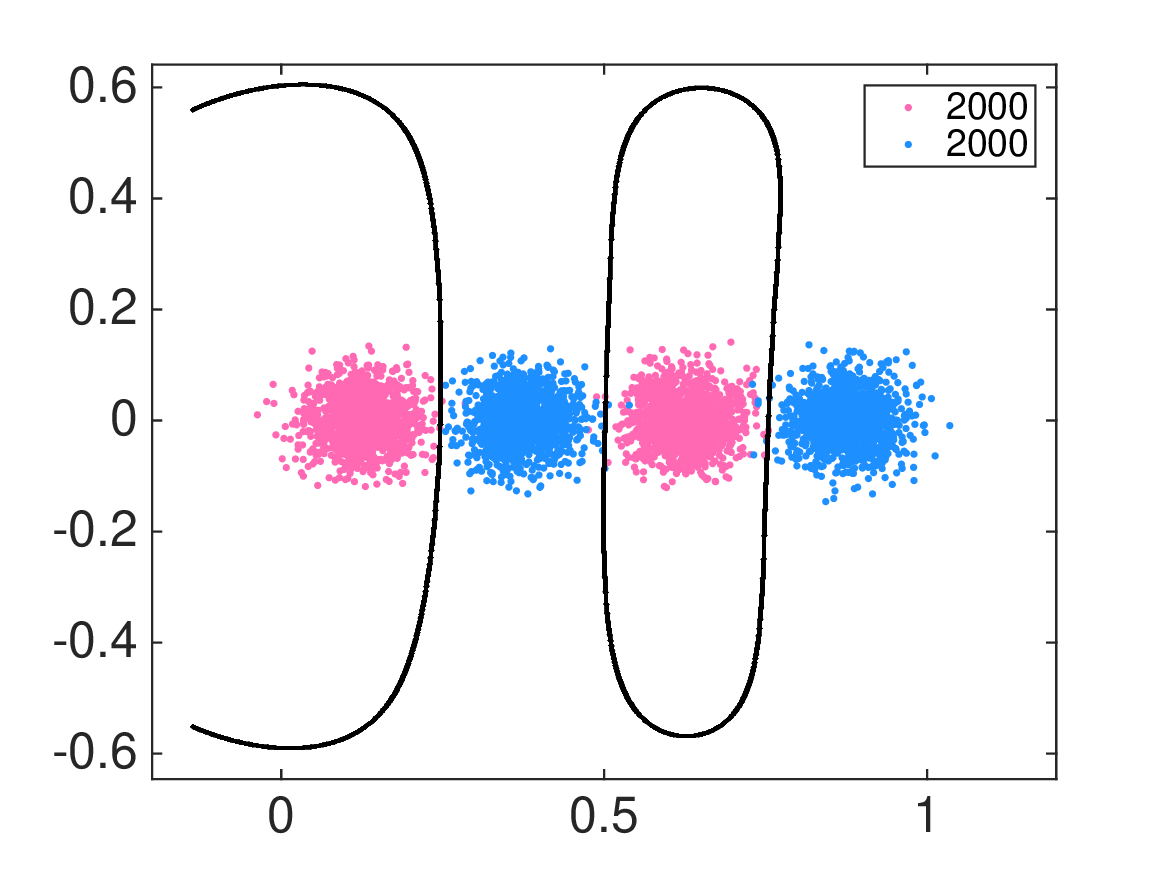}
        \caption{Data with 4 clusters}
    \end{subfigure}
    \begin{subfigure}[b]{0.45\textwidth}  
        \includegraphics[width=\textwidth]{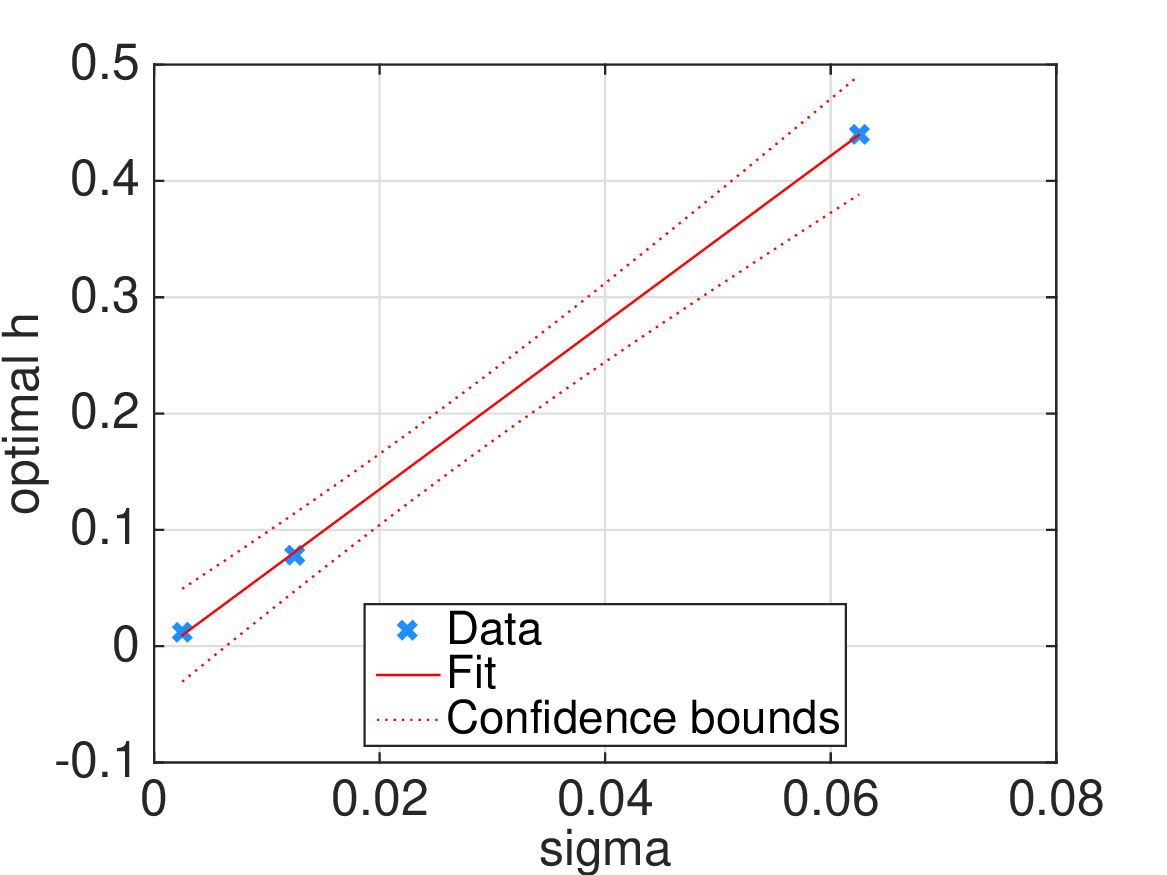}
        \caption{Relation of $\sigma$ and optimal $h$}
    \end{subfigure}
        
    \caption{Left: Data (4 clusters) with two alternating labels (20
    and 100 clusters cases are not shown). Each cluster is generated
    from a Gaussian distribution with standard deviation $\sigma$.
    The decision boundary (black curve) is associated with $h^* =
    0.44$.  Right: Linear relation of the standard deviation $\sigma$
    ($\approx$ half of cluster radius) of each cluster and $h^*$.}
    \label{fig:alternate}
\end{figure}

We first experimentally study the relation between $h^*$
and the smallest radius of curvature of the decision boundary.
\autoref{fig:alternate} shows Gaussian clusters with alternating
labels that are color coded. We decrease the radius of curvature
of the decision boundary by decreasing the radius of each cluster
while keeping the box size fixed. We quantify the smallest radius
of curvature of the decision boundary approximately by the standard
deviation $\sigma$ of each cluster. \autoref{fig:alternate}b shows
a linear correlation between $\sigma$ and $h^*$.
    
\begin{figure}[htbp]
    \centering
    \begin{subfigure}[b]{0.45\textwidth}
        \includegraphics[width=\textwidth]{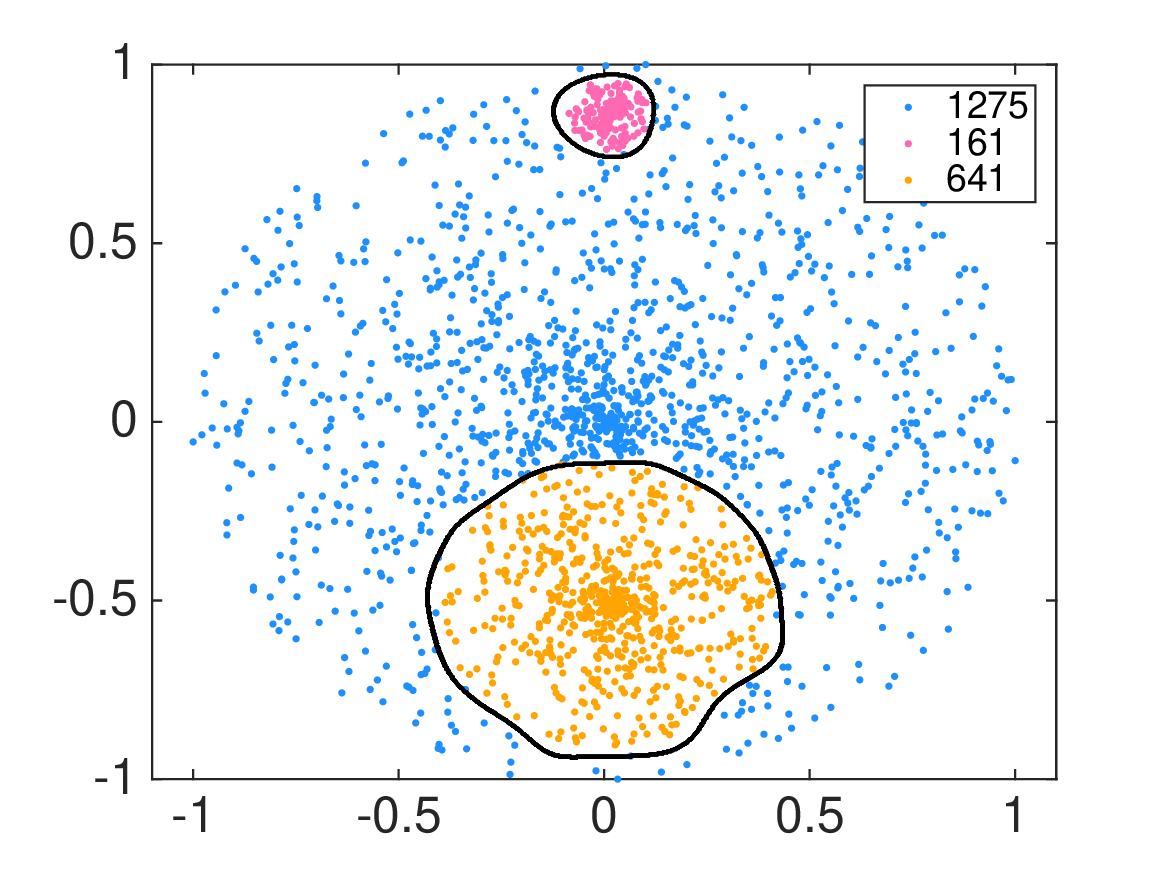}
        \caption{Data and decision boundary}
    \end{subfigure}
    \begin{subfigure}[b]{0.45\textwidth}
        \includegraphics[width=\textwidth]{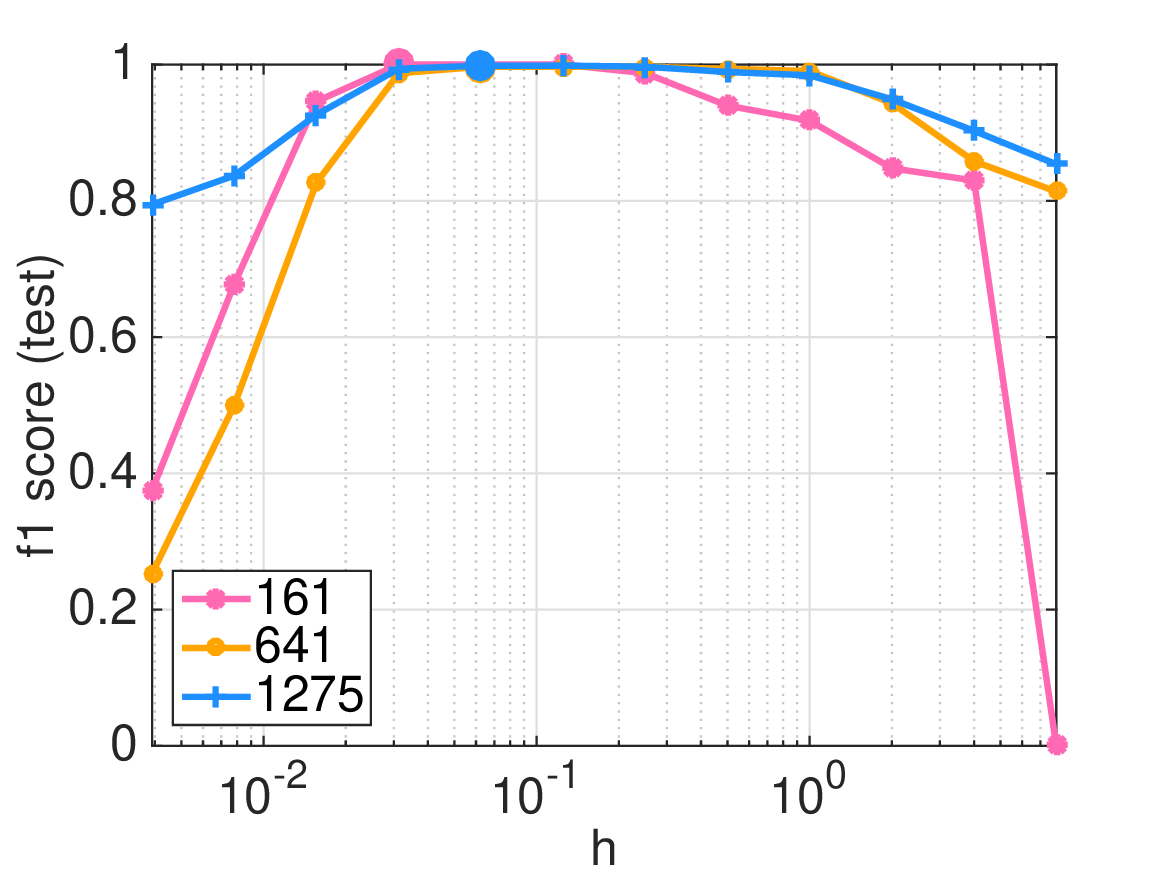}
        \caption{$F_1$ score for test data}
    \end{subfigure}

    \caption{Left: classes surrounded by a larger one; the legend
    shows the number of points in each class.  The decision boundary
    (black curve) is for $h = 0.125$.  Right: the test $F_1$ score
    for the test data versus $h$.}
    \label{fig:twocircles}
\end{figure}

\begin{figure}[htbp]
    \centering
    \begin{subfigure}[b]{0.45\textwidth}
        \includegraphics[width=\textwidth]{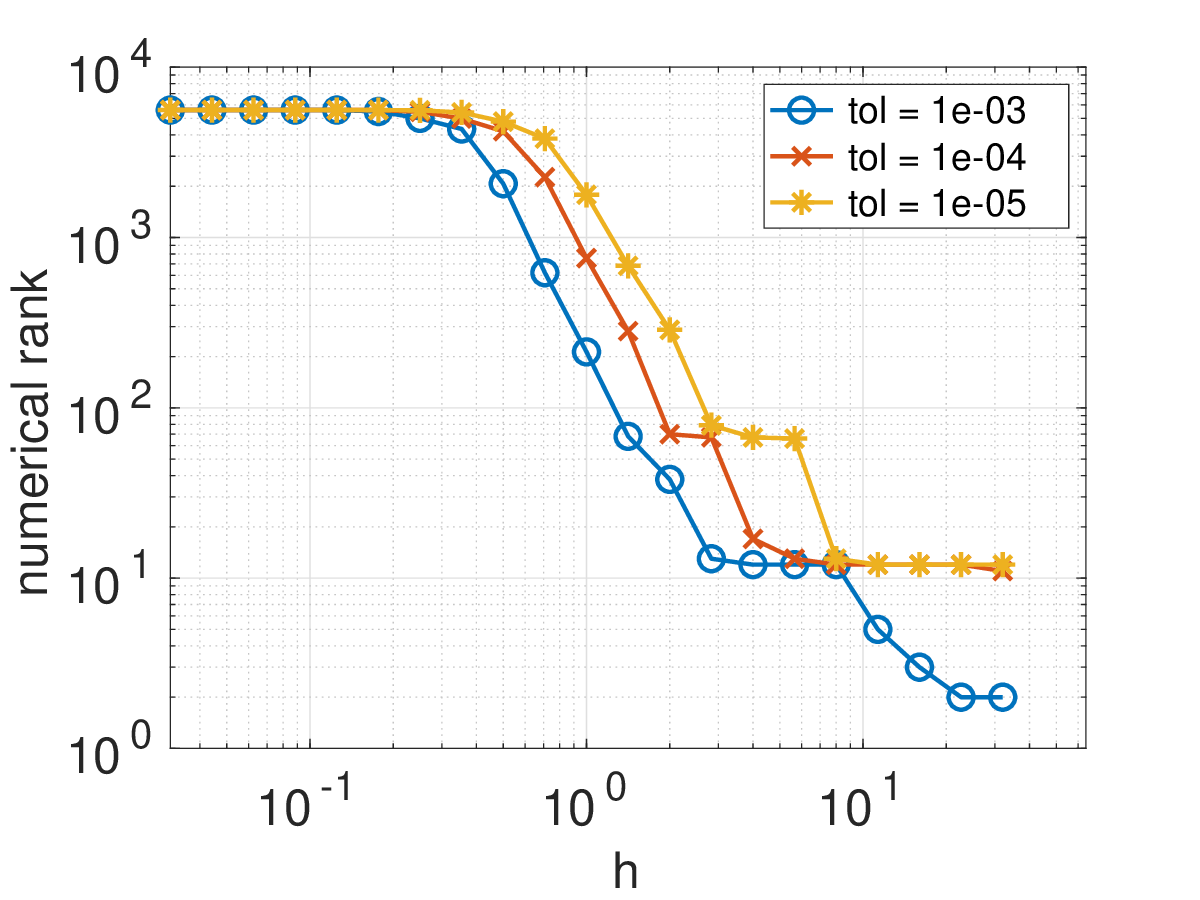}
        \caption{Numerical ranks}
    \end{subfigure}
    \begin{subfigure}[b]{0.45\textwidth}
        \includegraphics[width=\textwidth]{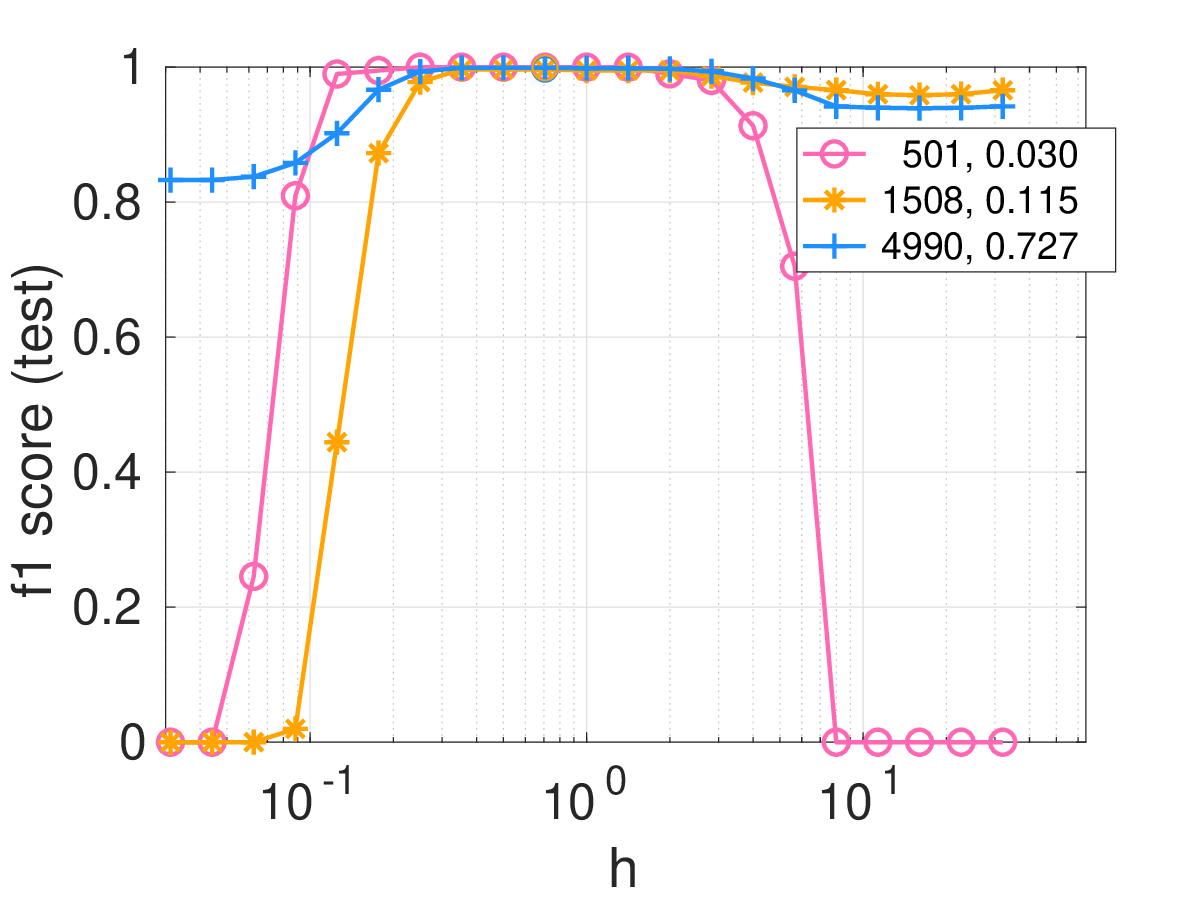}
        \caption{$F_1$ score for test data}
    \end{subfigure}

    \caption{Left: numerical ranks of the kernel matrix evaluated on
    the training data versus $h$.  Right: the $F_1$ score for the test
    data versus $h$. The synthetic data is in dimension 10 and has
    three clusters of different radii. The legend represents a pair
    $(n_i, r_i)$, where $n_i$ is the number of points in cluster $i$
    and $r_i$ is the cluster radius.} \label{fig:10dclusters}
\end{figure}

We study a couple more examples. \autoref{fig:twocircles} shows two
smaller circles with different radii surrounded by a large circle. For
this example, the smallest radius of curvature of the decision boundary
depends strongly on the cluster radius. Hence, the optimal $h$ for
the smaller class (pink colored) should be smaller than that for
the larger class (orange colored), which was verified by the $F_1$
score. Compared to the large cluster, the $F_1$ score for the small
cluster peaks at a smaller $h$ and drops faster as $h$ increases. A
similar observation was made in higher-dimensional data as well. We
generated two clusters of different radii which are surrounded by a
larger cluster in dimension 10. \autoref{fig:10dclusters} shows that
the intuition in dimension 2 nicely extends to dimension 10. Another
cluster example is in \autoref{fig:curvature}, which shows multiple
small clusters overlapping with a larger cluster at the boundary. The
3 reference decision boundaries correspond to $h$ being 1.5 (orange),
0.2 (blue) and 0.02 (black), respectively. The highest accuracy was
achieved at $h = 0.5$, which is close to the small cluster radius 0.2
and is large enough to tolerate the noises in the overlapping region.

The above examples, along with many that are not shown in this paper,
have experimentally suggested that the optimal parameter $h$ and the
smallest radius of curvature of the decision surface are positively
correlated. Hence, for datasets whose decision surfaces are highly
nonlinear, \emph{i.e.}, of small radius of curvature, a relatively
small $h$ is very likely needed to achieve a high accuracy.
 
In the following section, we will introduce our novel scheme to
accelerate kernel evaluations, which remains efficient in cases where
traditional low-rank methods are inefficient.
        
\section{Block Basis Factorization} 
\label{sec:BBF}

In this section, we propose the Block Basis Factorization
(BBF) that extends the availability of traditional low-rank
structures. \autoref{sec:BBF_structure} describes the BBF
structure. \autoref{sec:algo} proposes its fast construction algorithm.

\subsection{BBF Structure}
\label{sec:BBF_structure}

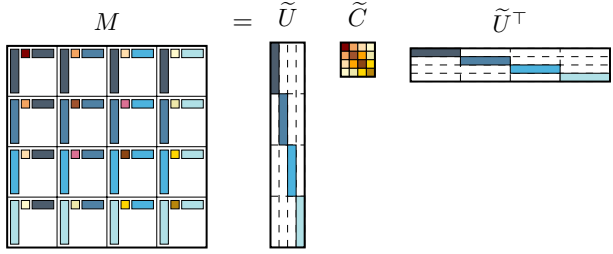
\begin{figure}[htbp]
    \centering
    \resizebox{3cm}{3.3cm}{
        \input{fig-M}
    }
    \begin{minipage}{.02\textwidth}
    \vspace{-5.9cm}$=$
    \end{minipage}
    \resizebox{!}{3.45cm}{
        \input{fig-U}
    }
    \resizebox{!}{3.45cm}{
        \input{fig-C}
    }
    \resizebox{3cm}{!}{
        \input{fig-VT}
    }\\
    \caption{$M = \widetilde{U} \widetilde{C} \widetilde{U}^\top$}
    \label{fig:bbf}
\end{figure}

This section defines and analyzes the \emph{Block Basis Factorization}
(BBF).  Given a symmetric matrix $M \in \mathbb{R}^{n
\times n}$ partitioned into $k$ by $k$ blocks, let $M_{i,j}$ denote
the $(i,j)$-th block for $i,j = 1, \ldots, k$.  Then, the {BBF} of $M$
is defined as: \abovedisplayskip=3pt \belowdisplayskip=3pt
\begin{equation} \label{eq:BBF}
    M = \widetilde{U} \widetilde{C} \widetilde{U}^\top,
\end{equation}
where $\widetilde{U}$ is a block diagonal matrix with the $i$-th
diagonal block $U_i$ being the column basis of $M_{i,j}$ for all $j$,
and $\widetilde{C}$ is a $k$ by $k$ block matrix with the $(i,j)$-th
block denoted by $C_{i,j} = U_i^\top M_{i,j} U_j$. The {BBF} structure
is depicted in \autoref{fig:bbf}.

We discuss the memory cost for the {BBF} structure. If the numerical
ranks of all the base $U_i$ are bounded by $r$, then the memory cost
for the {BBF} is $O(nr+(rk)^2)$.  Further, if $k\leq \sqrt{n}$ and $r$
is a constant independent of $n$, then the {BBF} gives a data-sparse
representation of matrix $M$.  In this case, the complexity for both
storing the {BBF} structure and applying it to a vector will be linear
in $n$.

It is important to distinguish between our {BBF} and a block low-rank
(BLR) structure~\cite{Amestoy2015}. There are two
main differences: 1). The memory usage of the BBF is much less than
BLR. BBF has one basis for all the blocks in the same row; while BLR
has a separate basis for each block. The memory for BBF is $nr +
(rk)^2$, whereas for BLR it is $2nkr$.  2). It is more challenging to
construct BBF in linear complexity while remaining accurate. A direct
approach using SVD to construct the low-rank base has a cubic cost;
while a simple randomized approach would be inaccurate and unstable.

In the next section, we will propose an efficient method to construct
the BBF structure, which uses randomized methods to reduce the cost
while still providing a robust approach. Our method is linear in $n$
for \replaced{R1}{5}{most kernels}{many kernels} used in machine learning applications.

\subsection{Fast construction algorithm for {BBF}}
\label{sec:algo}

In this section, we first introduce a theorem in
\autoref{sec:main_algo} that reveals the motivation behind our
{BBF} structure and addresses the applicable kernel functions.
We then propose a fast construction algorithm for BBF in
\autoref{sec:factorization}.

\subsubsection{Motivations}
\label{sec:main_algo}

Consider a RBF kernel function $\mathcal{K}: \mathbb{R}^d \times
\mathbb{R}^d \mapsto \mathbb{R}$. The following theorem in
\cite{wang2017numerical} provides an upper bound on the error for
the low-rank representation of kernel $\mathcal{K}$. The error is
expressed in terms of the function smoothness and the diameters of
the source domain and the target domain.
\begin{theorem} \label{thm:fourier_taylor}
    Consider a function $f$ and kernel $\calK(\vecx, \vecy ) =
    f(\norm{\vecx-\vecy}_2^2)$ with $\vecx=(x_1,\ldots,x_d)$ and
    $\vecy=(y_1,\ldots,y_d)$. We assume that $x_i \in [0,D/\sqrt{d}]$,
    $y_i \in [0,D/\sqrt{d}]$, where $D$ is a constant independent of
    $d$. This implies that $\|\vecx-\vecy\|_2^2\le D^2$. We assume
    further that there are $D_\vecx < D$ and $D_\vecy < D$, such that
    $\|\vecx_i-\vecx_j\|_2\le D_\vecx$ and $\|\vecy_i-\vecy_j\|_2\le
    D_\vecy$.

    Let $f_p(x) = \sum_{n} \mathcal{T} \circ f(x+4nD^2)$ be a
    $4D^2$-periodic extension of $f(x)$, where $\mathcal{T}(\cdot)$
    is 1 on $[-D^2, D^2]$ and smoothly decays to 0 outside of this
    interval. We assume that $f_p$ and its derivatives through
    $f_p^{(q-1)}$ are continuous, and the $q$-th derivative is
    piecewise continuous with its total variation over one period
    bounded by $V_q$.

    Then, $\forall~ M_f, M_t > 0$ with $9M_f \le M_t$, the kernel
    $\calK$ can be approximated in a separable form whose rank is at
    most $R = R(M_f, M_t, d) = 4M_f \binom{M_t+d}{d}$
    \begin{equation*}
        \calK(\vecx,\vecy) = \sum_{i=1}^{R} g_i(\vecx)h_i(\vecy)
        + \epsilon_{M_f, M_t}
    \end{equation*}
    The $L_\infty$ error is bounded by 
    \begin{equation*}
        |\epsilon_{M_f, M_t}| \le \norm{f}_\infty\left(\frac{D_\vecx
        D_\vecy}{D^2}\right)^{M_t+1} +\frac{V_q}{\pi q}
        \left(\frac{2D^2}{\pi M_f}\right)^q.
    \end{equation*}
\end{theorem}

In \autoref{thm:fourier_taylor}, the error is up bounded by the
summation of two terms. We first study the second term, which
is independent of \replaced{R1}{9}{$D_x$ or $D_y$}{$D_\vecx$ or $D_\vecy$}. The second term depends on
the smoothness of the function, and decays exponentially as the
smoothness of the function increases. \replaced{R1}{5}{Most}{Many} kernel functions used in
machine learning are sufficiently smooth; hence, the second term is
usually smaller than the first term.  Regarding the first term, the
domain diameter information influences the error through the factor
$\left(\frac{D_\vecx D_\vecy}{D^2}\right)^{M_t + 1}$, which suggests
for a fixed rank (positively related to $M_t$), reducing either
$D_\vecx$ or $D_\vecy$ reduces the error bound. It also suggests that
for a fixed error, reducing either $D_\vecx$ or $D_\vecy$ reduces
the rank.  This has motivated us to cluster points into distinct
clusters of small diameters, and by the theorem, the rank of the
submatrix that represents the local interactions from one cluster to
the entire dataset would be lower than the rank of the entire matrix.
    
Hence, we seek linear-complexity clustering algorithms that are able
to separate points into clusters of small diameters. $k$-means and
$k$-centers algorithms are natural choices. Both algorithms partition
$n$ data points in dimension $d$ into $k$ clusters at a cost of
$O(nkd)$ \added{R1}{10}{per iteration}. Moreover, they are based on the Euclidean distance between
points, which is consistent with the {RBF} kernels which are functions
of the Euclidean distance.  In practice, the algorithms converge to
slightly different clusters due to different objective functions, but
neither is absolutely superior. Importantly, the clustering results
from these algorithms yield a more memory efficient BBF structure
than random clusters.  A more task-specific clustering algorithm will
possibly yield better result; however, the main focus of this paper
is on factorizing the matrix efficiently, rather than proposing new
approaches to identify good clusters.

\begin{figure}[htp]
    \centering
    \begin{subfigure}{.35\linewidth}
        \centering
        \small
        \begin{tabular}{ccc}
            \toprule
            Cluster & Radius & Size \\
            \toprule
            1 & 42.4  &  752 \\
            2 & 41.3  &  603 \\ 
            3 & 37.6  &  729 \\
            4 & 24.3  &  733 \\
            5 & 24.2  & 1571 \\
            6 & 23.0  &  588 \\
            7 & 21.8  & 1006 \\
            8 & 21.6  &  421 \\
            9 & 21.4  &  236 \\
            10 & 21.3 &  855 \\
            \midrule
            Full & 62.8 & 7494 \\
            \bottomrule
        \end{tabular}
        \caption{}
        \label{tab:tenclusters}
    \end{subfigure}
    \begin{subfigure}{0.63\linewidth}
        \centering
        \includegraphics[width=\linewidth]{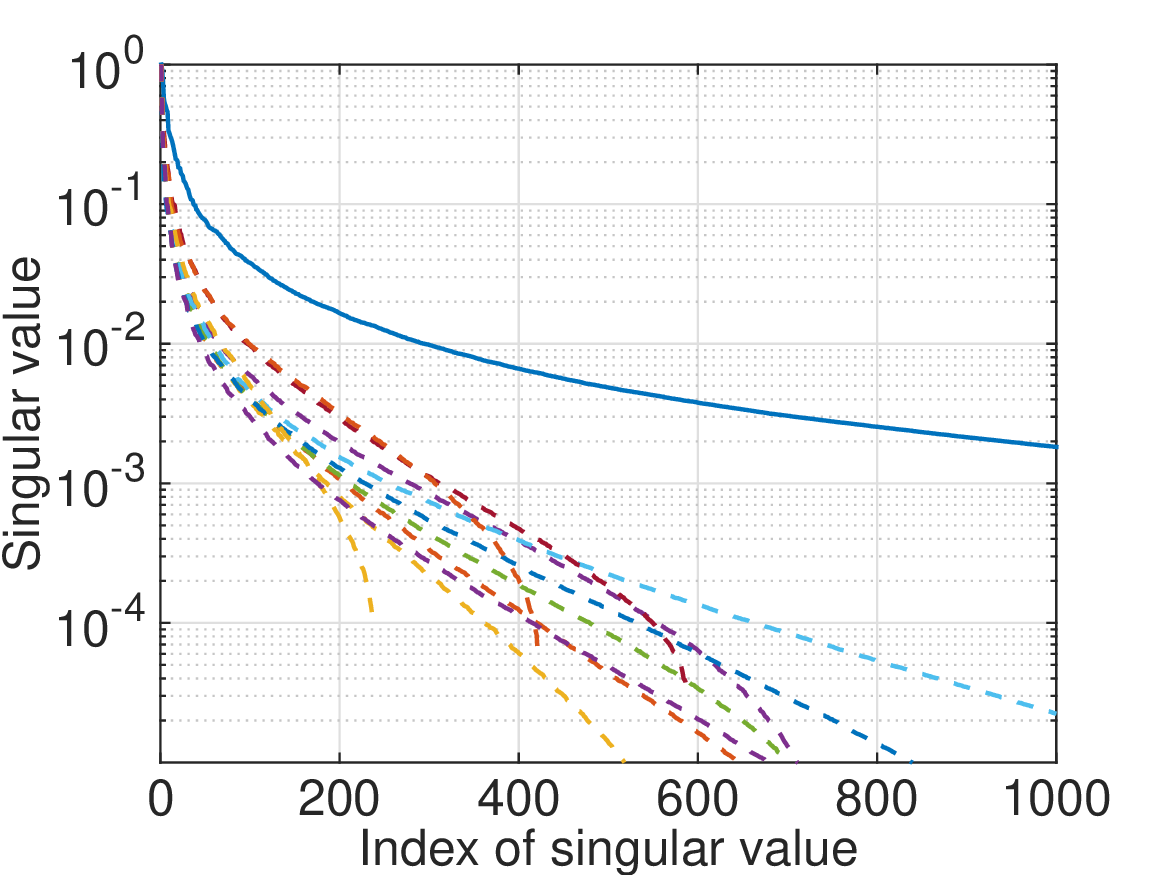}
        \caption{}
        \label{fig:sval_submat_forest}
    \end{subfigure}
    \caption{Left (a): Clustering result of the pendigits
    dataset. Right (b): Normalized singular value decay. In subplot
    (b), the solid curve represents the entire matrix $M$, and the
    dash curves represent the row-submatrices $M(C_i, :)$. The kernel
    used was the Gaussian kernel with bandwidth parameter $h = 2$.}
\end{figure}

We experimentally verify our motivation on a real-world dataset. We
clustered the pendigits dataset into 10 clusters ($C_1, C_2, \ldots,
C_{10}$) using the $k$-means algorithm, and reported the statistics
of each cluster in \autoref{tab:tenclusters}. We see that the radius
of each cluster is smaller than that of the full dataset. We further
plotted the normalized singular values of the entire matrix $M$ and its
sub-matrices $M(C_i, :)$ in \autoref{fig:sval_submat_forest}. Notably,
the normalized singular value of the sub-matrices shows a significantly
faster decay than that of the entire matrix. This suggests that
the ranks of sub-matrices are much lower than that of the entire
matrix. Hence, by clustering the data into clusters of smaller radius,
we are able to capture the local interactions that are missed by
the conventional low-rank algorithms which only consider global
interactions. As a result, we achieve a similar level of accuracy
with a much less memory cost.

\subsubsection{BBF Construction Algorithm}
\label{sec:factorization}

This section proposes a fast construction algorithm for the {BBF}
structure.  For simplicity, we assume the data points are evenly
partitioned into $k$ clusters, $\mathcal{C}_1,\dots,\mathcal{C}_k$,
and the numerical rank for each submatrix is $r$.  We first permute
the matrix according to the clusters:
\begin{equation} \label{eq:cluster}
    M = PKP^\top =
    \bordermatrix{
        {} & \mathcal{C}_1 & \mathcal{C}_2 & \cdots & \mathcal{C}_k \cr
        \mathcal{C}_1 & M_{1,1} & M_{1,2} & \cdots & M_{1,k} \cr
        \mathcal{C}_2 & M_{2,1} & M_{2,2} & \cdots & M_{2,k} \cr
        \vdots & \vdots  & \vdots  & \ddots & \vdots  \cr
        \mathcal{C}_k & M_{k,1} & M_{k,2} & \cdots & M_{k,k}
    },
\end{equation}
where $P$ is a permutation matrix, and $M_{i,j} =
\mathcal{K}(\mathcal{C}_i, \mathcal{C}_j)$ is the interaction matrix
between cluster $\mathcal{C}_i$ and cluster $\mathcal{C}_j$.

Our fast construction algorithm consists of two components: basis
construction and inner matrix construction.  In the following,
we adopt Matlab's notation for submatrices. We use the colon to
represent $\texttt{1:end}$, \emph{e.g.}, $M_{i,:} = \begin{pmatrix}
M_{i,1} & \cdots & M_{i,k} \end{pmatrix}$, and use the index vectors
$\mathcal{I}$ and $\mathcal{J}$ to represent sub-rows and sub-columns,
\emph{e.g.}, $M(\mathcal{I}, \mathcal{J})$ represents the intersection
of rows and columns whose indices are $\mathcal{I}$ and $\mathcal{J}$,
respectively.

{\bf 1. Basis construction} 

We consider first the basis construction algorithm. The most accurate
approach is to explicitly construct the submatrix $M_{i,:}$ and apply
an {SVD} to obtain the column basis; regrettably, it has a cubic cost
to compute all the bases. Randomized SVD \cite{halko2011finding}
reduces the cost to quadratic while being accurate; however, a
quadratic complexity is still expensive in practice. In the following,
we describe a linear algorithm that is accurate and stable. Since
the proposed algorithm adopts randomness, by ``stable'' we mean the
variance of the output is small under multiple runs. The key idea is
to restrict us in a subspace by sampling columns of large volume.
    
The algorithm is composed of two parts. In the first part, we select
some columns of $M_{i,:}$ that are representative of the column
space. By representative, we mean the $r$ sampled columns have volume
approximating the maximum $r$-dimensional volume among all column
sets of size $r$. In the second part, we apply the randomized SVD
algorithm to the representative columns to extract the column basis.

{\emph{Part 1: Randomized sampling algorithm}}
    
We seek a sampling method that samples columns with approximate maximum
volume. Strong rank revealing QR (RRQR) \cite{gu1996efficient} returns
columns whose volume is proportional to the maximum volume obtained by
SVD. QR with column pivoting (pivoted QR) is a practical replacement
for the strong RRQR due to its inexpensive computational cost. To
ensure a linear complexity, we use the pivoted QR factorization with
a randomized approach.
    
\begin{figure}[htbp]
    \centering
    \includegraphics[width=.8\textwidth]{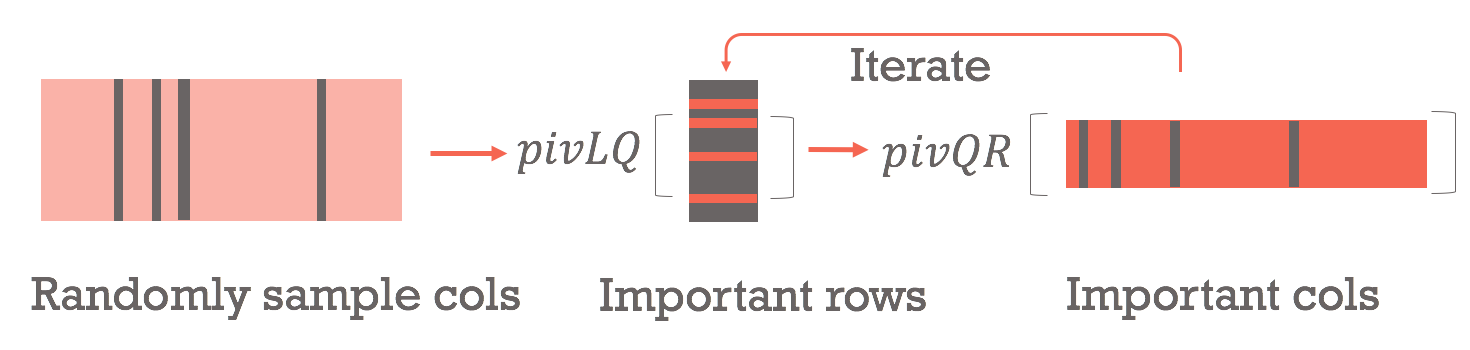}

    \caption{A pictorial description of the sampling algorithm.
    We start with sampling random columns, and iterate between
    important rows (by pivoted LQ) and important columns (by pivoted
    QR) to obtain our refined important columns. This procedure is
    usually repeated for a few times to ensure the stability of the
    important indices.}
    \label{fig:sample}
\end{figure}

We describe the randomized sampling method \cite{engquist2009fast}
used in our BBF algorithm; the algorithm detail is in
\autoref{alg:randomized_sampling} with the procedure depicted in
\autoref{fig:sample}. The complexity of sampling $r$ columns from an
$m \times n$ matrix is $O(r^2(m+n))$. The size of the output index
sets $\Pi_r$ and $\Pi_c$ could grow as large as $qr$, but it can be
controlled by some practical algorithmic modifications. One is that
given a tolerance, we truncate the top columns based on the magnitudes
of the diagonal entries of matrix $R$ from the pivoted QR. Another
is to apply an early stopping once the important column index set do
not change for two consecutive iterations. For the numerical results
reported in this paper, we used $q=2$. Note that any linear sampling
algorithm can substitute \autoref{alg:randomized_sampling}, and in
practice, \autoref{alg:randomized_sampling} returns columns whose
volume is very close to the largest.

\begin{algorithm}[htbp]
    \Fn{Randomized\_Sampling($M_{i,:}$, $r_i$, $q$)}{
        \SetKwInOut{Input}{input}
        \SetKwInOut{Output}{output}

        \Input{(1) Row-submatrix $M_{i,:}$ to sample from in its implicit form
        (given data and kernel function); (2) Sample size $r_i$; (3)
        Iteration parameter $q$}

        \Output{Important column index set $\Pi_c$ for $M_{i,:}$}

        $\Pi_r = \emptyset$\\

        \For{iter=1, \dots, $q$}{
            {\textit{Important columns}}. Uniformly sample $r_i$ rows, and
            denote the index set as $\Gamma_r$, update $\Pi_r =
            \Pi_r \cup \Gamma_r$.  Apply a pivoted QR factorization
            on $M_{i,:}(\Pi_r, :)$ to get the important columns index set,
            denoted as $\Pi_c$.\\

            {\textit{Important rows}}. Uniformly sample $r_i$ columns, and
            denote the index set as $\Gamma_c$. Update $\Pi_c =
            \Gamma_c \cup \Pi_c$.  Apply a pivoted LQ factorization
            on $M_{i,:}(:, \Pi_c)$ to get the important row index set,
            denoted as $\Pi_r$.\\
        }
        
        \Return $\Pi_c$
    }
    \emph{Note: The pivoted QR is the QR factorization with column
    pivoting based on the largest column norm.}
    
    \caption{Randomized sampling algorithm for each submatrix}
    
    \label{alg:randomized_sampling}
\end{algorithm}

\begin{algorithm}[htbp]
    \Fn{BBF\_Sampling($\{M_{i,:}\}_{i=1}^k$,
    $\{r_i\}_{i=1}^k$, $q$)}{
        \SetKwInOut{Input}{input}
        \SetKwInOut{Output}{output}

        \Input{(1) Sub-matrices $\{M_{i,:}\}_{i=1}^k$ to sample from
        in their implicit forms (given data and kernel function); (2)
        Sample sizes $\{r_i\}_{i=1}^k$ for each sub-matrix $M_{i,:}$;
        (3) Iteration parameter $q$}

        \Output{Important column index set $\Pi_i$ for each
        row-submatrix}

        \For{$i = 1, \ldots, k$}{
            $\Pi_i$ = Randomized\_Sampling($M_{i, :}(:, \Gamma), r_i,
            q)$ (using \autoref{alg:randomized_sampling})
        }
      
        \Return $\Pi_i$ for $i = 1, \ldots, k$ 
    }
    \caption{BBF sampling algorithm}
    \label{alg:sample_BBF}
\end{algorithm}

Applying \autoref{alg:randomized_sampling} to $k$ submatrices
$\{M_{i,:}\}_{i=1}^k$ will return the desired $k$ sets of important
columns for BBF, which is described in \autoref{alg:sample_BBF}. The
complexity of \autoref{alg:sample_BBF} depends on $k$ (see
\autoref{sec:complexity_analysis} for details), and we can remove
this dependence by applying \autoref{alg:randomized_sampling}
on a pre-selected and refined set of columns instead of all the
columns. This leads to a more efficient procedure to sample columns
for the $k$ submatrices as described in \autoref{alg: more_fast}. Our
final BBF construction algorithm will use \autoref{alg:sample_BBF}
for column sampling.

\begin{algorithm}[htbp]
    \Fn{BBF\_Sampling\_I($\{M_{i,:}\}_{i=1}^k$,
    $\{r_i\}_{i=1}^k$, $q$)}{
        \SetKwInOut{Input}{input}
        \SetKwInOut{Output}{output}

        \Input{(1) Sub-matrices $\{M_{i,:}\}_{i=1}^k$ to sample from
        in their implicit forms (given data and kernel function); (2)
        Sample sizes $\{r_i\}_{i=1}^k$ for each sub-matrix $M_{i,:}$;
        (3) Iteration parameter $q$}

        \Output{Important column index set $\Pi_i$ for each
        row-submatrix}

        \For{$i = 1, \ldots, k$}{
            Randomly sample $r_i$ columns from $M_i$ and denote the
            index set as $\Pi_i$\\

            Apply a pivot LQ on $M_{i, :}(:, \Pi_i)$ to obtain $r$
            important rows, and we denote the index set as $\Gamma_i$
        }

        Stack all the sampled rows $\Gamma = [\Gamma_1, \ldots,
        \Gamma_k]$\\

        \For{$i = 1, \ldots, k$}{
            $\Pi_i$ = Randomized\_Sampling($M_{i, :}(:, \Gamma), r_i,
            q)$ (using \autoref{alg:randomized_sampling})
        }
      
        \Return $\Pi_i$ for $i = 1, \ldots, k$ 
    }
    \caption{More efficient BBF sampling algorithm}
    \label{alg: more_fast}
\end{algorithm}

{\emph{Part 2: Orthogonalization algorithm}}

Having sampled the representative columns $M_{i,:}(:,\Pi_i)$,
the next step is to obtain the column basis \added{R2}{3}{that approximates the span of the selected columns}. This can be achieved
through any orthogonalization methods, \emph{e.g.}, pivoted QR,
SVD, randomized SVD \cite{halko2011finding}, etc. According to
\autoref{alg:randomized_sampling}, the size of the sampled index
set $\Pi_c$ can be as large as $qr$. In practice, we found that the
randomized SVD works efficiently. The randomized SVD algorithm was
proposed to reduce the cost of computing a rank-$r$ approximation
of an $m \times n$ matrix to $O(mnr)$. The algorithm is described
in \autoref{alg:randomized_svd}. The practical implementation of
\autoref{alg:randomized_svd} involves an oversampling parameter $\ell$
to reduce the iteration parameter $q$. For simplicity, we eliminate
$\ell$ from the pseudo code.
 
\begin{algorithm}[htbp]
    \Fn{ Randomized\_SVD($M$, $r$, $q$)}{
        \SetKwInOut{Input}{input}
        \SetKwInOut{Output}{output}
        
        \Input{(1) Matrix $M \in \mathbb{R}^{m \times n}$; (2)
        desired rank $r$; (3) iteration parameter $q$}

        \Output{$U$, $\Sigma$, and $V$ such that $M \approx U\Sigma
        V^\top$}
        
        Randomly generate a Gaussian matrix $\Omega \in \mathbb{R}^{n
        \times r}$\\

        $M \Omega$ = $QR$\\
        
        \For{$i$=1, \dots, $q$}{
            $M^\top Q = \hat Q \hat R$ \\
            $M\hat Q = QR$
        }
      
        $\hat U \Sigma V^\top = Q^\top M$\\
      
        $U = Q\hat U$
      
        \Return $U$, $\Sigma$, $V$
    }
    \caption{Randomized SVD}
    \label{alg:randomized_svd}
\end{algorithm}
   
{\bf 2. Inner matrix construction}

We then consider the inner matrix construction. Given column base
$U_i$ and $U_j$, we seek a matrix $C_{i,j}$ such that it minimizes
$$\|M_{i,j} - U_iC_{i,j} U_j^\top \|.$$ The minimizer is given by
$C_{i,j} = U_i^{\dagger} M_{i,j}(U_j^\top)^{\dagger}$. Computing
$C_{i,j}$ exactly has a quadratic cost. Again, we restrict ourselves
in a subspace and propose a sampling-based approach that is efficient
yet accurate.  The following proposition provides a key theoretical
insight behind our algorithm.
\begin{proposition} \label{prop:inner_matrix}
    If a matrix $M \in \mathbb{R}^{m \times n}$ can be written
    as $M = U C V^\top$, where $U \in \mathbb{R}^{m \times r}$ and
    $V \in \mathbb{R}^{n \times r}$.  Further, if for some index
    set $\mathcal{I}$ and $\mathcal{J}$, $U(\mathcal{I},:)$ and
    $V(\mathcal{J},:)$ are full rank, then, the inner matrix $C$
    is given by
    \begin{equation} \label{eq:inner_matrix}
        C = (U(\mathcal{I},:))^{\dagger} \; M(\mathcal{I}, \mathcal{J})
        \;(V(\mathcal{J},:)^\top)^{\dagger},
    \end{equation}
    where $\dagger$ denotes the pseudo-inverse of the matrix.
\end{proposition}
        
\begin{proof}
    To simplify the notations, we denote $\widehat U
    = U(\mathcal{I},:)$, $\widehat V = V(\mathcal{J},:)$, and
    $\widehat M = M(\mathcal{I}, \mathcal{J})$, where $\mathcal{I}$
    is the sampled row index set for $U$ and $\mathcal{J}$ is the
    sampled row index set for $V$. We apply the sampling matrices
    $P_{\mathcal{I}}$ and $P_{\mathcal{J}}$ (matrices of 0's and 1's)
    to both sides of equation $M = UCV^\top$, and obtain
    \begin{equation*}
        P_{\mathcal{I}} M P_{\mathcal{J}}^\top = P_{\mathcal{I}} UC
        V^\top P_{\mathcal{J}}^\top,
    \end{equation*}
    \emph{i.e.}, 
    \begin{equation*}
        \widehat M = \widehat U C \widehat V^\top.
    \end{equation*}
    The assumption that $\widehat U$ and $\widehat V$ are tall and
    skinny matrices with full column ranks implies that $\widehat
    U^\dagger \hat U= I$ and $\widehat V^\top (\widehat V^\top)^\dagger
    = I$. We then multiply $\widehat U^{\dagger}$ and $ (\widehat
    V^\top)^{\dagger}$ on both sides and obtain the desired result:
    \begin{equation*}
        \widehat U^{\dagger} \widehat M (\widehat V^\top)^\dagger =
        \widehat U^\dagger \widehat UC \widehat V^\top (\widehat
        V^\top )^\dagger = C.
    \end{equation*}
\end{proof}
    
Prop.~\ref{prop:inner_matrix} provides insights into an efficient,
stable and accurate construction of the inner matrix. In practice,
the equality \added{R1}{11}{$M=UCV^\top$} in Prop.~\ref{prop:inner_matrix} often holds with an
error term and we seek index sets $\mathcal{I}$ and $\mathcal{J}$
such that the computation for $C$ is accurate and numerically
stable. \autoref{eq:inner_matrix} suggests that a good choice leads
to an $M(\mathcal{I}, \mathcal{J})$ with a large volume. However,
finding such a set can be computationally expensive and a heuristic is
required for efficiency. We used a simplified approach where we sample
$\mathcal{I}$ (resp.\ $\mathcal{J}$) such that $U(\mathcal{I},:)$
(resp.\ $V(\mathcal{J},:)$) has a large volume. This leads to good
numerical stability, because having a large volume is equivalent to
being nearly orthogonal, which implies a good condition number. In
principle, a pivoted QR strategy could be used but, fortunately,
we are able to skip it by using the results from the basis
construction. Recall that in the basis construction, the important
rows were sampled using a pivoted LQ factorization, hence, they
already have large volumes.
    
Therefore, the inner matrix construction is described in
what follows.  We first uniformly sample $r$ column indices
$\Gamma_j$ and $r$ row indices $\Gamma_i$, respectively, from
$\mathcal{C}_j$ and $\mathcal{C}_i$.  Then, the index sets are
constructed as $\mathcal{I} = \Pi_i\cup\Gamma_i$ and $\mathcal{J}
= \Pi_j\cup\Gamma_j$, where $\Pi_i$ and $\Pi_j$ are the important
row index sets from the basis construction. Finally, $C_{i,j}$ is
given by $$(U_i(\mathcal{I},:))^{\dagger} \; M_{i,j}(\mathcal{I},
\mathcal{J}) \;(U_j(\mathcal{J},:)^\top)^{\dagger}.$$
    
We also observed small entries in some off-diagonal blocks of the inner
matrix. Those blocks normally represent far-range interactions. We can
set the blocks for which the norm is below a preset threshold to 0. In
this way, the dense inner matrix becomes a block-wise sparse matrix,
further reducing the memory.
    
Having discussed the details for the construction algorithm, we
summarize the procedure in \autoref{alg: fast}, which is the algorithm
used for all the numerical results.

\begin{algorithm}[htbp]
    \Fn{BBF\_Construction ($k$, $\{\mathcal{C}_i\}_{i=1}^k$,
    $\{r_i\}_{i=1}^k$, $M$, $q_\text{Samp}$, $q_\text{SVD}$)}{
        \SetKwInOut{Input}{Input}
        \SetKwInOut{Output}{Output}

        \Input{(1) Number of clusters $k$; (2) Clustering assignments
        $\{\mathcal{C}_i\}_{i=1}^k$; (3) Rank $\{r_i\}_{i=1}^k$ for
        each column basis; (4) Matrix $M$ in its implicit form; (5)
        Iteration parameter $q_\text{Samp}$ for randomized sampling;
        (6) Iteration parameter $q_\text{SVD}$ for randomized SVD.}

        \Output{Block diagonal matrix $\widetilde{U}$ and block-wise
        sparse matrix $\widetilde{C}$ s.t. $M \approx \widetilde{U}
        \widetilde{C} \widetilde{U}^\top$}

        $[\Pi_1, \cdots, \Pi_k]$ = BBF\_Sampling($\{M_{i,:}\}_{i=1}^k$,
    $\{r_i\}_{i=1}^k$,
        $q_\text{Samp}$) (\autoref{alg:sample_BBF})\\

        \For{$i = 1, \ldots, k$}{
            $U_i =$ Randomized\_SVD($M_{i:}(:, \Pi_i)$, $r_i$,
            $q_\text{SVD}$) (\autoref{alg:randomized_svd})
        }

        \For{$i = 1, \ldots, k$}{
            \For{$j = 1, \ldots, i$} {
                \eIf{cutoff criterion is not satisfied}{
                    Uniformly sample $\Gamma_i$ and $\Gamma_j$ from
                    $\mathcal{C}_i$ and $\mathcal{C}_j$, respectively\\

                    $\mathcal{I} = \Pi_i \cup \Gamma_i$
                    and $\mathcal{J} = \Pi_j \cup \Gamma_j$\\

                    $C_{i,j} = (U_i(\mathcal{I},:))^\dagger
                    M_{i,j}(\mathcal{I},\mathcal{J})
                    (U_j(\mathcal{J},:)^\top)^\dagger$\\

                    $C_{j,i} = C_{i,j}^\top$
                }{
                    $C_{i,j} = 0$ \\
                    $C_{j,i} = 0$
                }
            }
        }
        
        \Return $\tilde{U}$, $\tilde{C}$
    }
    \caption{{\bf Main Algorithm} --- Fast construction algorithm for BBF}
    \label{alg: fast}
\end{algorithm}

In this section, for simplicity, we only present BBF for symmetric
kernel matrices. However, the extension to general non-symmetric cases
is straightforward by applying similar ideas, and the computational
cost will be roughly doubled. Asymmetric BBF can be useful in
compressing the kernel matrix in the testing phase.

{\bf 3. Pre-computation: Parameter Selection}

We present a heuristic algorithm to identify input parameters for BBF.
The algorithm takes $n$ input points $\{\vecx_i\}_{i=1}^n$ and a
requested error (tolerance) $\epsilon$, and outputs the suggested
parameters for the BBF construction algorithm, specifically, the
number of clusters $k$, the index set for each cluster $\mathcal{I}$,
and the estimated rank $r_i$ for the submatrix corresponding to the
cluster $\mathcal{I}$.  We seek a set of parameters that minimizes
the memory cost while keeping the approximation error below $\epsilon$.

\textbf{Choice of column ranks.} Given the tolerance $\epsilon$ and
the number of clusters $k$, we describe our method of identifying the
column ranks. To maintain a low cost, the key idea is to consider only
the diagonal blocks instead of the entire row-submatrices. For each
row-submatrix in the RBF kernel matrices (after permutation), the
diagonal block, which represents the interactions within a cluster,
usually has a slower spectral decay than that of off-diagonal blocks
which represent the interactions between clusters. Hence, we minimize
the input rank for the diagonal block and use this as the rank for
those off-diagonal blocks in the same row.
    
Specifically, we denote $\sigma_{1,i} \ge \sigma_{2,i} \ge \ldots \ge
\sigma_{n_i, i}$ as the singular values for $M_{i,i}$. Then for block
$M_{i,i} \in \mathbb{R}^{n_i \times n_i}$, the rank $r_i$ is chosen as
\begin{equation*}
    r_i = \min \Big\{m~ \mid~ \sum_{p = m+1}^{n_i} \sigma_{p,i}^2 <
    \frac{n_i^2}{n^2} \|M_{i,i}\|_F^2~ \epsilon^2\Big\}.
\end{equation*}

\textbf{Choice of number of clusters $\bm k$.}
Given the tolerance $\epsilon$, we consider the number
of clusters $k$. For $k$ clusters, the upper bound on the memory usage
of BBF is $\sum_{i = 1}^k n_i r_i + \left(\sum_{i=1}^k r_i\right)^2$,
where $r_i$ is computed as described above. Hence, the optimal $k$
is the solution to the following optimization problem
\begin{align*}
    & \text{minimize}_k\;\; g(k) = \sum_{i = 1}^k n_i r_i +
    \left(\sum_{i=1}^k r_i\right)^2 \\
    & \text{subj to}\;\;  r_i =\min \Big \{m~|~
    \sum_{p = m+1}^{n_i} \sigma_{p}^2 < \frac{n_i^2}{n^2}
    \|M_{i,i}\|_F^2~ \epsilon^2 \Big\}, \;\;
    \forall i
\end{align*}
We observed empirically that in general, $g(k)$ is close to convex
in the interval $[1, O(\sqrt{n})]$, which enables us to perform
a dichotomy search algorithm with complexity $O(\log n)$ for the
minimal point.

\subsubsection{Complexity Analysis}
\label{sec:complexity_analysis}

In this section, we analyze the algorithm complexity. We will provide
detailed analysis on the factorization step including the basis
construction and the inner matrix construction, and skip the analysis
for the pre-computation step. We first introduce some notations.

\textit{Notations.} Let $k$ denote the number of clusters,
$\{n_i\}_{i=1}^k$ denote the number of point in each cluster,
$\{r_i\}_{i=1}^k$ denote the requested rank for the blocks in the
$i$-th submatrix, and $l$ denote the oversampling parameter.

{\bf Basis construction.} The cost comes from two parts, the column
sampling and the randomized SVD. We first calculate the cost for
the $i$-th row-submatrix. For the column sampling, the cost is
$n_i (r_i + l)^2 + n(r_i+l)^2$, where the first term comes from
the pivoted $LQ$ factorization, and the second term comes from
the pivoted $QR$ factorization. For the randomized SVD, the cost
is $n_i(r_i+l)^2$. Summing up the costs from all the submatrices,
we obtain the overall complexity
\begin{equation*}
    O\left(\sum_{i = 1}^kn_i (r_i + l)^2 + n(r_i+l)^2 + n_i(r_i+l)^2
    \right).
\end{equation*}
We simplify the result by \replaced{R1}{12}{assuming that the clusters are of equal
size and the numerical rank for each block is $r$}{denoting the maximum numerical rank of all blocks as $r$}. Then, the above
complexity is simplified to $O(nkr^2)$.
    
{\bf Inner matrix construction.} The cost for computing inner matrix
$C_{i,j}$ with sampled $M_{i,j}$, $U_i$ and $U_j$ is $r_i^2r_j + r_i
r_j^2$. Summing over all the $k^2$ blocks, the overall complexity is
given by
\begin{equation*}
    \sum_{i = 1}^k \sum_{j = 1}^k r_i^2r_j + r_i r_j^2.
\end{equation*}
With the same assumptions as above, the simplified complexity is
$O(k^2r^3)$. Note that $k$ can reach up to $O(\sqrt{n})$ while still
maintaining a linear complexity for this step.
    
Finally, we summarize the complexity of our algorithm in
\autoref{tab:complexity}.

\begin{table}[htbp]
    \centering

    \caption{Complexity table. $n$ is the total number of points,
    $k$ is the number of clusters, $r$ is the numerical rank for
    column basis.}
    \label{tab:complexity}
    \vspace{.4cm}
    \begin{tabular}{ccccc}
        \toprule
        \multicolumn{2}{c}{Pre-computation} &
        \multicolumn{2}{c}{Factorization} &
        Application\\
        \toprule
        Each block & $O(n_i r^2)$ &
        Basis      & $O(nkr^2)$ &
        \multirow{3}{*}{$O(nr + (rk)^2)$} \\
        \cmidrule(r){1-2} \cmidrule(r){3-4}
        Compute $g(k)$ & $O(nr^2)$ &
        Inner matrix   & $O(k^2r^3)$ & \\
        \cmidrule(r){1-2} \cmidrule(r){3-4}
        Total          & $O(r^2n \log n)$ &
        Total          & $O(nk r^2 + k^2r^3)$& \\
        \bottomrule
    \end{tabular}
\end{table}
    
From \autoref{tab:complexity}, we note that the factorization and
application cost (storage) depend quadratically on the number of
clusters $k$. This suggests that a large $k$ will spoil the linearity
of the algorithm. However, this may not be the case for most machine
learning kernels, and we will discuss the influence of $k$ on three
types of kernel matrices: 1) well-approximated by a low-rank matrix;
2) full-rank but approximately sparse; 3) full-rank and dense.

\begin{enumerate}[leftmargin=*]
    \item {\emph{Well-approximated by a low-rank matrix}}. When the
    kernel matrix is well approximated by a low-rank matrix, $kr$ is
    up bounded by a constant (up to the approximation accuracy). In
    this case, both the factorization and application costs are linear.
    
    \item {\emph{Full-rank but approximately sparse}}. When the kernel
    matrix is full-rank ($kr = O(n)$) but approximately sparse, the
    application cost (storage) remains linear due to the sparsity. By
    ``sparsity'', we mean that as $h$ decreases, the entries in the
    off-diagonal blocks of the inner matrices become small enough
    that setting them to 0 does not cause much accuracy loss. The
    factorization cost, however, becomes quadratic when using
    \autoref{alg:sample_BBF}. One solution is to use \autoref{alg:
    more_fast} for column sampling, which removes the dependence on
    $k$, assuming $k < O(\sqrt{n})$.
    
    \item {\emph{Full-rank and dense}}. In this case, BBF would be
    sub-optimal. However, we experimentally observed that \replaced{R1}{5}{most}{many} kernel
    matrices generated by {RBF} functions with high dimensional data
    are in case 1 or 2.
\end{enumerate}
    
\begin{figure}[htbp]
    \centering
    \begin{subfigure}[b]{0.45\textwidth}
        \includegraphics[width=\textwidth]{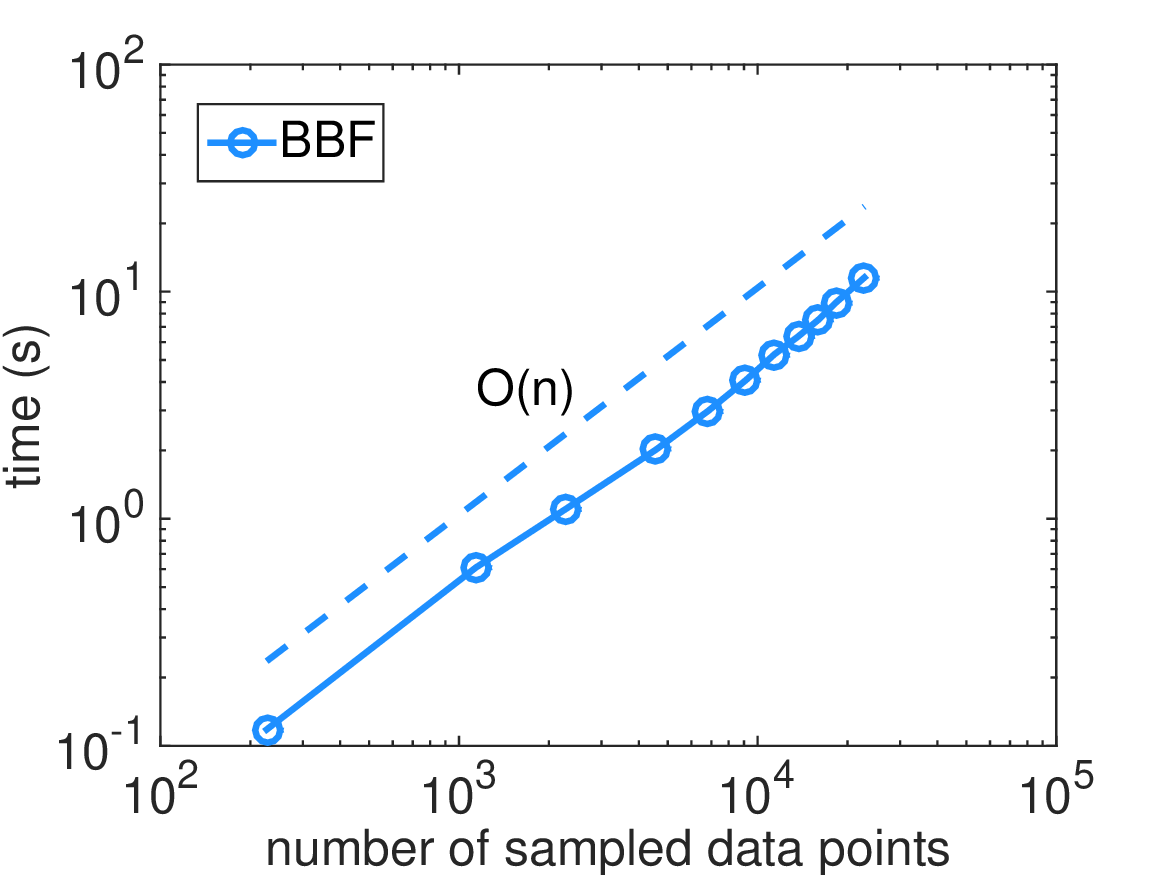}
        \caption{Census housing, $h = 2$}
    \end{subfigure}
    \begin{subfigure}[b]{0.45\textwidth}
        \includegraphics[width=\textwidth]{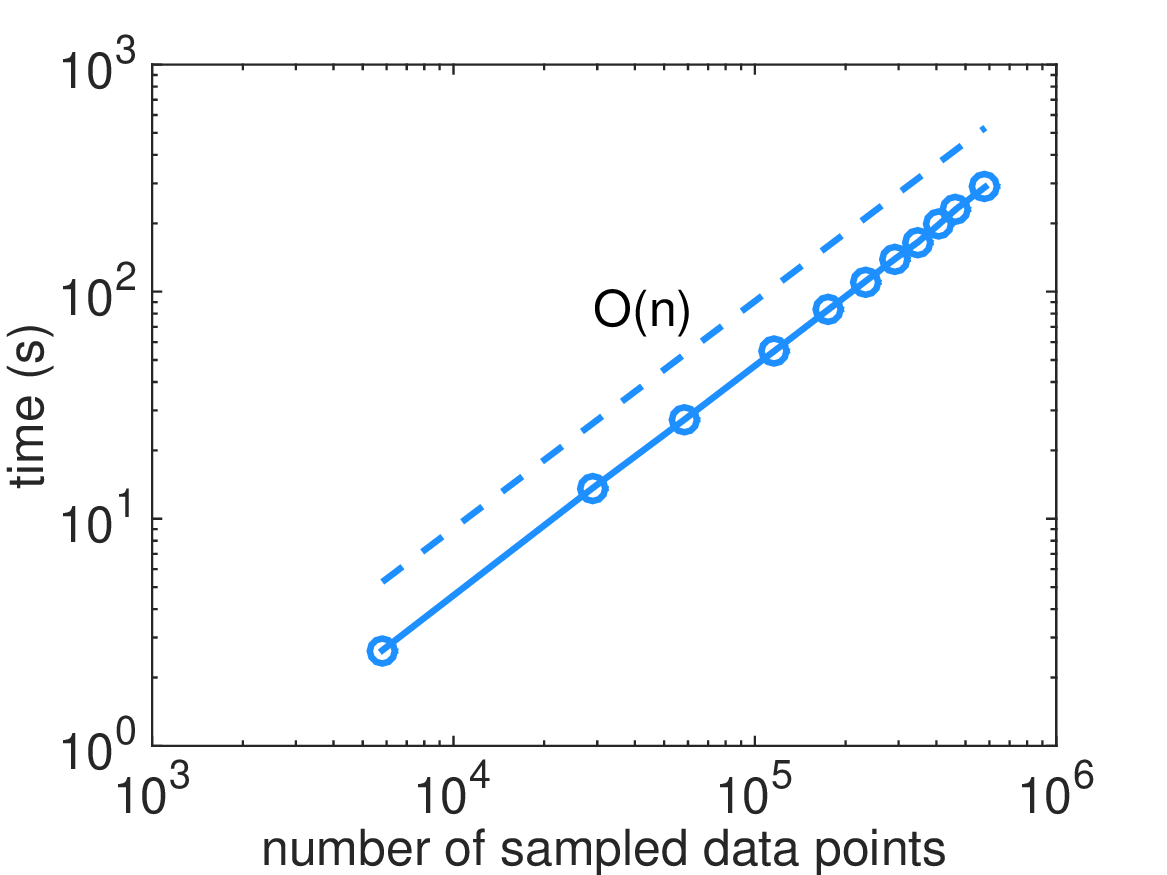}
        \caption{Forest covertype, $h = 2$}
    \end{subfigure}

    \caption{Factorization time (loglog scale) for kernel matrices from
    real datasets. To illustrate the linear growth of the complexity,
    we generated datasets with a varying number of points with the
    following strategy. We first clustered the data into 15 groups,
    and sampled a portion $p\%$ from each group, then increased $p$. To
    avoid the influence from other factors on the timing, we fixed
    the input rank for each block. As we can see, the timing grows
    linearly with the data size (matrix size).}
    \label{fig:linear}
\end{figure}

In the end, we empirically verify the linear complexity of our
method. \autoref{fig:linear} shows the factorization time (in second)
versus the number of data points on some real datasets. The trend is
linear, confirming the linear complexity of our algorithm.

\section{Experimental Results}
\label{sec:results}

In this section, we experimentally verify the advantages of the BBF
structure in \autoref{sec:advantage_bbf} and the BBF algorithm in
\autoref{sec:advantage_algo}. By BBF algorithm we refer to the BBF
structure and the proposed fast construction algorithm.

The datasets are listed in \autoref{tab:datasets} and
\autoref{tab:datasets_classifcation}, and they can be downloaded from the
UCI repository \cite{Bache+Lichman:2013}, the libsvm website \cite{CC01a}
and Kaggle. All the data were normalized such that each dimension has
mean 0 and standard deviation 1. All the experiments were performed on
a computer with 2.4 GHz CPU and 8 GB memory.
    
\begin{table}[htbp]
    \centering
    \caption{Real datasets used in the experiments.}
    \vspace{.4cm}
        \begin{tabular}{cccc}
        \toprule
        \textbf{Dataset} & Abalone  &Mushroom& Cpusmall  \\
        \midrule
        \textbf{\# Instance} & 4,177 &8,124&  8,192  \\
        \textbf{\# Attributes} & 8 &112& 16  \\
        \midrule
        \textbf{Dataset} & Pendigits & Census house  &Forest covertype   \\
        \midrule
        \textbf{\# Instance} & 10,992 & 22,748 &581,012 \\
        \textbf{\# Attributes} & 11 & 16 &54 \\
        \bottomrule
        \end{tabular}
    \label{tab:datasets}
\end{table}

\subsection{{BBF} structure}
\label{sec:advantage_bbf}

In this section, we will experimentally analyze the key factors in
our {BBF} structure that contribute to its advantages over competing
methods.  Many factors contribute and we will focus our discussions
on the following two: 1) The {BBF} structure has its column base
constructed from the entire row-submatrix, which is an inherently more
accurate representation than from diagonal blocks only (see MEKA);
2) The {BBF} structure considers local interactions instead of only
global interactions used by a low-rank scheme.

\subsubsection{Basis from the row-submatrix versus diagonal blocks} 
\label{sec:diag_entirerow}

\begin{figure}[htbp]
    \centering
    \renewcommand{\thesubfigure}{a1}
    \begin{subfigure}[b]{0.38\textwidth}
        \includegraphics[width=\textwidth]{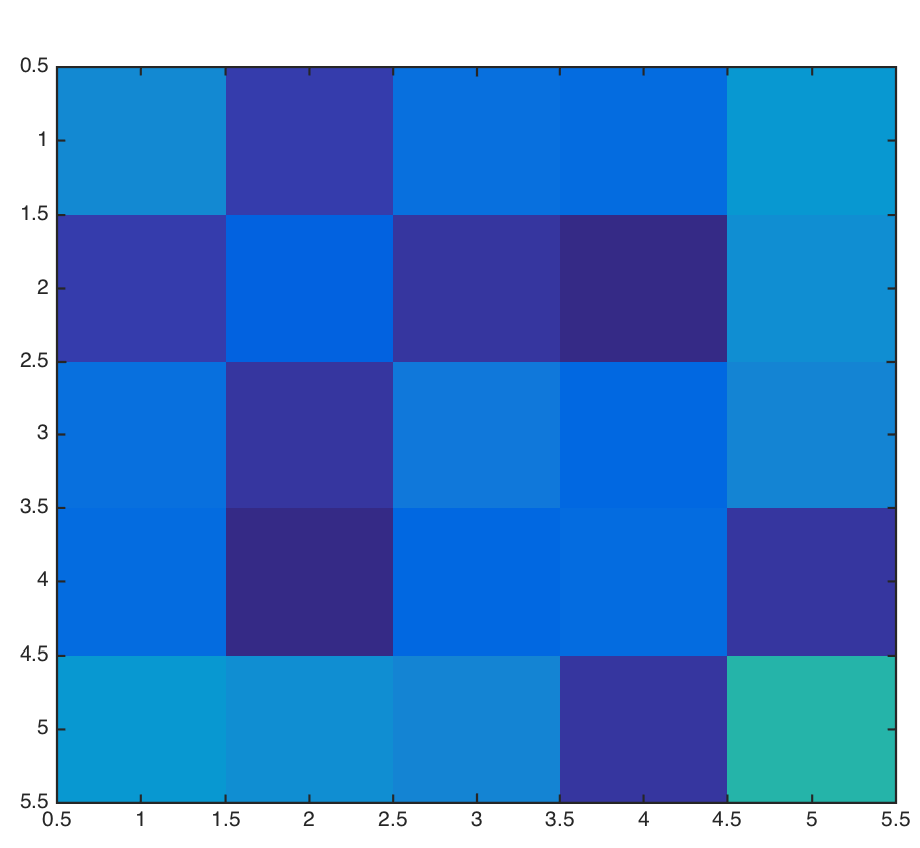}
        \caption{$M_{i,:}$, {SVD}}
    \end{subfigure}
    \renewcommand{\thesubfigure}{a2}
    \begin{subfigure}[b]{0.45\textwidth}
        \includegraphics[width=\textwidth]{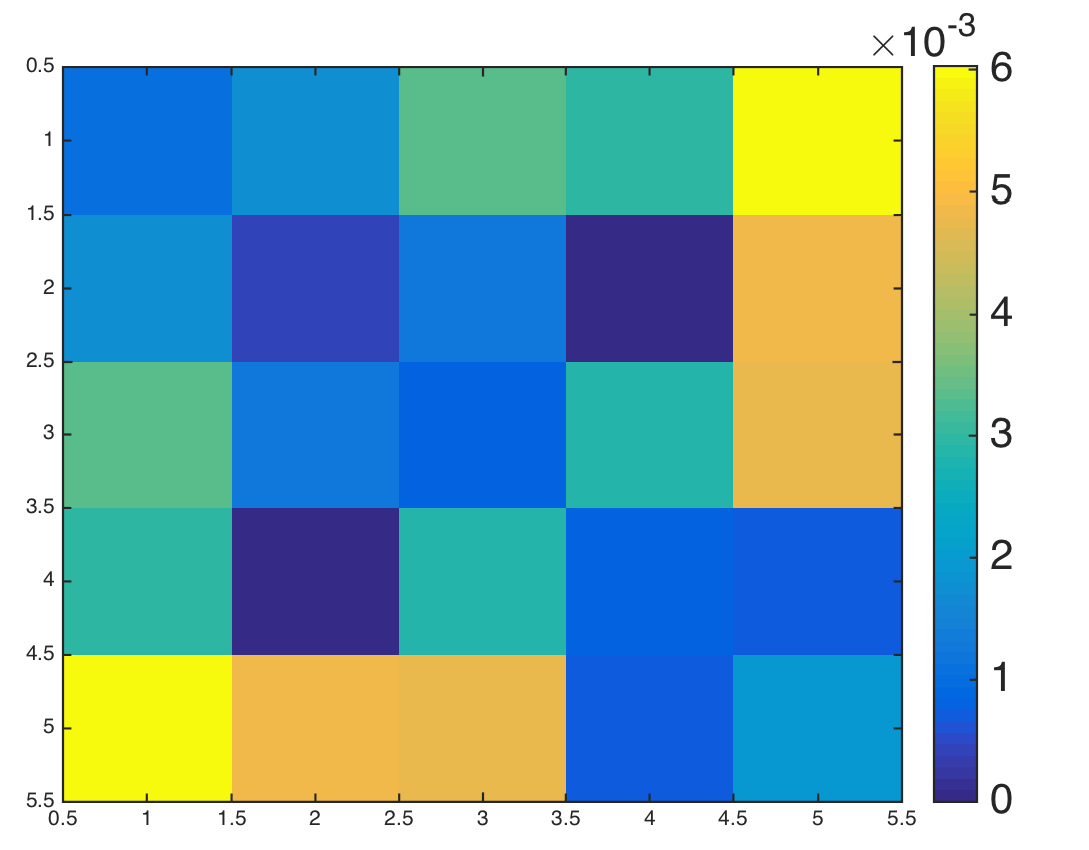}
        \caption{$M_{i,i}$, {SVD}}
    \end{subfigure}
    \renewcommand{\thesubfigure}{b1}
    \begin{subfigure}[b]{0.37\textwidth}
        \includegraphics[width=\textwidth]{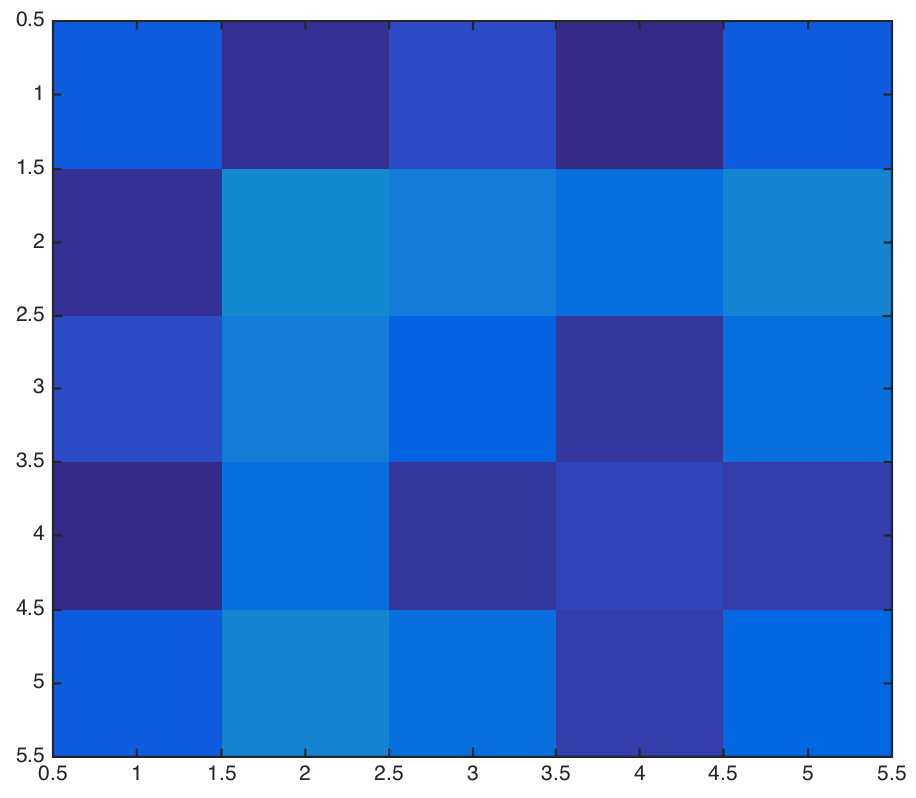}
        \caption{$M_{i,:}$, randomized}
    \end{subfigure}
    \renewcommand{\thesubfigure}{b2}
    \begin{subfigure}[b]{0.45\textwidth}
        \includegraphics[width=\textwidth]{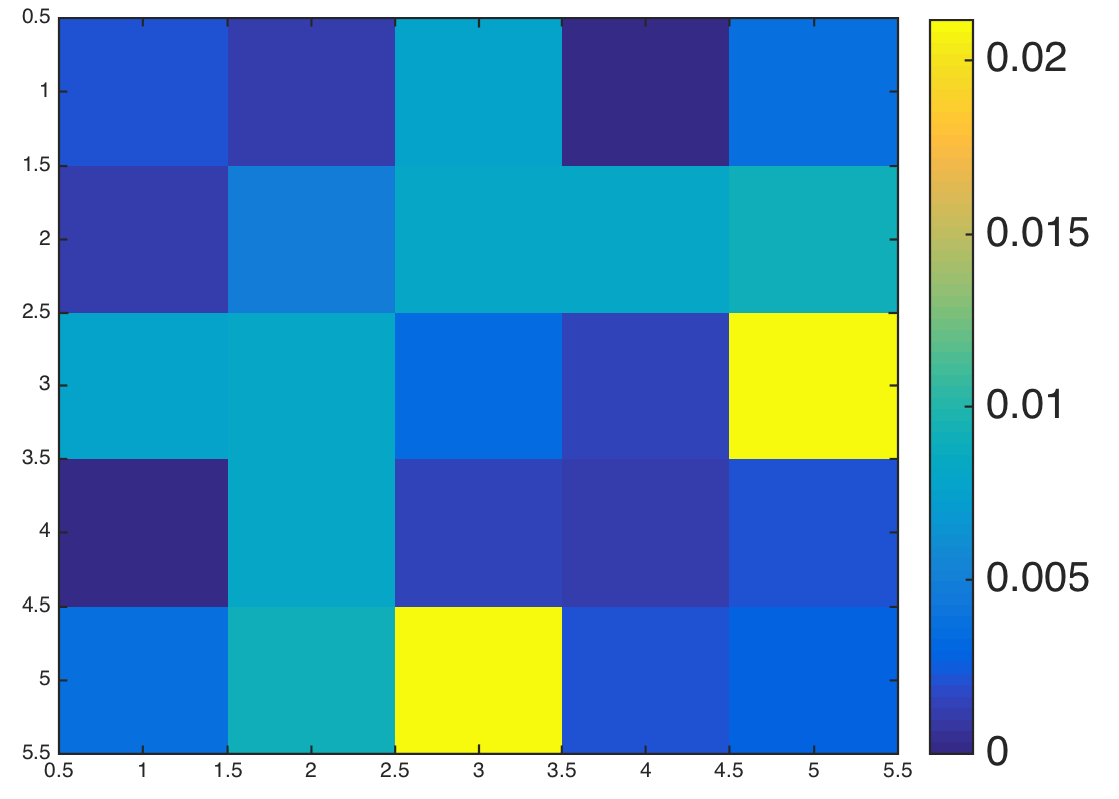}
        \caption{$M_{i,i}$, randomized}
    \end{subfigure}

    \caption{Errors for each block in the approximated matrix from
    the Abalone dataset. Warmer color represents larger error. Subplot
    (a1) and (a2) shares the same colorbar; and (b1) and (b2) shares
    the same colorbar.  The error for block $(i,j)$ is computed
    as $\|M_{i,j} - \widehat M_{i,j}\|_F / \|M\|_F$, where $\widehat M_{i,j}$ is the approximation of $M_{i,j}$. The basis in
    subplot (a1) and (a2) are computed by an {SVD}; and in (b1) and
    (b2) are computed by the randomized sampling algorithm.  As we
    can see, computing the column basis from the diagonal blocks
    leads to lower error in the diagonal blocks; however, the errors
    in the off-diagonal blocks are much larger. The relative error
    in subplot (a1), (a2), (b1) and (b2) are $1.4 \times 10^{-3}$,
    $6.3 \times 10^{-3}$, $1.5 \times 10^{-2}$, $4.0 \times 10^{-2}$,
    respectively.}
    \label{fig:diag}
\end{figure}

We verify that computing the column basis from the entire row-submatrix
$M_{i,:}$ is generally more accurate than from the diagonal blocks
$M_{i,i}$ only. Column basis computed from the diagonal blocks
only preserves the column space information in the diagonal blocks,
and will be less accurate in approximating the off-diagonal blocks.
\autoref{fig:diag} shows that computing the basis from the entire
row-submatrix is more accurate.

\subsubsection{BBF structure versus low-rank structure}

We compare the {BBF} structure and the low-rank structure. The {BBF}
structure refers to \autoref{fig:bbf}, and the low-rank structure
means $K \approx UU^\top$, where $U$ is a tall and skinny matrix.
For a fair comparison, we fixed all the factors to be the same except
for the structure. For example, for both the BBF and the low-rank
schemes, we used the same sampling method for the column selection,
and computed the inner matrices exactly to avoid randomness introduced
in that step. The columns for BBF and low-rank scheme, respectively,
were sampled from each row-submatrix $M_{i,:} \in R^{n_i \times n}$ and
the entire matrix $M \in R^{n \times n}$. For BBF with leverage-score
sampling, we sampled columns of $M_{i,:}$ based on its column leverage
scores computed from the algorithm in \cite{drineas2012fast}.

\begin{figure}[htbp]
    \centering
    \begin{subfigure}[b]{0.45\textwidth}
        \includegraphics[width=\textwidth]{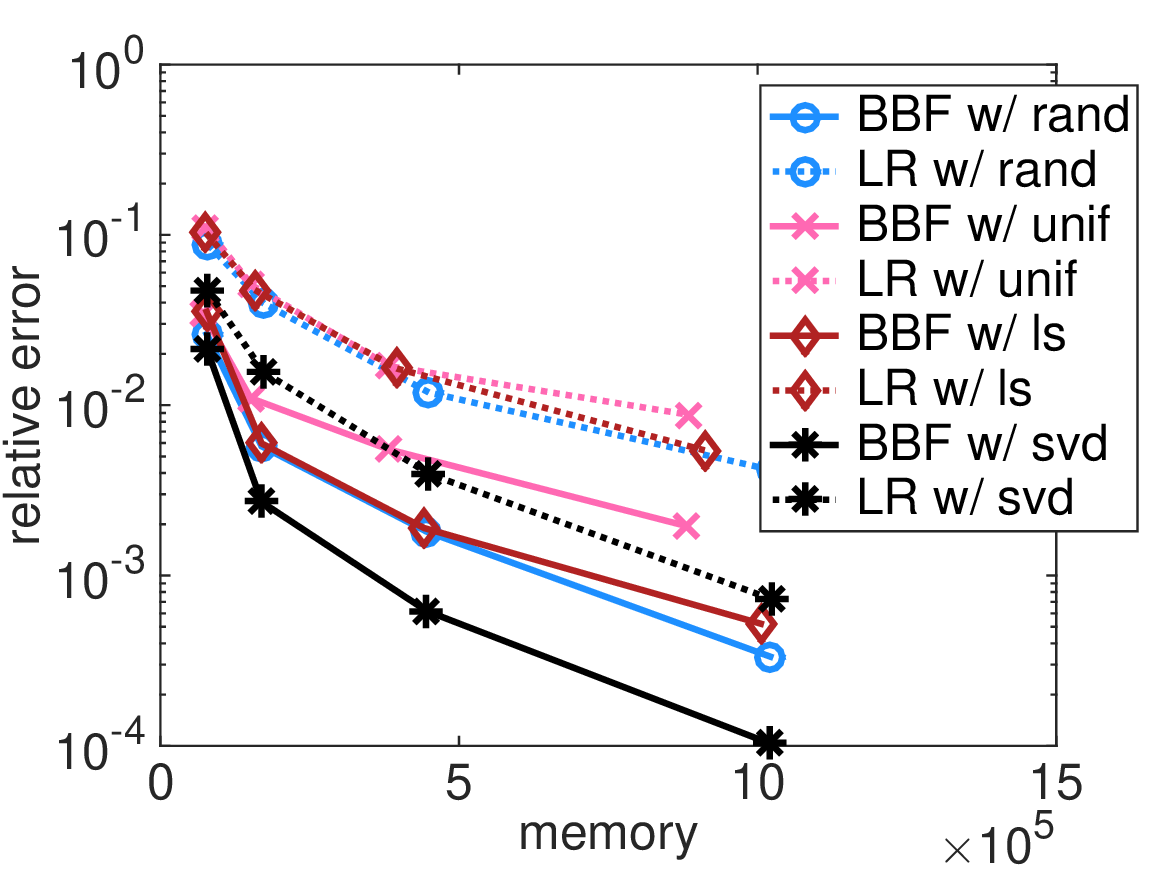}
        \caption{Abalone, $h$=2}
    \end{subfigure}
    \begin{subfigure}[b]{0.45\textwidth}
        \includegraphics[width=\textwidth]{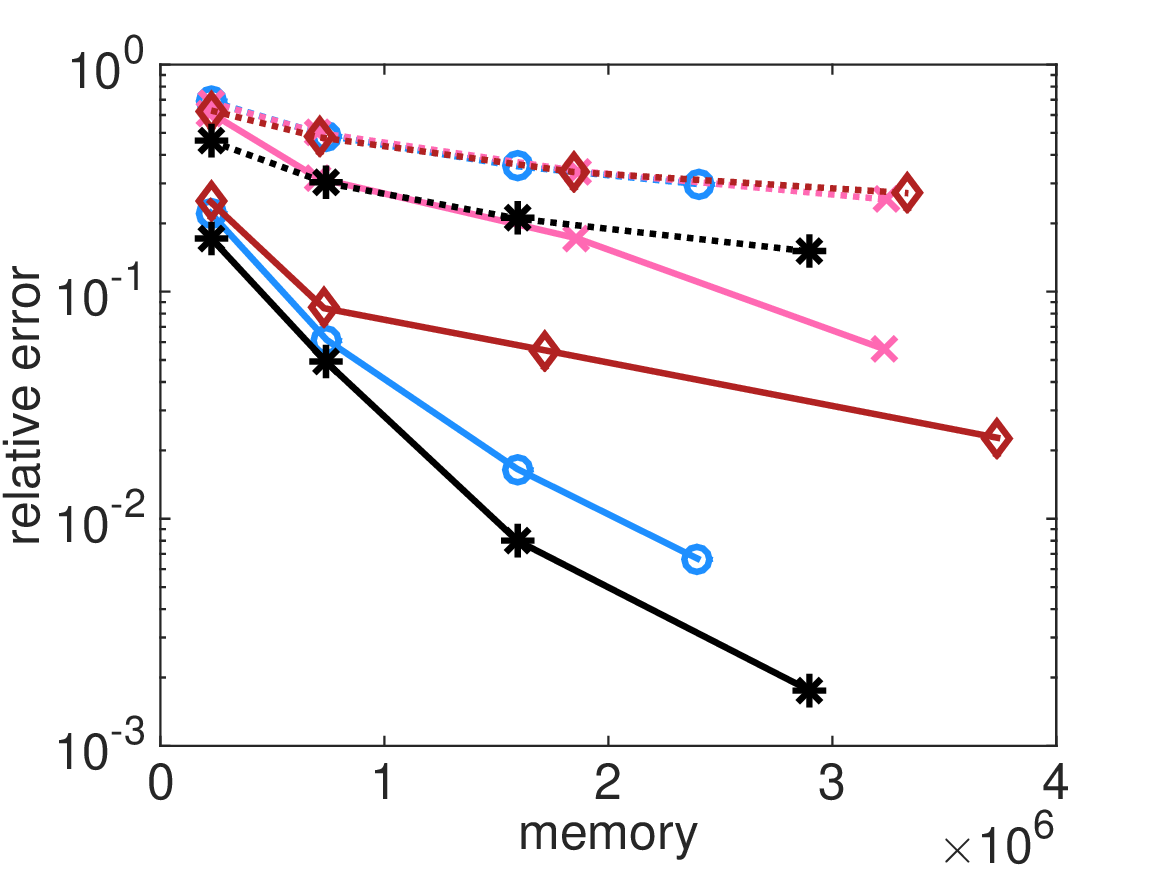}
        \caption{Pendigits, $h$ = 2}
    \end{subfigure}
    
    \caption{Kernel approximation error versus memory cost for {BBF}
    and low-rank structure with different sampling methods. Gaussian
    kernel is used. The results are averaged over 5 runs.  {BBF} (solid
    lines) uses the structure described in \autoref{fig:bbf}, and {LR}
    (dash lines) uses a low-rank structure. ``rand'': randomized
    sampling; ``unif'':  uniform sampling; ``ls'': 
    leverage score sampling; ``svd'': an {SVD} is used for computing
    the basis.}
    \label{fig:bbf_lr}
\end{figure}

\autoref{fig:bbf_lr} shows the relative error versus the memory
cost for different sampling methods. The relative error is
computed by $\frac{\|\hat K - K\|_F}{\|K\|_F}$, where $\hat K$
is the approximated kernel matrix, $K$ is the exact kernel matrix,
and $\|\cdot\|_F$ denotes the Frobenius norm. As can be
seen, the BBF structure is strictly a generalization of the low-rank
scheme, and achieves lower approximation error regardless of the
sampling method used. Moreover, for most sampling methods, the {BBF}
structure outperforms the best low rank approximation computed by an
{SVD}, which strongly implies that the {BBF} structure is favorable.

\subsection{BBF algorithm} \label{sec:advantage_algo}

In this section, we experimentally evaluate the performance of
our {BBF} algorithm with other state-of-art kernel approximation
methods. \autoref{sec:err_vs_mem} and \autoref{sec:err_vs_param}
examine the matrix reconstruction error under varying memory budget
and kernel bandwidth parameters. \autoref{sec:ridge} applies the
approximations to the kernel ridge regression problem. Finally,
\autoref{sec:ifgt} compares the linear complexity of {BBF} with the
IFGT~\cite{yang2003improved}. Throughout the experiments, we use {BBF}
to denote our algorithm, whose input parameters are computed from
our pre-computation algorithm.
    
In what follows, we briefly introduce some implementation and input
parameter details for the methods we are comparing to.
\begin{itemize}
    \item \emph{The na\"ive Nystr\"om (Nys)}. We uniformly sampled $2k$
    columns without replacement for a rank $k$ approximation.

    \item \emph{$k$-means Nystr\"om (kNys)}.  It uses $k$-means
    clustering and sets the centroids to be the landmark points.
    We used the code provided by the author.

    \item \emph{Leverage score Nystr\"om (lsNys)}. It samples
    columns with probabilities proportional to the statistical
    leverage scores.  We calculated the approximated leverage scores
    \cite{drineas2012fast} and sampled $2k$ columns with replacement
    for a rank-$k$ approximation.

    \item \emph{Memory Efficient Kernel Approximation (MEKA)}. We
    used the code provided by the author.

    \item \emph{Random Kitchen Sinks (RKS)}. We used our own MATLAB
    implementation based on their algorithm.

    \item \emph{Improved Fast Gauss Transform (IFGT)}. We used the
    \verb|C++| code provided by the author.
\end{itemize}
    
\subsubsection{Approximation with varying memory budget}
\label{sec:err_vs_mem}

We consider the reconstruction errors from different methods
when the memory cost varies. The memory cost (storage) is also
a close approximation of the running time for a matrix-vector
multiplication. In addition, computing memory is more accurate than
running time, which is sensitive to the implementation and algorithmic
details.  In our experiments, we indirectly increased the memory
cost by requesting a lower tolerance in BBF. The memories for all the
methods were fixed to be roughly the same in the following way. For
low rank methods, the input rank was set to be the memory of {BBF}
divided by the matrix size.  For MEKA, the input number of clusters
was set to be the same as {BBF}; the ``eta'' parameter (the percentage
of blocks to be set to zeros) was also set to be similar as {BBF}.
    
\begin{figure}[htbp]
    \centering
    \begin{subfigure}[b]{0.45\textwidth}
        \includegraphics[width=\textwidth]{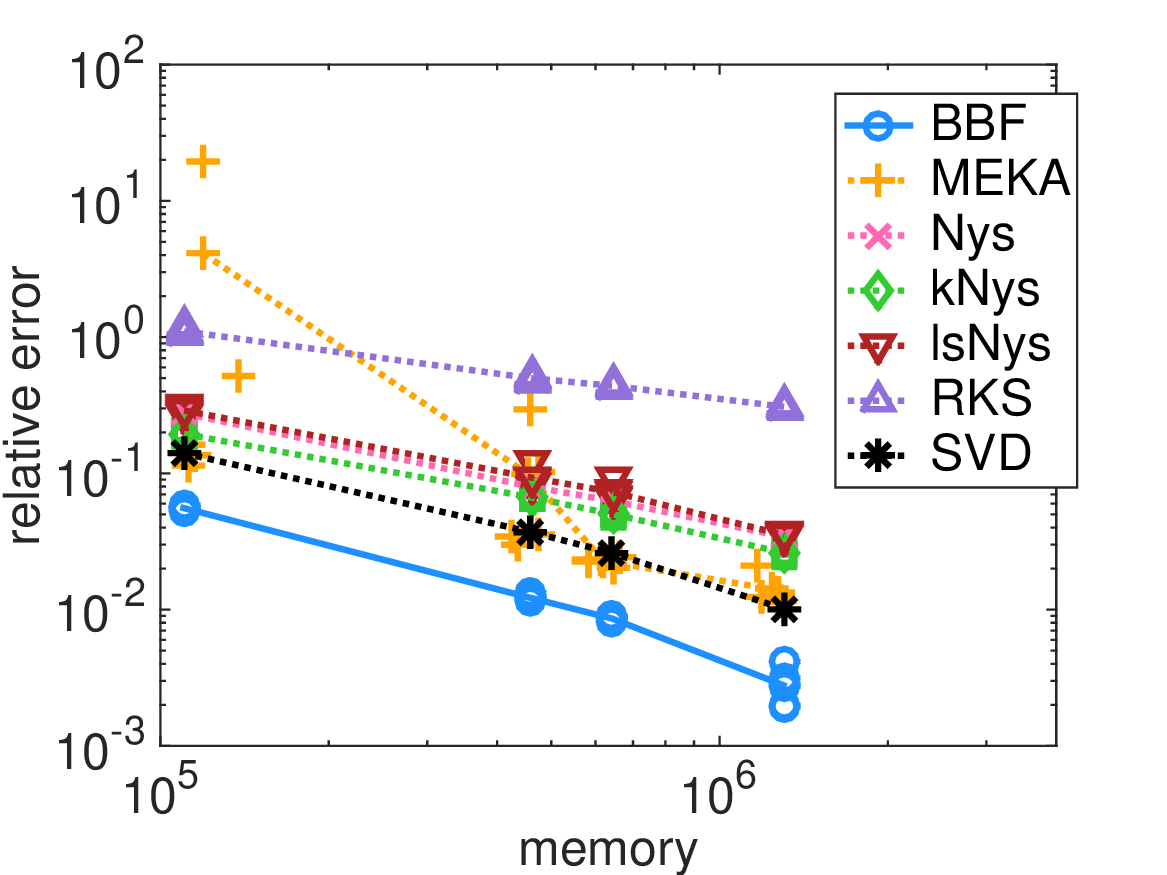}
        \caption{Abalone, $h$ = 1 }
    \end{subfigure}
    \begin{subfigure}[b]{0.45\textwidth}
        \includegraphics[width=\textwidth]{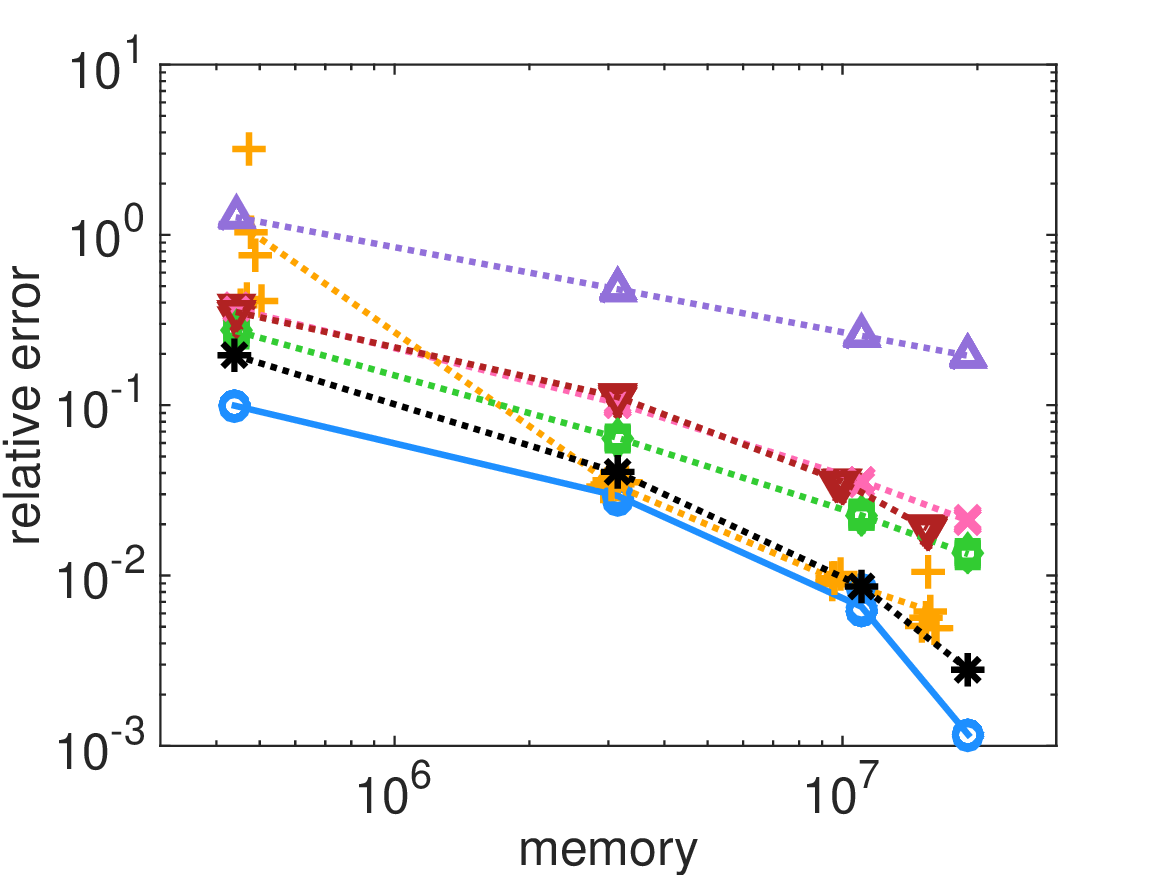}
        \caption{Pendigits, $h$ = 2}
    \end{subfigure}
    \begin{subfigure}[b]{0.45\textwidth}
        \includegraphics[width=\textwidth]{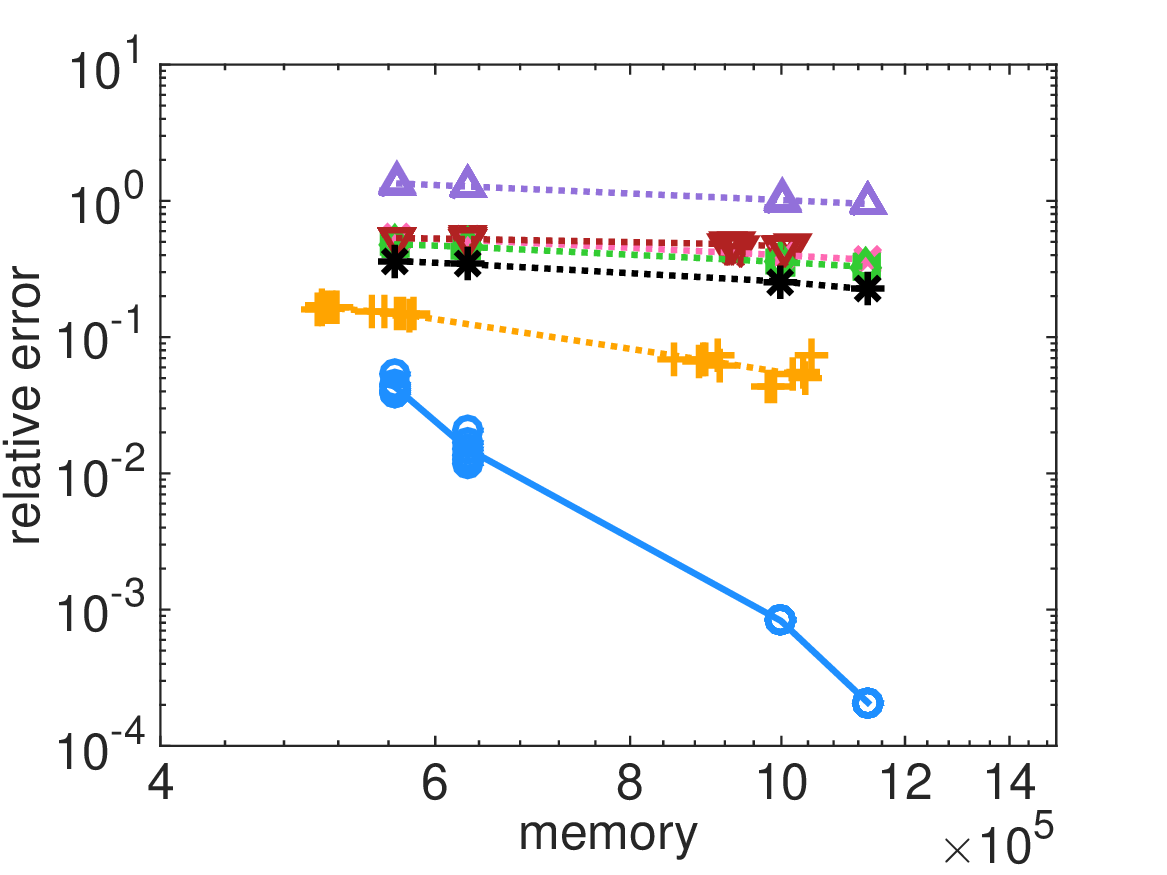}
        \caption{CTG, $h = 0.5$}
    \end{subfigure}
    \begin{subfigure}[b]{0.45\textwidth}
        \includegraphics[width=\textwidth]{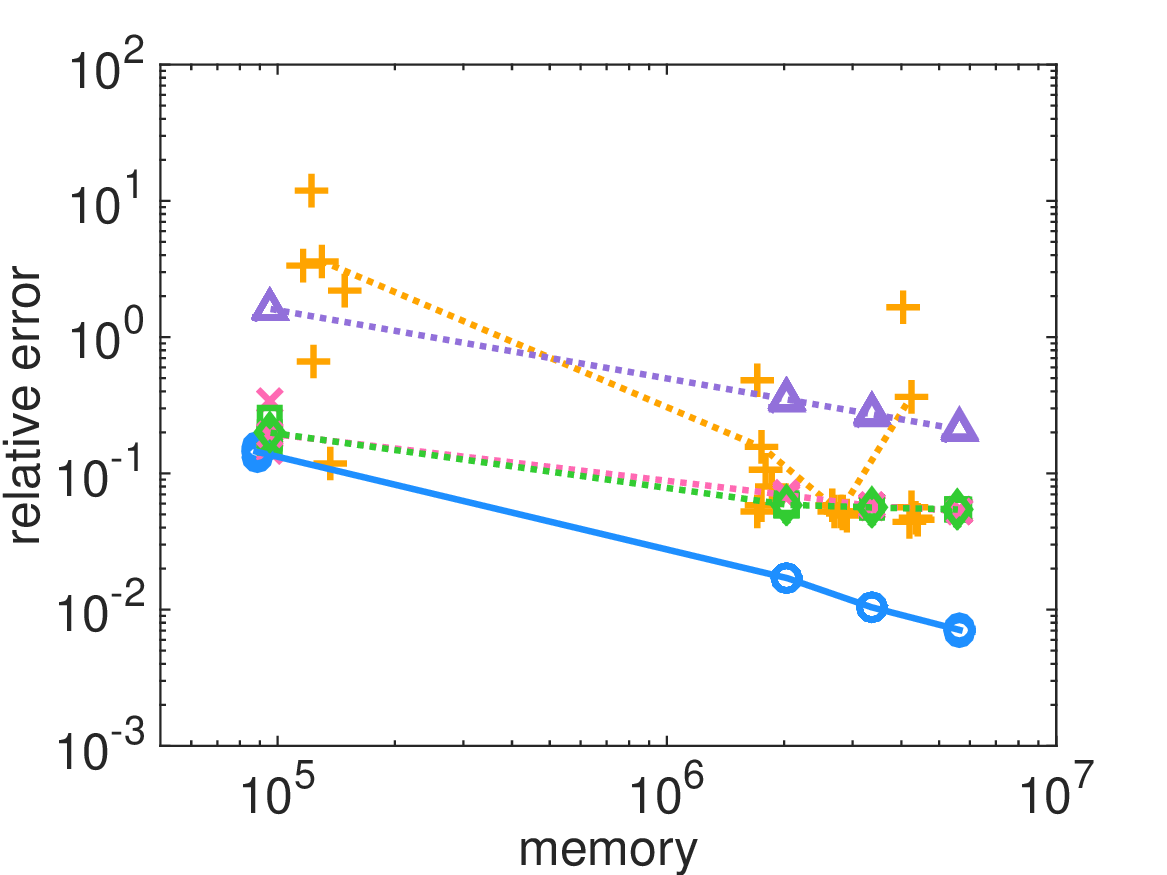}
        \caption{EMG, $h = 0.1$}
    \end{subfigure}
      \begin{subfigure}[b]{0.45\textwidth}
        \includegraphics[width=\textwidth]{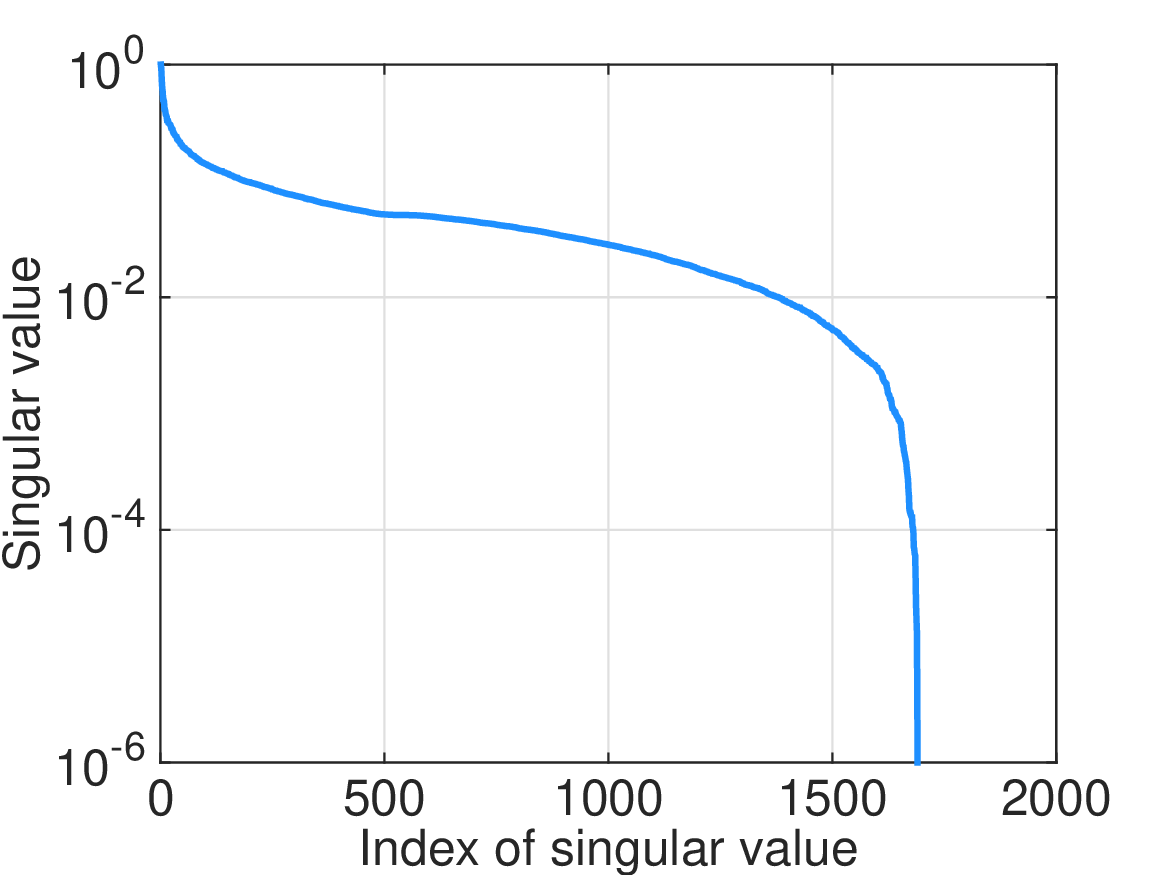}
        \caption{Singular values (CTG)}
    \end{subfigure}
    \begin{subfigure}[b]{0.45\textwidth}
        \includegraphics[width=\textwidth]{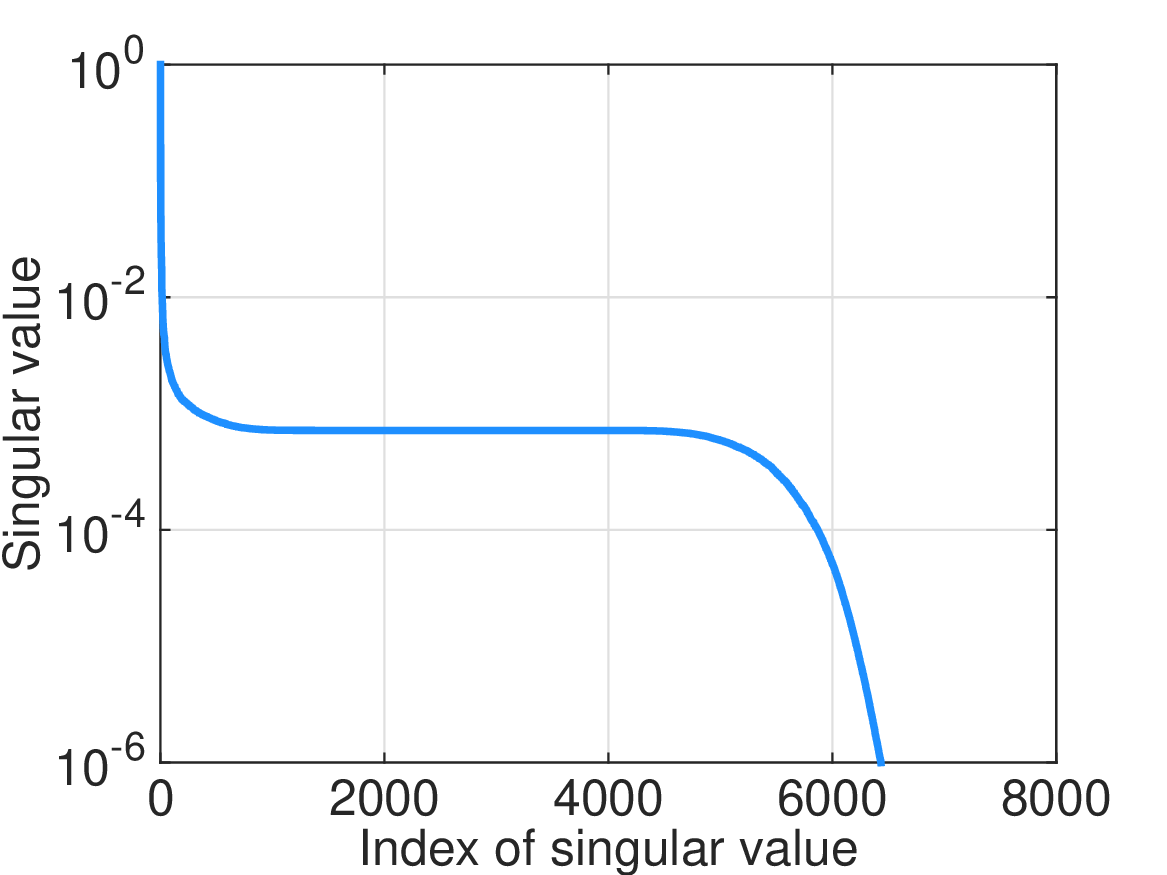}
        \caption{Singular values  (EMG)}
    \end{subfigure}
    \caption{Comparisons of BBF (our algorithm) with competing
    methods.  Top four plots (loglog scale) share the same legend.
    For each method, we show the improvement in error as more memory
    is available.  For each memory footprint, we report the error
    of 5 runs of each algorithm.  Each run is shown with a marker,
    while the lines represent the average error.  For CTG and EMG
    datasets, the parameter $h$ was chosen to achieve higher $F_1$
    score on smaller classes, which leads to matrices with higher
    ranks as shown by the plateau or slow decay of singular values in
    the bottom plots subplots (e) and (f).}
    \label{fig:err_vs_memory}
\end{figure}

\begin{figure}[htbp]
    \centering
    \begin{subfigure}[b]{0.45\textwidth}
        \includegraphics[width=\textwidth]{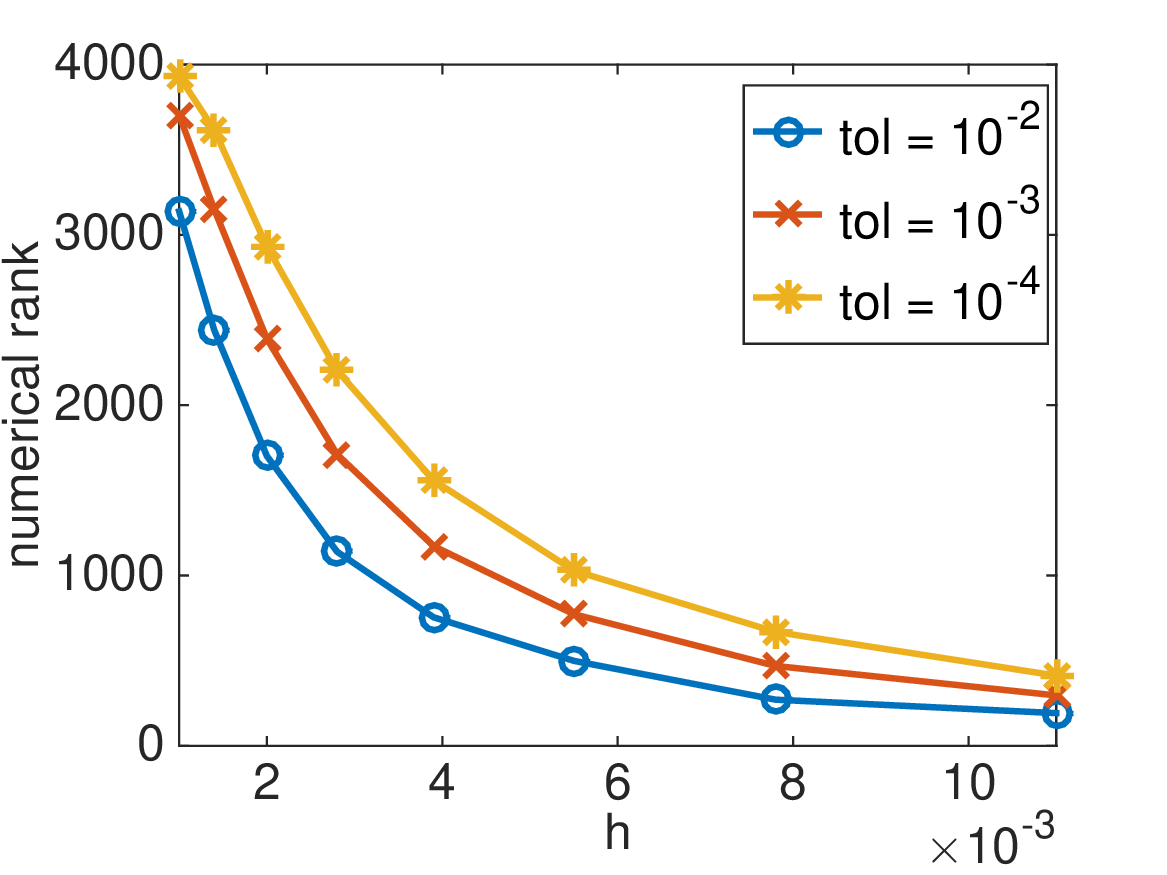}
        \caption{Numerical ranks of the kernel matrix}
    \end{subfigure}
    \begin{subfigure}[b]{0.45\textwidth}
        \includegraphics[width=\textwidth]{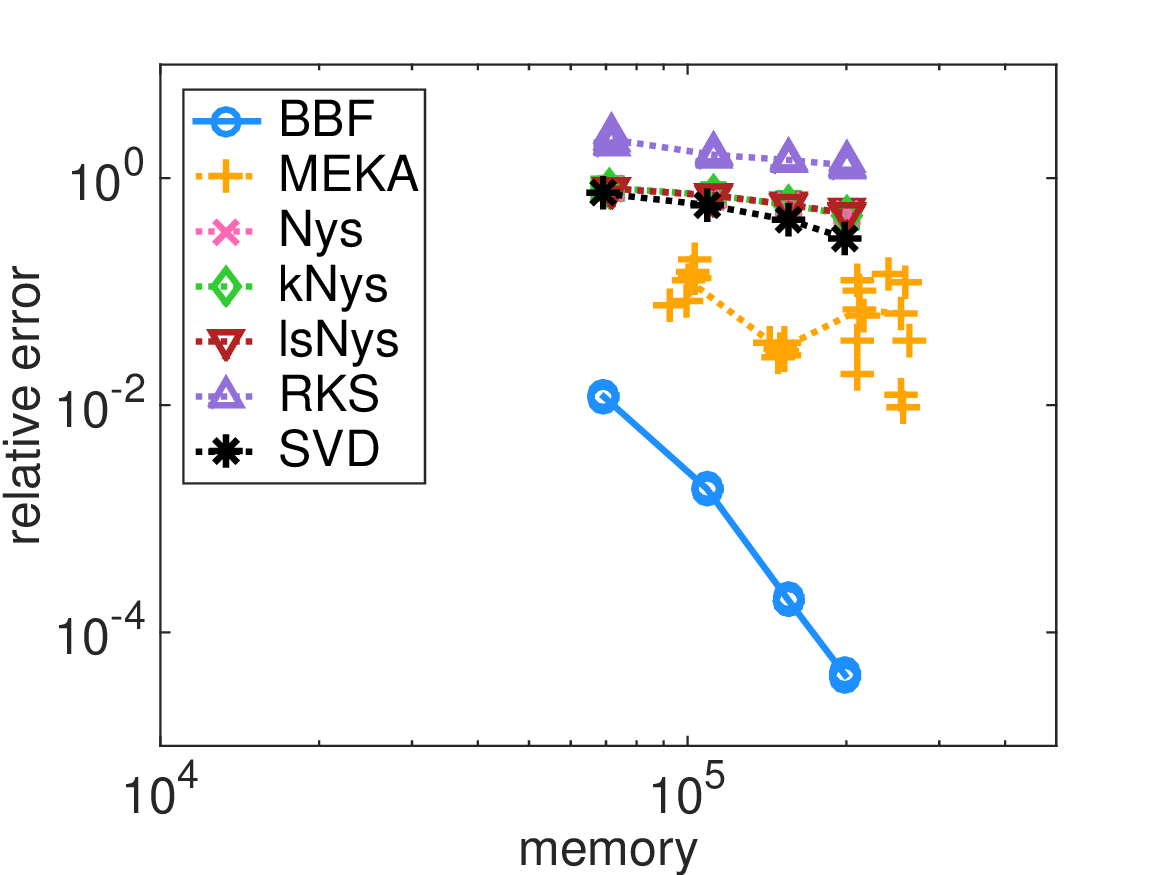}
        \caption{Relative error versus memory cost}
    \end{subfigure}
    \begin{subfigure}[b]{0.45\textwidth}
        \includegraphics[width=\textwidth]{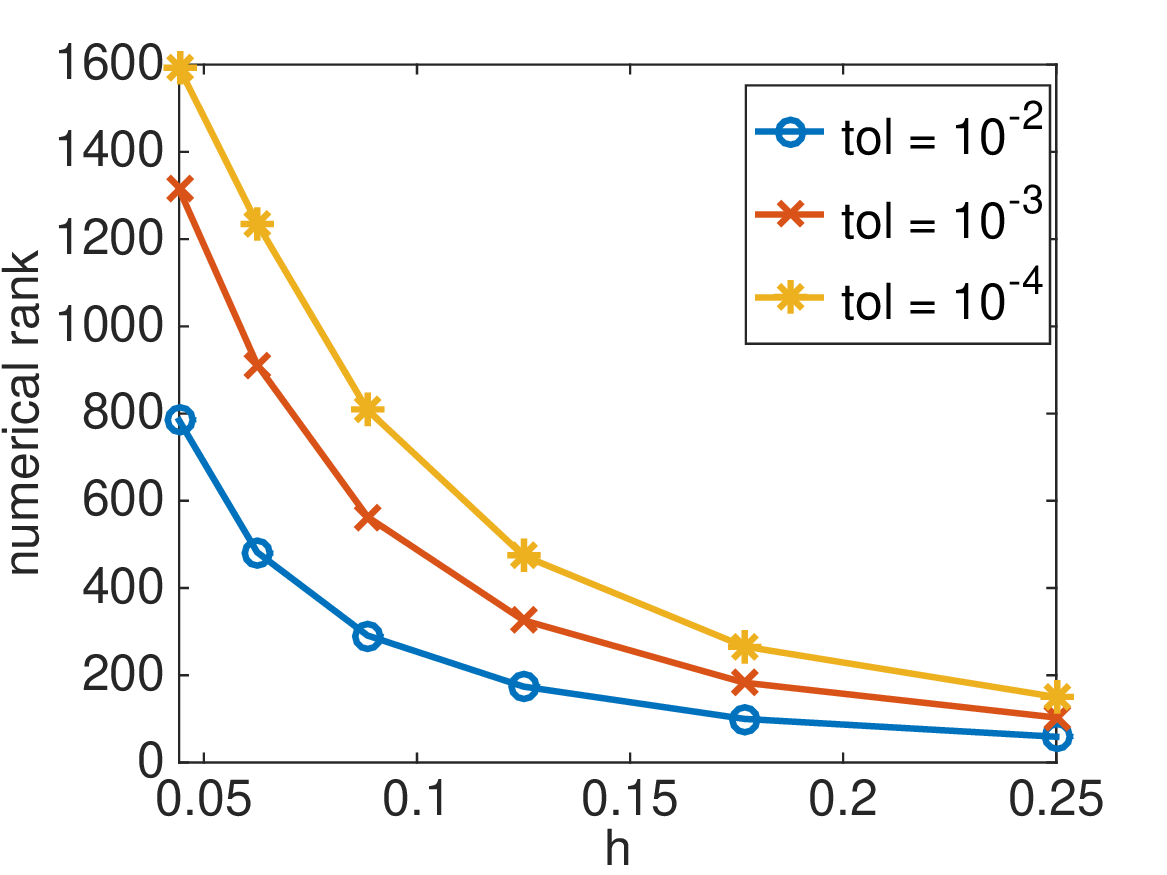}
        \caption{Numerical ranks of the kernel matrix}
    \end{subfigure}
    \begin{subfigure}[b]{0.45\textwidth}
        \includegraphics[width=\textwidth]{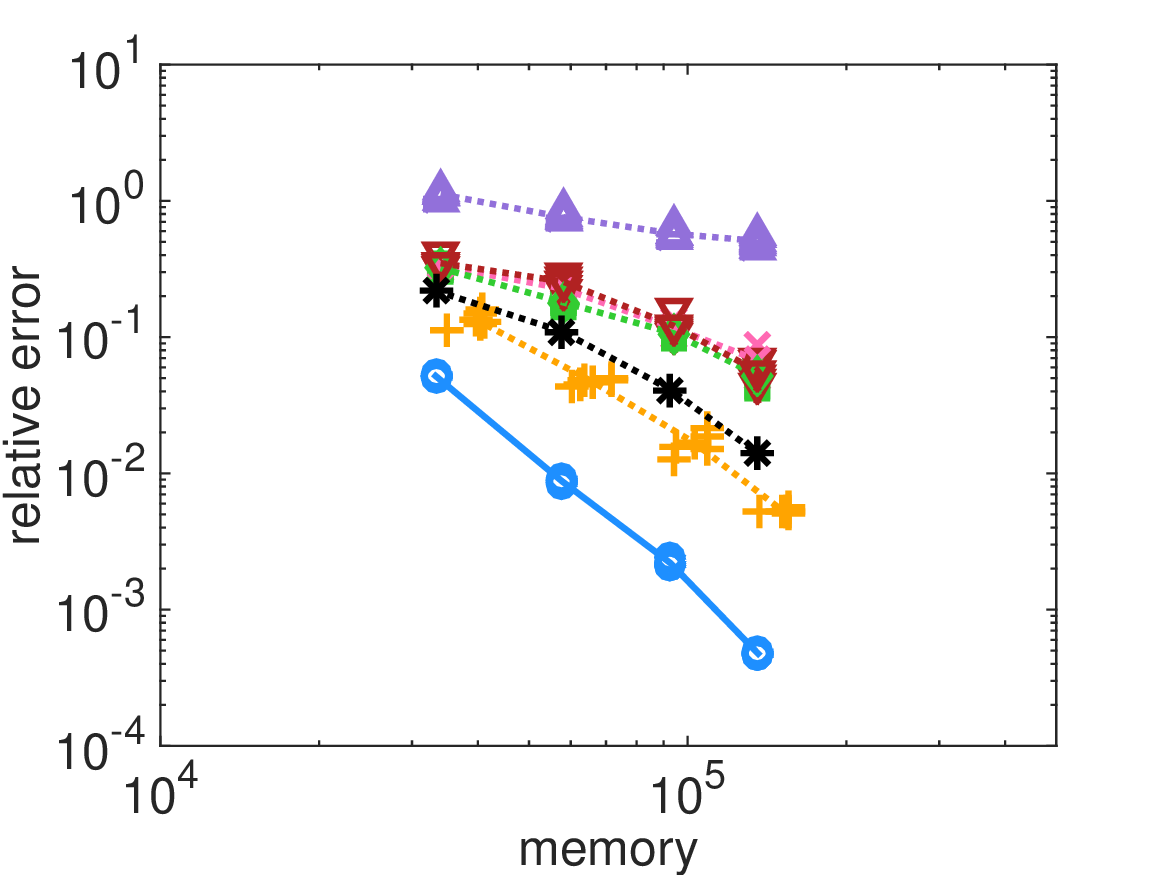}
        \caption{Relative error versus memory cost}
    \end{subfigure}
    \caption{The left plots report the numerical ranks of matrices
    versus $h$ and the right plots report the relative error
    versus memory. The data for top plots has alternating labels
    (see \autoref{fig:alternate} with 100 clusters) while the
    data for the bottom plots has smaller clusters surrounded by
    larger ones (see \autoref{fig:twocircles}). The values of $h$
    reported in subplot $(a)$ and $(c)$ yield test accuracy greater
    than 0.99. The $h$ in $(b)$ and $(d)$ are the largest optimal
    $h$ with value 0.0127 and 0.25 respectively.  For each memory
    cost, we report the relative error of 5 runs of each algorithm.
    The number of clusters for BBF was fixed at 20 for subplot $(b)$
    and selected automatically for subplot $(d)$.}
    \label{fig:morealternate_small_large_BBF}
\end{figure}

\autoref{fig:err_vs_memory} and
\autoref{fig:morealternate_small_large_BBF} show the reconstruction
error versus memory cost on real datasets and 2D synthetic datasets,
respectively. We see that {BBF} achieves comparable and often
significantly lower error than the competing methods regardless of the
memory cost. There are two observations worth noting. First, the {BBF}
outperforms the exact {SVD} which is the best rank-$r$ approximation,
and it outperforms with a factorization complexity that is only linear
rather than cubic. This has demonstrated the superiority of the {BBF}
structure over the low-rank structure. Second, even when compared to a
similar structure as MEKA, {BBF} achieves a lower error whose variance
is also smaller, and it achieves so with a similar factorization
complexity. These have verified that the representation of {BBF}
is more accurate and the constructing algorithm is more stable.
    
\subsubsection{Approximation with varying kernel bandwidth parameters}
\label{sec:err_vs_param}

We consider the reconstruction errors with varying decay patterns of
singular values, which we achieve by choosing a wide range of kernel
bandwidth parameters. The memory for all methods are fixed to be
roughly the same.
    
\begin{figure}[htbp]
    \centering
    \begin{subfigure}[b]{0.45\textwidth}
        \includegraphics[width=\textwidth]{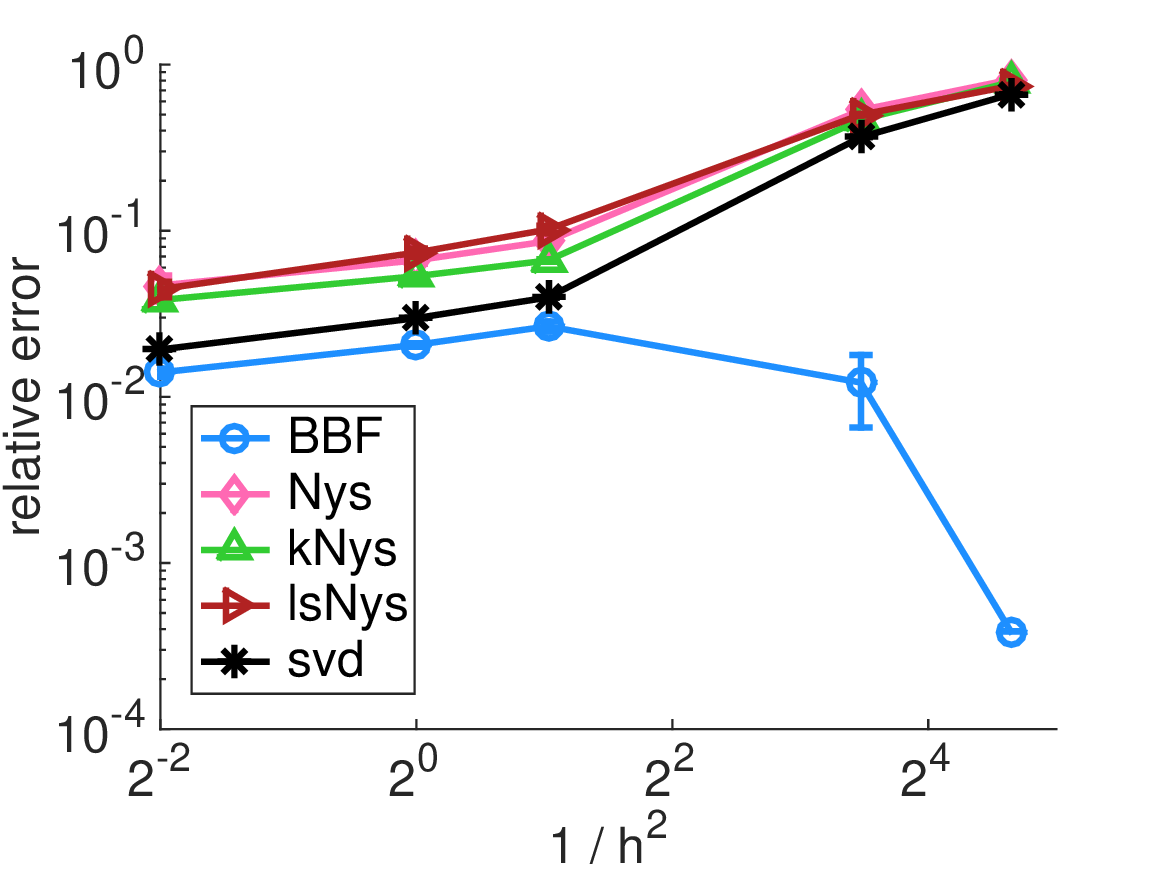}
        \caption{Abalone}
    \end{subfigure}
    \begin{subfigure}[b]{0.45\textwidth}
        \includegraphics[width=\textwidth]{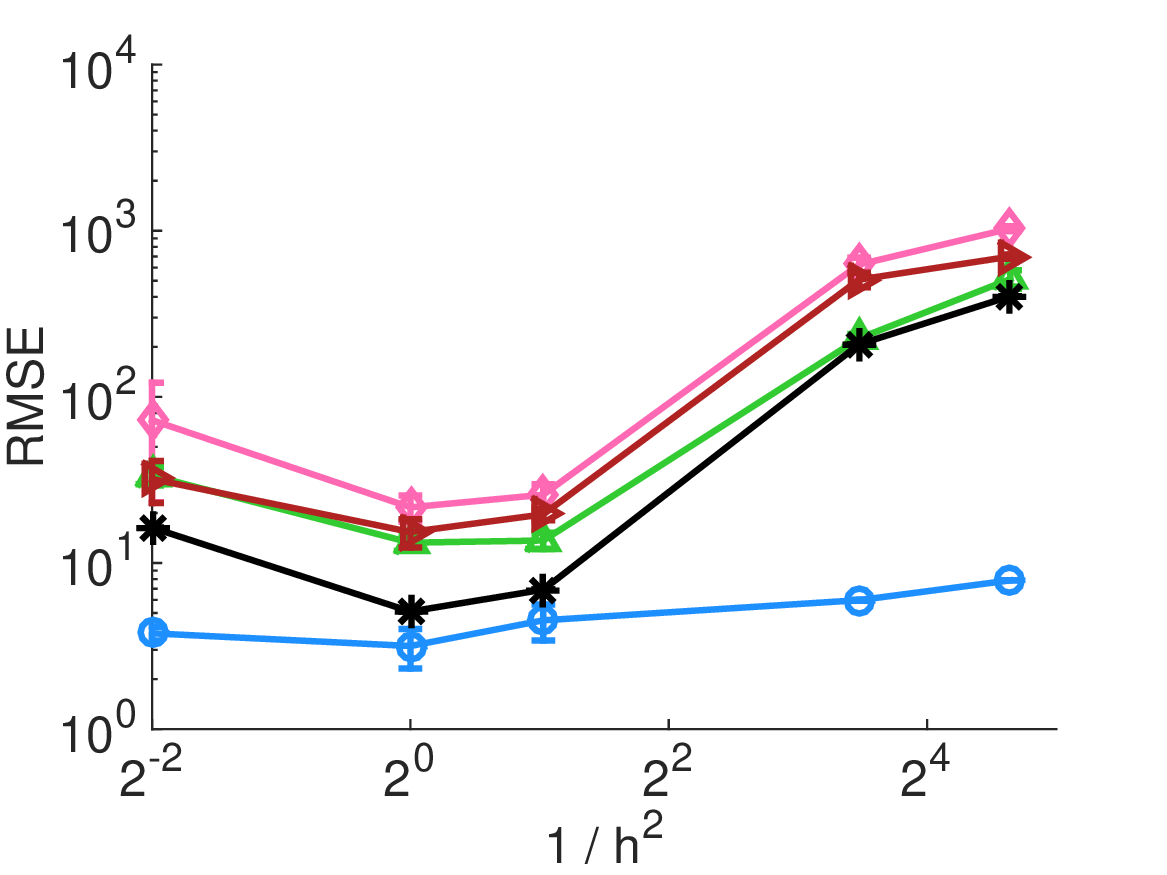}
        \caption{Abalone}
    \end{subfigure}
    \begin{subfigure}[b]{0.45\textwidth}
        \includegraphics[width=\textwidth]{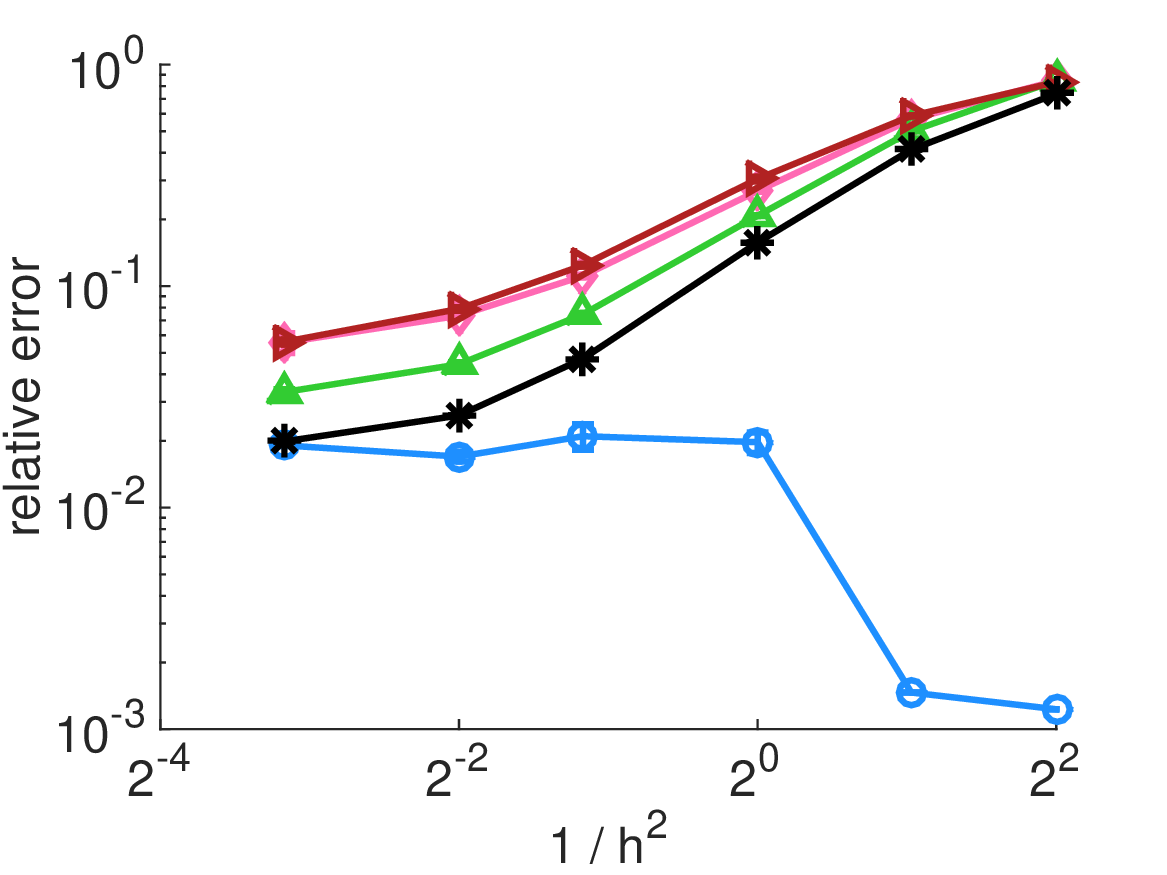}
        \caption{Pendigits}
    \end{subfigure}
    \begin{subfigure}[b]{0.45\textwidth}
        \includegraphics[width=\textwidth]{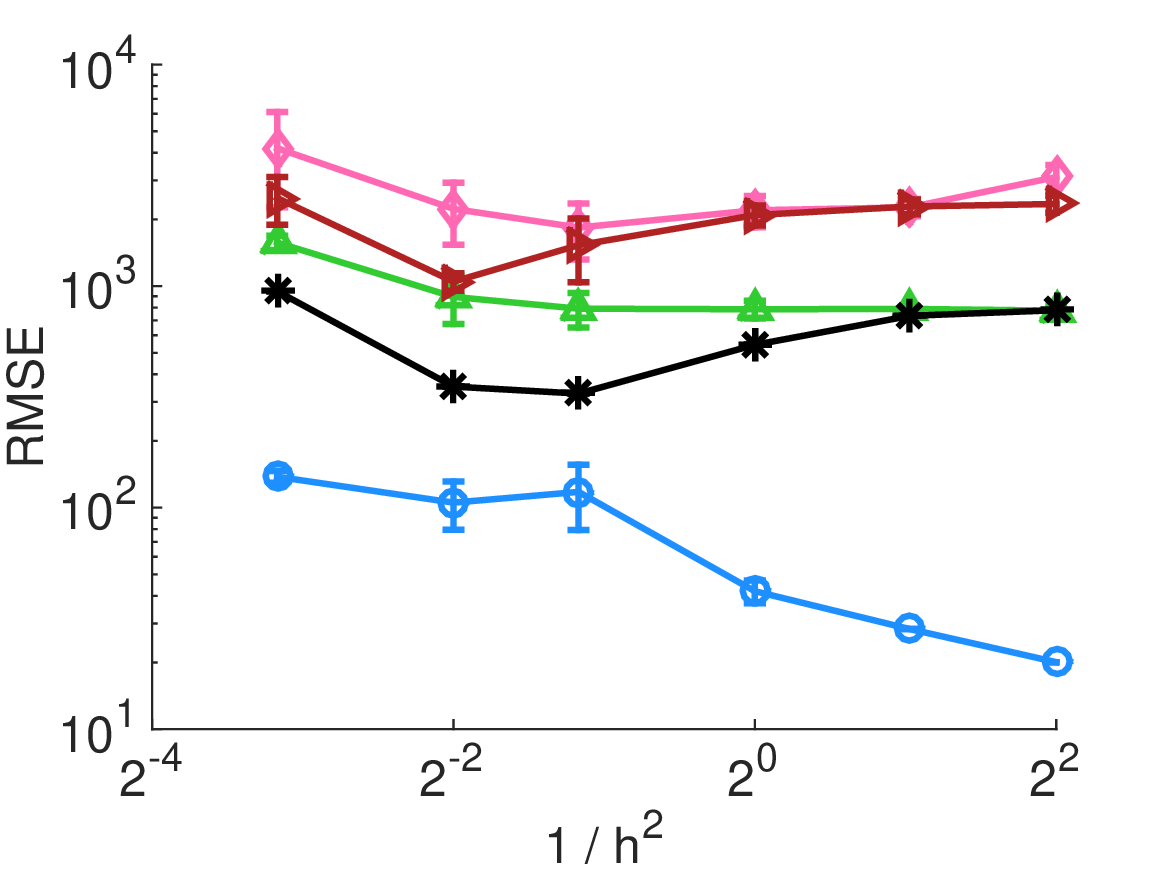}
        \caption{Pendigits}
    \end{subfigure}
    \caption{Plots for relative error vers $1/h^2$ for different kernel
    approximation methods.  The memory costs for all methods and all
    kernel parameters are fixed to be roughly the same.  The kernel
    function used is the Gaussian kernel.  As the rank of the matrix
    increases ($h$ decreases), the gap in error between low-rank
    approximation methods and BBF increases.}
    \label{fig:params_vs_err}
\end{figure}

The plots on the left of \autoref{fig:params_vs_err} show the average
matrix reconstruction error versus $1/h^2$.  We see that for all
the low-rank methods, the error increases when $h$ decreases. When
$h$ becomes smaller, the kernel function becomes less smooth,
and consequently the matrix rank increases. This relation between
$h$ and the matrix rank are revealed in some statistics listed in
\autoref{tab:stats_data}. The results in the table are consistent
with the results shown in \cite{gittens2016revisiting} for varying
kernel bandwidth parameters.
    
\begin{table*}[htbp]
    \small
    \centering

    \caption{Summary statistics for abalone and pendigits datasets
    with the Gaussian kernel, where $r$ is the rank, $M$ is the exact
    matrix, and $M_r$ is the best rank-$r$ approximation for $M$.
    $\Big\lceil \frac{\|M\|_F^2}{\|M\|_2^2} \Big\rceil$ is referred
    to as the stable rank and is an underestimate of the rank.
    $l_r$ represents the $r$-th largest leverage score scaled by
    $\tfrac{n}{r}$.}
    \label{tab:stats_data}

    \vskip 0.15in
    \begin{tabular}{cccc| cccc}
        \toprule
        \multicolumn{4}{c|}{{\bf{Abalone}} ($r$ = 100)}
        & \multicolumn{4}{c}{{\bf{Pendigits}}  ($r$ = 252)} \\
        \midrule
        $\frac{1}{h^2}$
        & $\Big\lceil \frac{\|M\|_F^2}{\|M\|_2^2} \Big\rceil$
        & $100\frac{\|M_r\|_F}{\|M\|_F}$ & $l_r$
        & $\frac{1}{h^2}$
        & $\Big\lceil \frac{\|M\|_F^2}{\|M\|_2^2} \Big\rceil$
        & $100\frac{\|M_r\|_F}{\|M\|_F}$ & $l_r$ \\
        \midrule
        0.25 &    2    & 99.99 &  4.34 &
        0.1 &    3  & 99.99 &  2.39 \\
        1 &    4 & 99.86 &  2.03 &  
        0.25 &    6 & 99.79 &  1.83 \\
        4 &    5  & 97.33 &  1.94 & 
        0.44 &    8 & 98.98 &  1.72 \\
        25 &   15  & 72.00 &  5.20 &  
        1 &   12 & 93.64 &  2.02 \\
        100 &  175  & 33.40 & 12.60 &  
        2 &   33 & 77.63 &  2.90 \\
        400 &  931 & 19.47 & 20.66 &  
        4 &  207  & 49.60 &  4.86 \\
        1000 & 1155 & 16.52 & 20.88 &  
        25 & 2794  & 19.85 & 14.78 \\
        \bottomrule
    \end{tabular}
\end{table*}
    
In the large-$h$ regime, the gap in error between {BBF} and other
methods is small. In such regime, the matrix is low rank, and the
low-rank algorithms work effectively. Hence, the difference in
error is not significant. In the small-$h$ regime, the gap starts
to increase. In this regime, the matrix becomes close to diagonal
dominant, and the low-rank structure, as a global structure,
cannot efficiently capture the information along the diagonal;
while for {BBF}, the pre-computation procedure will increase the
number of clusters to better approximate the diagonal part, and
the off-diagonal blocks can be set to 0 due to their small entries.
By efficiently using the memory, {BBF} is favorable in all cases,
from low-rank to nearly diagonal.

\subsubsection{Kernel ridge regression}
\label{sec:ridge}

We consider the kernel ridge regression.  The standard optimization
problem for the kernel ridge regression is
\begin{equation}
    \min_{\vecalpha} \|K\vecalpha - \vecy\|^2 + \lambda \| \vecalpha
    \|^2, \label{eq:ridge}
\end{equation}
where $K$ is a kernel matrix, $\vecy$ is the target, and $\lambda >
0$ is the regularization parameter. The minimizer is given by the
solution of the following linear system
\begin{equation} \label{eq:linear_system}
    (K + \lambda I) \vecalpha = \vecy.
\end{equation}
The linear system can be solved by an iterative solver,
\emph{e.g.} MINRES \cite{paige1975solution}, and the complexity
is $O(n^2T)$ where $n^2$ is from matrix-vector multiplications
and $T$ denotes the iteration number. If we can approximate $K$
by $\widehat{K}$ which can be represented in lower memory, then
the solving time can be accelerated. This is because the memory
is a close approximation for the running time of a matrix-vector
multiplication. We could also solve the approximated system directly
when the matrix can be well-approximately by a low-rank matrix, that
is, we compute the inversion of $\widehat K$ first by the Woodbury
formula\footnote{https://en.wikipedia.org/wiki/Woodbury\_matrix\_identity}
and then apply the inversion to $\vecy$.

\begin{figure}[htbp]
    \centering
    \begin{subfigure}[b]{0.45\textwidth}
        \includegraphics[width=\textwidth]{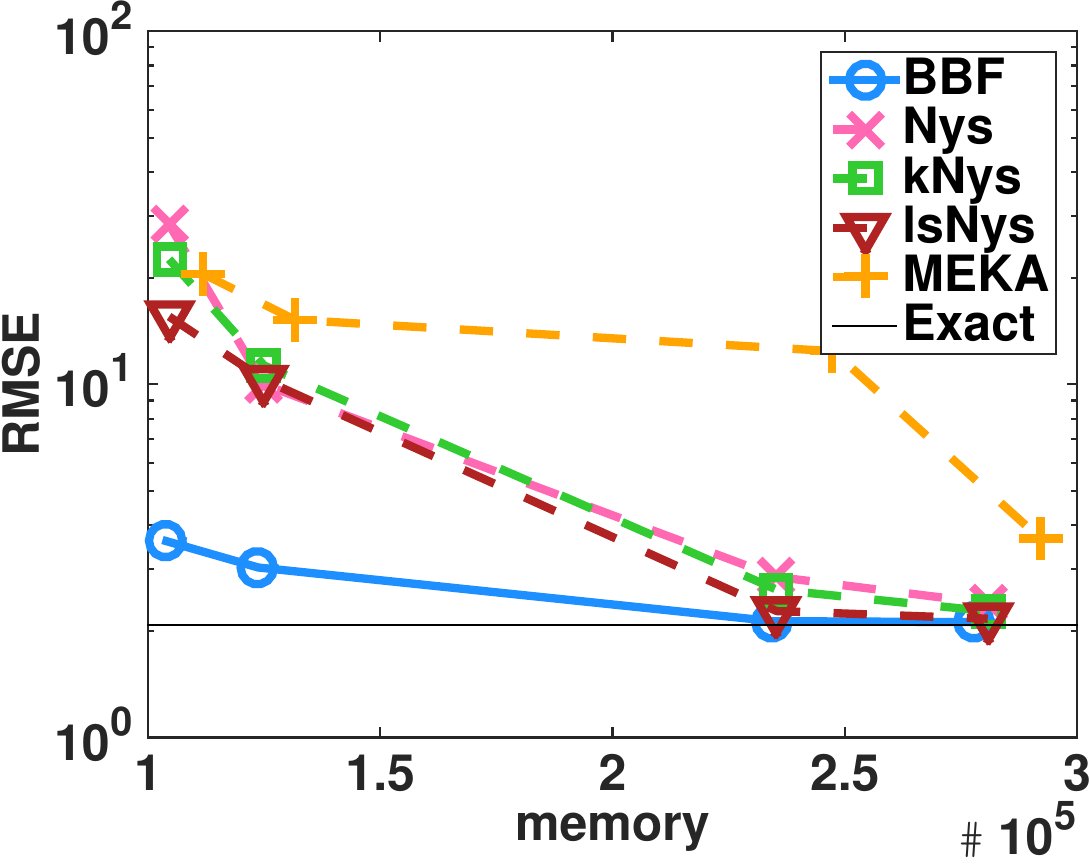}
        \caption{Abalone, (4, $2^{-4}$)}
    \end{subfigure}
    \begin{subfigure}[b]{0.45\textwidth}
        \includegraphics[width=\textwidth]{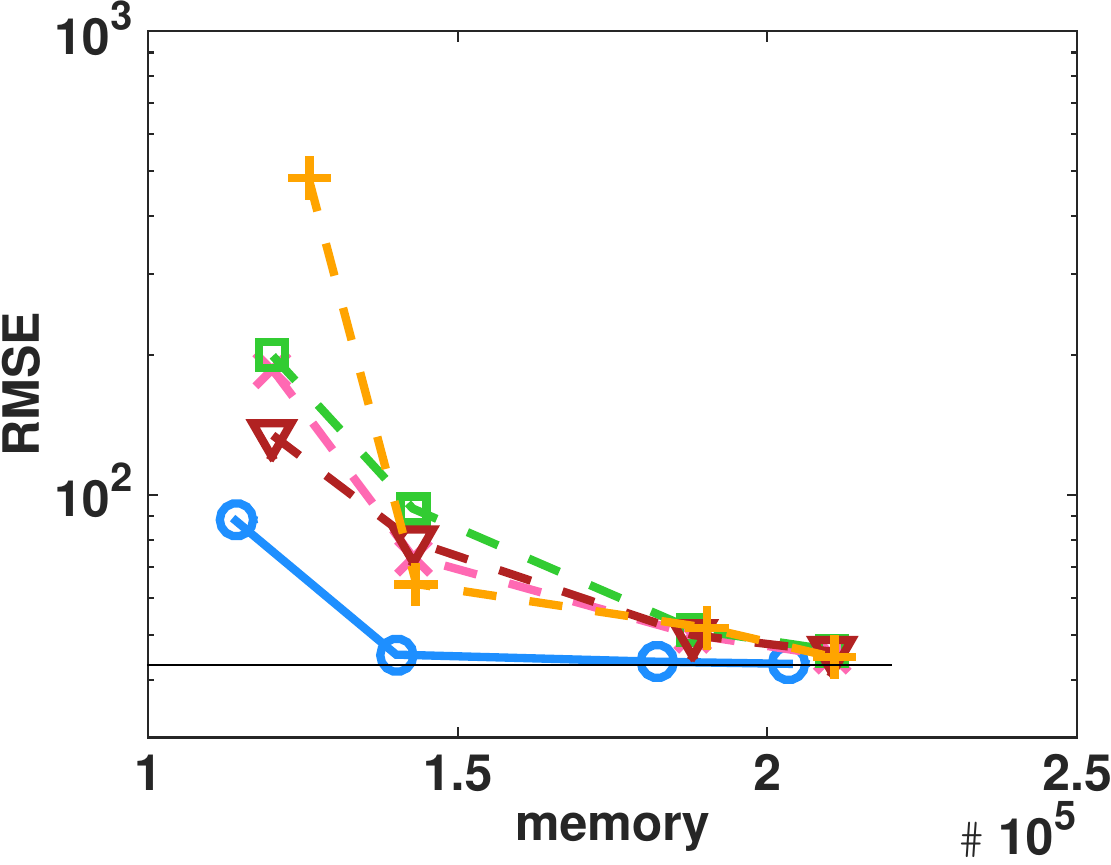}
        \caption{Pendigits, (22.6, $2^{-5}$)}
    \end{subfigure}
    \begin{subfigure}[b]{0.45\textwidth}
        \includegraphics[width=\textwidth]{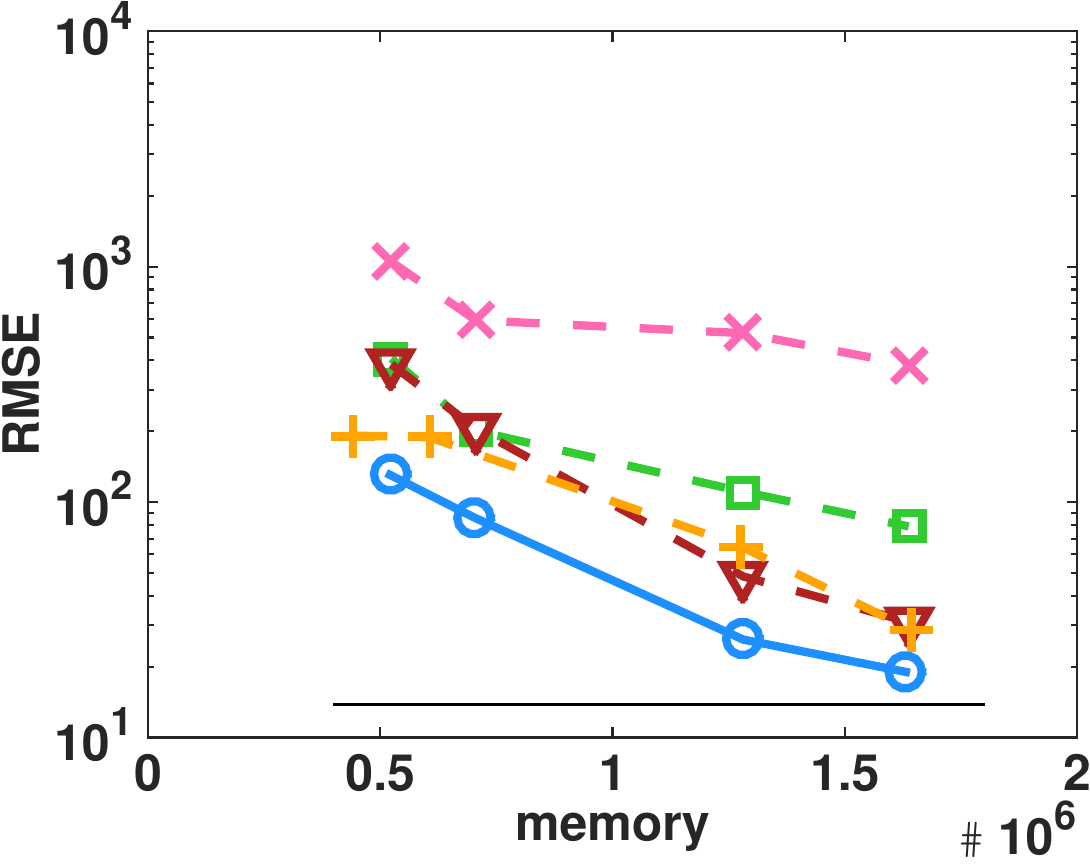}
        \caption{Cpusmall, (4.94, $2^{-5}$)}
    \end{subfigure}
    \begin{subfigure}[b]{0.45\textwidth}
        \includegraphics[width=\textwidth]{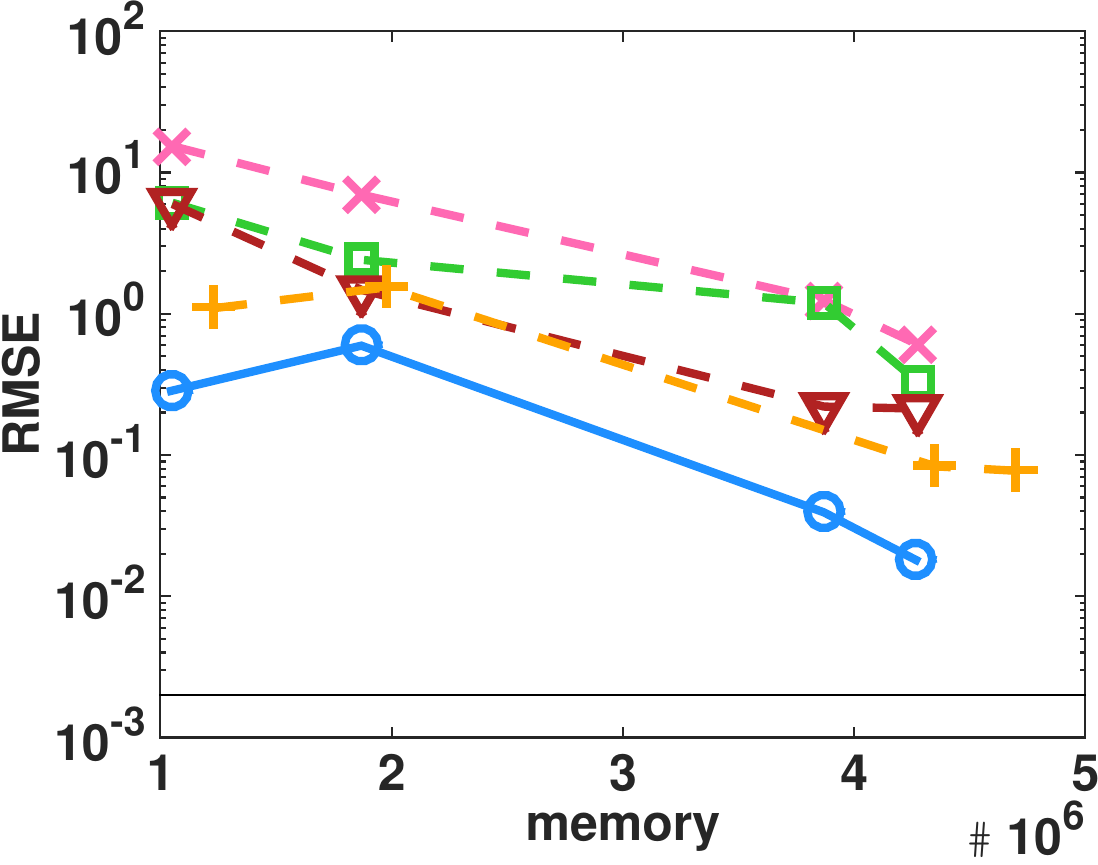}
        \caption{Mushroom, (4.63, $2^{-5}$)}
    \end{subfigure}
    \caption{Test RMSE versus memory for kernel ridge regression. For
    each memory cost, we report the results averaged over 5 runs. The
    black line on the bottom of each plot represents the test RMSE
    when using the exact matrix. The kernel parameter $h$ and the
    regularization parameter $\lambda$ were selected by a 5-fold
    cross-validation with a grid search, and the selected $(h,\lambda)$
    pairs are listed in the subcaption.}
    \label{fig:rmse}
\end{figure}

In the experiments, we approximated $K$ by $\widehat{K}$, and solved
the following approximated system with MINRES.
\begin{equation} \label{eq:approx_linear_system}
    (\widehat K + \lambda I) \vecalpha = \vecy.
\end{equation}
The dataset was randomly divided into training set (80\%) and
testing set (20\%). \added{}{}{The kernel used is the Laplacian kernel $K(\vecx, \vecy) = \exp(\|\vecx-\vecy\|/h)$ for this subsection}. We report the test root-mean-square error (RMSE)
which is defined as
\begin{equation}
    \sqrt{\frac{1}{n_{\text{test}}}\| K_{\text{test}} \hat \vecalpha -
    \vecy_{\text{test}}\|_F^2},
\end{equation}
where $K_{\text{test}}$ is the interaction matrix between the test
data and training data, $\hat \vecalpha$ is the solution from solving
\autoref{eq:approx_linear_system}, and $\vecy_{\text{test}}$ is the
true test target. \autoref{fig:rmse} shows the test RMSE with varying
memory cost of the approximation. We see that with the same memory
footprint, the {BBF} achieves lower test error.

\added{R1}{}{

\emph{Discussion.}
For downstream prediction tasks, better generalization error could be achieved by using the surrogate kernel, which is the kernel matrix between the testing points and landmark points, instead of the exact kernel matrix for na\"ive Nystr\"om, $k$-means Nystr\"om, leverage score Nystr\"om, and random kitchen sink. Based on our experience, using surrogate kernels with Nystr\"om methods and random Fourier methods achieves competitive testing accuracy as that of BBF. Hence, although BBF significantly outperforms Nystr\"om methods and random Fourier methods in the approximation of kernel matrices, the advantage of BBF in prediction compared with surrogate kernels is less pronounced.

Meanwhile, an easy modification of BBF can be used to construct a surrogate kernel for downstream predictions as well. Specifically, for $U_i$, we can set $U_i$ as the carefully sampled important columns with points denoted as $X_i$, instead of the column basis of those sampled columns. This further reduces our factorization cost due to the removal of the orthonormalization step. Once these important columns are available, the middle matrix $C$ can be constructed identical to that in Algorithm~\ref{alg: fast}. These steps construct the modified BBF, which can be used to accelerate the linear system solve of \eqref{eq:approx_linear_system} and obtain $\vecalpha$ efficiently. Then, the coefficient for the surrogate kernel is computed as $\widetilde{\vecalpha} = C U^\dagger \vecalpha$. We denote $\widetilde{\vecalpha}_i$ as the coefficient for cluster $i$. The downstream prediction task, then, is divided into two steps. First, for a testing point $\vecx_{test}$, we find the cluster $i$ that $\vecx_{test}$ belongs to. Second, we compute the predictions $y_{test}$ as $K(\vecx_{test},X_i)\widetilde{\vecalpha_i}$.

With this modified BBF and the corresponding prediction procedure, assuming a surrogate kernel of the same size is used, it will be more efficient to compute the coefficients of the surrogate kernel as well as the predictions through BBF than Nystr\"om methods or random Fourier methods.
}

\subsubsection{Comparison with the improved fast gauss transform (IFGT)}
\label{sec:ifgt}

We benchmarked the linear complexity of the Improved Fast Gauss
Transform (IFGT)~\cite{yang2003improved} and {BBF}. IFGT was proposed
to alleviate the dimensionality issue for the Gaussian kernel.  For a
fixed dimension $d$, the IFGT has a linear complexity in terms of
application time and memory cost; regrettably, when $d$ increases
(\emph{e.g.}, $d \ge 10$) the algorithm requires a large number of
data points $n$ to make this linear behavior visible. BBF, on the
other hand, does not require a large $n$ to observe a linear growth.
    
\begin{figure}[htbp]
    \centering
    \begin{subfigure}[b]{0.45\textwidth}
        \includegraphics[width=\textwidth]{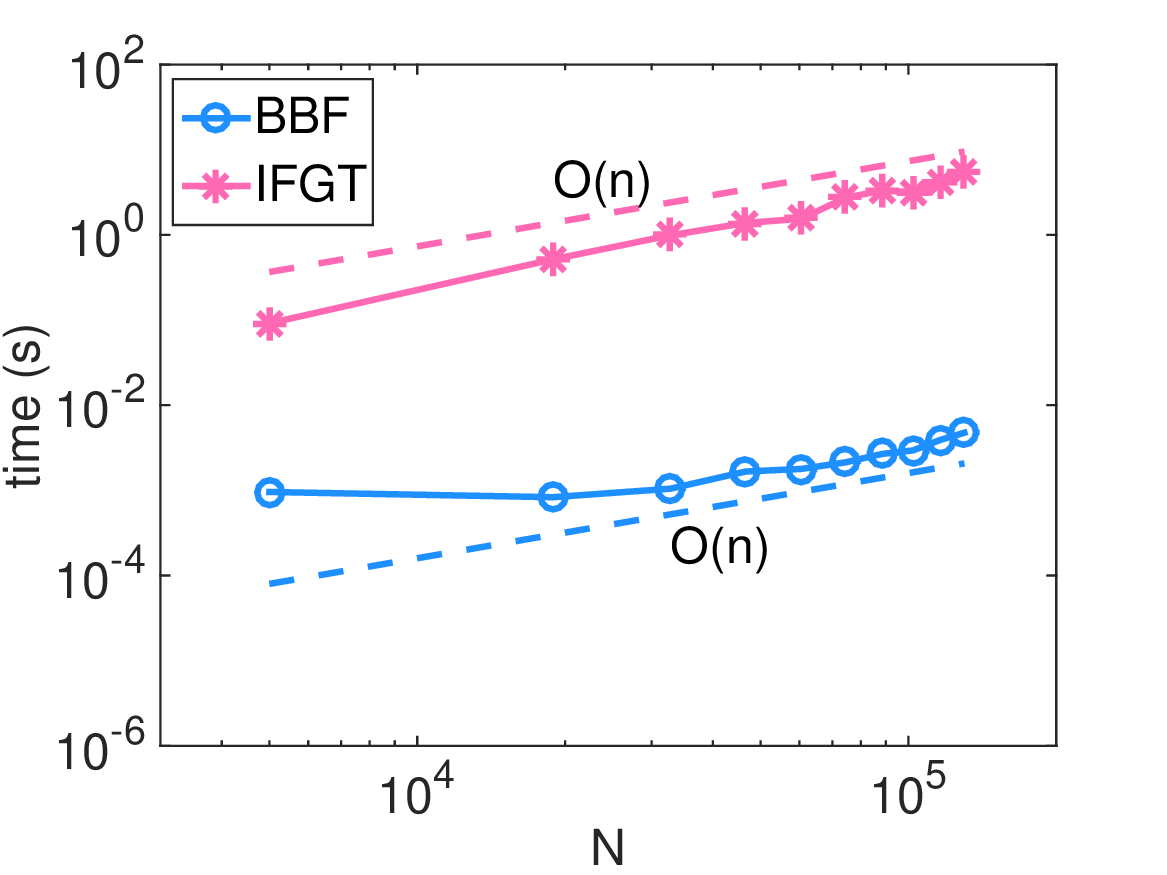}
        \caption{Application time (d = 5)}
    \end{subfigure}
    \begin{subfigure}[b]{0.45\textwidth}
        \includegraphics[width=\textwidth]{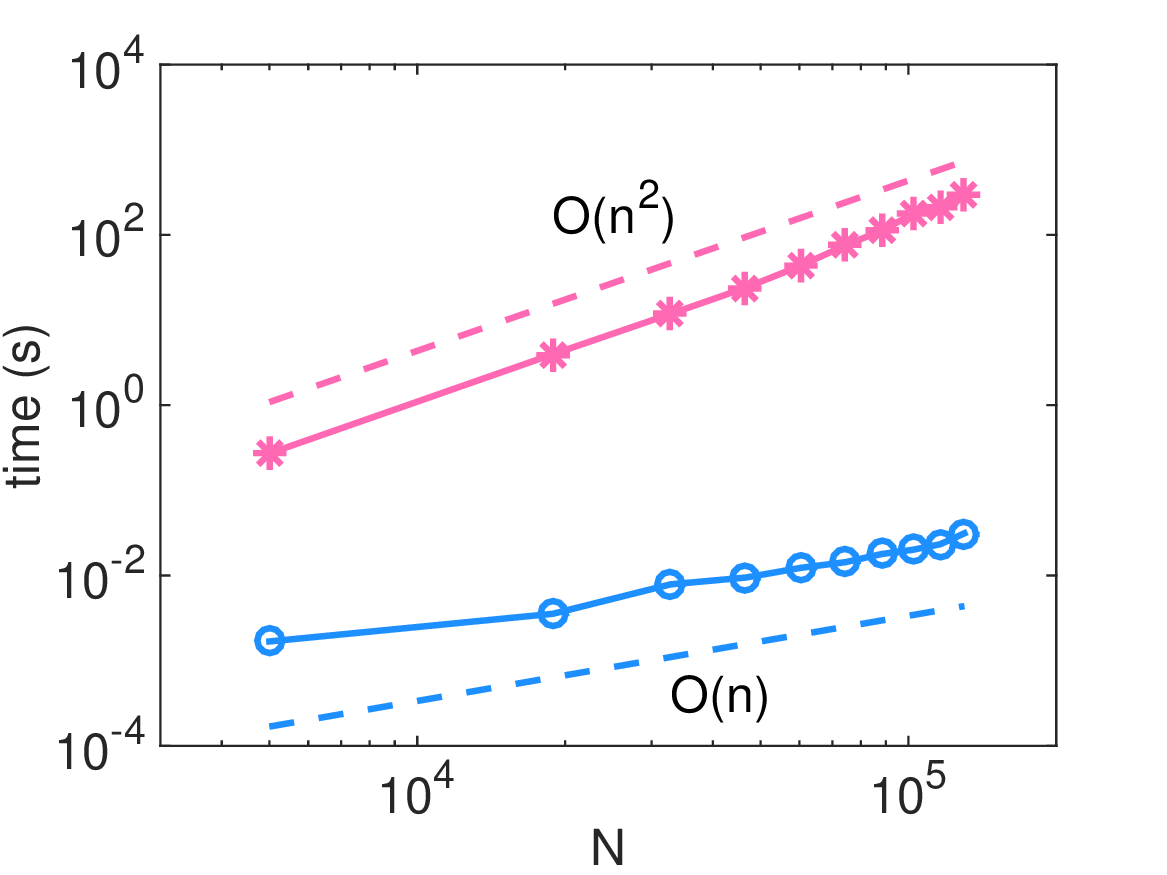}
        \caption{Application time (d = 40)}
    \end{subfigure}
    \begin{subfigure}[b]{0.45\textwidth}
        \includegraphics[width=\textwidth]{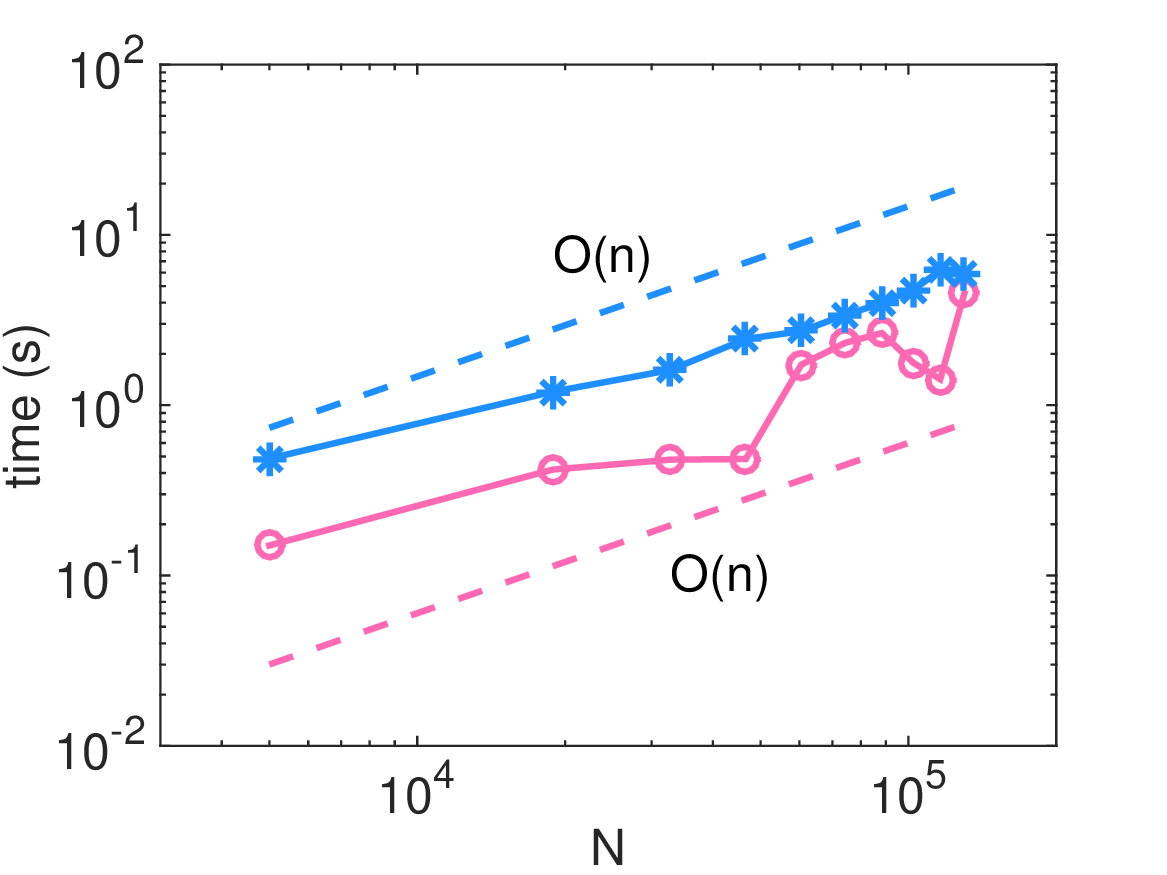}
        \caption{Total time (d = 5)}
    \end{subfigure}
    \begin{subfigure}[b]{0.45\textwidth}
        \includegraphics[width=\textwidth]{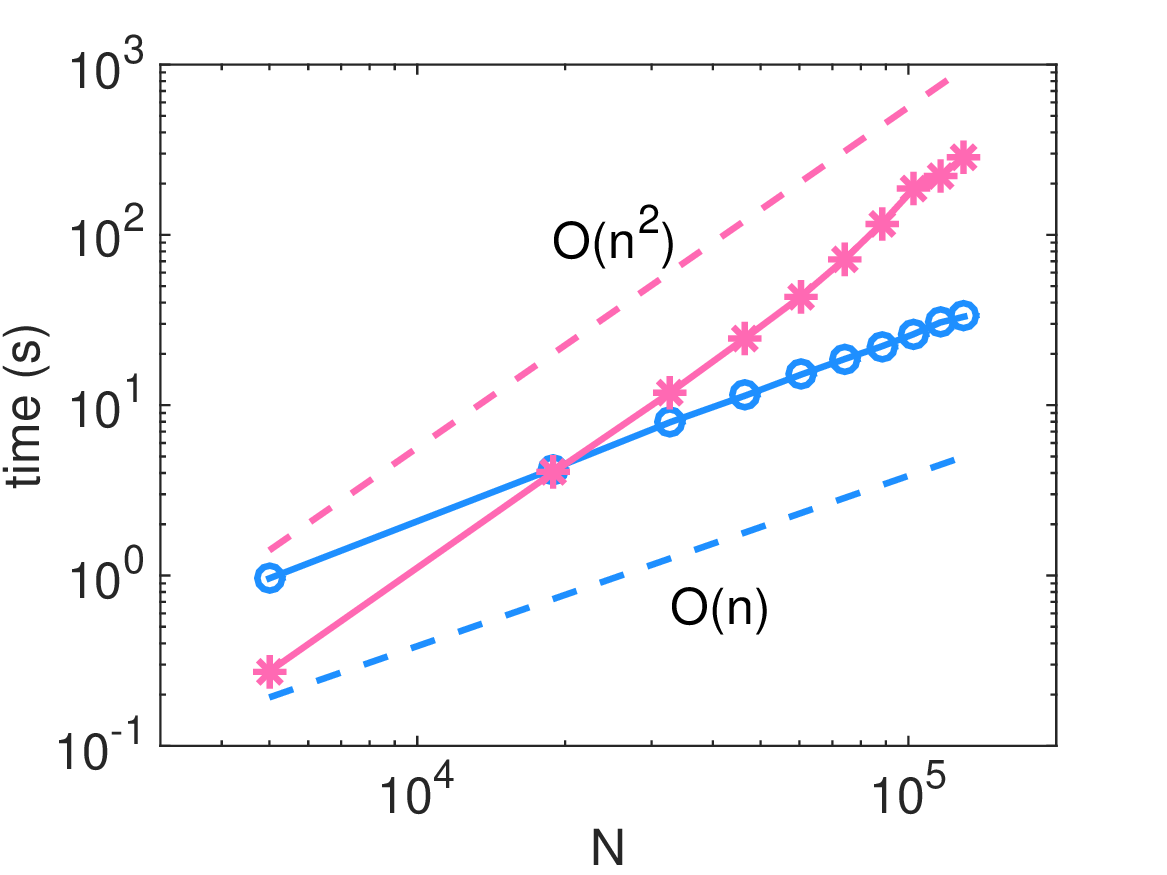}
        \caption{Total time (d = 40)}
    \end{subfigure}
    \caption{Timing (loglog scale) for IFGT and BBF on a synthetic
    dataset with dimension $d = 5$ and $40$.  We generated 10
    centers uniformly at random in a unit cube, and around each
    center we randomly generated data with standard deviation 0.1
    along each dimension.  The tolerance and the kernel parameter
    $h$ were set to $10^{-3}$, and 0.5, respectively.  All the plots
    share the same legends.  The top plots show the application time
    (matrix-vector product), and bottom plots show the total time
    (factorization and application).  The timing for BBF is linear
    for all dimensions, while the timing for the IFGT falls back to
    quadratic when $d$ increases.}
    \label{fig:ifgt}
\end{figure}

We verify the influence of dimension $d$ on the complexity of BBF and
the IFGT on synthetic datasets. We fixed the tolerance to be $10^{-3}$
throughout the experiments.  \autoref{fig:ifgt} shows the time versus
the number of points. We focus only on the trend of time instead of
the absolute value, because the IFGT was implemented in \verb|C++|
while {BBF} was in MATLAB. We see that the growth rate of IFGT is
linear when $d = 5$ but falls back to quadratic when $d = 40$; the
growth rate of BBF, however, remains linear.

\section{Conclusions and future work}

In this paper, we observed that for classification datasets whose
decision boundaries are complex, \emph{i.e.}, of small radius of
curvature, a small bandwidth parameter is needed for a high prediction
accuracy. In practical datasets, this complex decision boundary occurs
frequently when there exist a large variability in class sizes or
radii. These small bandwidths result in kernel matrices whose ranks
are not low and hence traditional low-rank methods are no longer
efficient. Moreover, for many machine-learning applications, low-rank
approximations of dense kernel matrices are inefficient. Hence, we
are interested in extending the domain of availability of low-rank
methods and retain computational efficiency.  Specifically, we
proposed a structured low-rank based algorithm that is of linear
memory cost and floating point operations, that remains accurate
even when the kernel bandwidth parameter is small, \emph{i.e.},
when the matrix rank is not low. We experimentally demonstrated that
the algorithm works in fact for a wide range of kernel parameters.
Our algorithm achieves comparable and often orders of magnitude
higher accuracy than other state-of-art kernel approximation methods,
with the same memory cost. It also produces errors with smaller
variance, thanks to the sophisticated randomized algorithm.  This is
in contrast with other randomized methods whose error fluctuates
much more significantly. Applying our algorithm to the kernel ridge
regression also demonstrates that our method competes favorably with
the state-of-art approximation methods.

There are a couple of future directions. One direction is on the
efficiency and performance of the downstream inference tasks. The
focus of this paper is on the approximation of the kernel matrix
itself. Although the experimental results have demonstrated good
performances in real-world regression tasks, we could further improve
the downstream tasks by relaxing the orthogonality constrains of the
$U_i$ matrices. That is, we can generate $U_i$ from interpolative
decomposition~\cite{Cheng2005}, which allows us to share the same
advantages as algorithms using Nys and RKS. Another direction is on
the evaluation metric for the kernel matrix approximation. This paper
used the conventional Frobenius norm to measure the approximation
performance. Zhang et al.~\cite{Zhang2018} proposed a new metric that
better measures the downstream performance. The new metric suggests
that to achieve a good generalization performance, it is important
to have a high-rank approximation. This suggestion aligns well with
the design of BBF and further emphasizes its advantage. Evaluating
BBF under the new metric will be explored in the future. Last but
not least, the BBF construction did not consider the regularization
parameter used in many learning algorithms. We believe that the
regularization parameter could facilitate the low-rank compression
of the kernel matrix in our BBF, while the strategy requires further
exploration.

\bibliographystyle{siam}
\bibliography{bbf}

\end{document}

%% file: fig-M.tex
\begin{tikzpicture}[scale = 0.12,baseline=-0.5ex]
\tikzset{every left delimiter/.style={xshift=-1ex},
    every right delimiter/.style={xshift=1ex},anchor=base}

\def\bdgap{0.4}
\def\igap{0.3}


\definecolor{col1}{RGB}{76,92,108}
\definecolor{col2}{RGB}{79,129,164}
\definecolor{col3}{RGB}{76,179,222}
\definecolor{col4}{RGB}{176,224,230}

\definecolor{row1}{RGB}{76,92,108}
\definecolor{row2}{RGB}{79,129,164}
\definecolor{row3}{RGB}{76,179,222}
\definecolor{row4}{RGB}{176,224,230}


\definecolor{col11}{RGB}{128,0,0}
\definecolor{col21}{RGB}{244,164,96}
\definecolor{col22}{RGB}{160,82,45}
\definecolor{col31}{RGB}{255,222,173}
\definecolor{col32}{RGB}{219,112,147}
\definecolor{col33}{RGB}{139,69,19}
\definecolor{col41}{RGB}{255,250,205}
\definecolor{col42}{RGB}{238,232,170}
\definecolor{col43}{RGB}{255,215,0}
\definecolor{col44}{RGB}{184,134,11}

\draw[fill=col4]  (0+\bdgap,0+\bdgap) rectangle 
    (1+\bdgap,6-\bdgap);
\draw[fill=col41] (1+\bdgap+\igap,5-\bdgap) rectangle
    (2+\bdgap+\igap,6-\bdgap);
\draw[fill=row1]  (2+\bdgap+2*\igap,5-\bdgap) rectangle
    (6-\bdgap,6-\bdgap);
\draw[fill=col4]  (0+6*1+\bdgap,0+\bdgap) rectangle
    (1+6*1+\bdgap,6-\bdgap);
\draw[fill=col42] (1+6*1+\bdgap+\igap,5-\bdgap) rectangle 
    (2+6*1+\bdgap+\igap,6-\bdgap);
\draw[fill=row2]  (2+6*1+\bdgap+2*\igap,5-\bdgap) rectangle
    (6+6*1-\bdgap,6-\bdgap);
\draw[fill=col4]  (0+6*2+\bdgap,0+\bdgap) rectangle
    (1+6*2+\bdgap,6-\bdgap);
\draw[fill=col43] (1+6*2+\bdgap+\igap,5-\bdgap) rectangle
    (2+6*2+\bdgap+\igap,6-\bdgap);
\draw[fill=row3]  (2+6*2+\bdgap+2*\igap,5-\bdgap) rectangle
    (6+6*2-\bdgap,6-\bdgap);
\draw[fill=col4]  (0+6*3+\bdgap,0+\bdgap) rectangle
    (1+6*3+\bdgap,6-\bdgap);
\draw[fill=col44] (1+6*3+\bdgap+\igap,5-\bdgap) rectangle    
    (2+6*3+\bdgap+\igap,6-\bdgap);
\draw[fill=row4]  (2+6*3+\bdgap+2*\igap,5-\bdgap) rectangle
    (6+6*3-\bdgap,6-\bdgap);

\draw[fill=col3]  (0+\bdgap,0+6*1+\bdgap) rectangle 
    (1+\bdgap,6+6*1-\bdgap);
\draw[fill=col31] (1+\bdgap+\igap,5+6*1-\bdgap) rectangle
    (2+\bdgap+\igap,6+6*1-\bdgap);
\draw[fill=row1]  (2+\bdgap+2*\igap,5+6*1-\bdgap) rectangle
    (6-\bdgap,6+6*1-\bdgap);
\draw[fill=col3]  (0+6*1+\bdgap,0+6*1+\bdgap) rectangle
    (1+6*1+\bdgap,6+6*1-\bdgap);
\draw[fill=col32] (1+6*1+\bdgap+\igap,5+6*1-\bdgap) rectangle 
    (2+6*1+\bdgap+\igap,6+6*1-\bdgap);
\draw[fill=row2]  (2+6*1+\bdgap+2*\igap,5+6*1-\bdgap) rectangle
    (6+6*1-\bdgap,6+6*1-\bdgap);
\draw[fill=col3]  (0+6*2+\bdgap,0+6*1+\bdgap) rectangle
    (1+6*2+\bdgap,6+6*1-\bdgap);
\draw[fill=col33] (1+6*2+\bdgap+\igap,5+6*1-\bdgap) rectangle
    (2+6*2+\bdgap+\igap,6+6*1-\bdgap);
\draw[fill=row3]  (2+6*2+\bdgap+2*\igap,5+6*1-\bdgap) rectangle
    (6+6*2-\bdgap,6+6*1-\bdgap);
\draw[fill=col3]  (0+6*3+\bdgap,0+6*1+\bdgap) rectangle
    (1+6*3+\bdgap,6+6*1-\bdgap);
\draw[fill=col43] (1+6*3+\bdgap+\igap,5+6*1-\bdgap) rectangle    
    (2+6*3+\bdgap+\igap,6+6*1-\bdgap);
\draw[fill=row4]  (2+6*3+\bdgap+2*\igap,5+6*1-\bdgap) rectangle
    (6+6*3-\bdgap,6+6*1-\bdgap);

\draw[fill=col2]  (0+\bdgap,0+6*2+\bdgap) rectangle 
    (1+\bdgap,6+6*2-\bdgap);
\draw[fill=col21] (1+\bdgap+\igap,5+6*2-\bdgap) rectangle
    (2+\bdgap+\igap,6+6*2-\bdgap);
\draw[fill=row1]  (2+\bdgap+2*\igap,5+6*2-\bdgap) rectangle
    (6-\bdgap,6+6*2-\bdgap);
\draw[fill=col2]  (0+6*1+\bdgap,0+6*2+\bdgap) rectangle
    (1+6*1+\bdgap,6+6*2-\bdgap);
\draw[fill=col22] (1+6*1+\bdgap+\igap,5+6*2-\bdgap) rectangle 
    (2+6*1+\bdgap+\igap,6+6*2-\bdgap);
\draw[fill=row2]  (2+6*1+\bdgap+2*\igap,5+6*2-\bdgap) rectangle
    (6+6*1-\bdgap,6+6*2-\bdgap);
\draw[fill=col2]  (0+6*2+\bdgap,0+6*2+\bdgap) rectangle
    (1+6*2+\bdgap,6+6*2-\bdgap);
\draw[fill=col32] (1+6*2+\bdgap+\igap,5+6*2-\bdgap) rectangle
    (2+6*2+\bdgap+\igap,6+6*2-\bdgap);
\draw[fill=row3]  (2+6*2+\bdgap+2*\igap,5+6*2-\bdgap) rectangle
    (6+6*2-\bdgap,6+6*2-\bdgap);
\draw[fill=col2]  (0+6*3+\bdgap,0+6*2+\bdgap) rectangle
    (1+6*3+\bdgap,6+6*2-\bdgap);
\draw[fill=col42] (1+6*3+\bdgap+\igap,5+6*2-\bdgap) rectangle    
    (2+6*3+\bdgap+\igap,6+6*2-\bdgap);
\draw[fill=row4]  (2+6*3+\bdgap+2*\igap,5+6*2-\bdgap) rectangle
    (6+6*3-\bdgap,6+6*2-\bdgap);

\draw[fill=col1]  (0+\bdgap,0+6*3+\bdgap) rectangle 
    (1+\bdgap,6+6*3-\bdgap);
\draw[fill=col11] (1+\bdgap+\igap,5+6*3-\bdgap) rectangle
    (2+\bdgap+\igap,6+6*3-\bdgap);
\draw[fill=row1]  (2+\bdgap+2*\igap,5+6*3-\bdgap) rectangle
    (6-\bdgap,6+6*3-\bdgap);
\draw[fill=col1]  (0+6*1+\bdgap,0+6*3+\bdgap) rectangle
    (1+6*1+\bdgap,6+6*3-\bdgap);
\draw[fill=col21] (1+6*1+\bdgap+\igap,5+6*3-\bdgap) rectangle 
    (2+6*1+\bdgap+\igap,6+6*3-\bdgap);
\draw[fill=row2]  (2+6*1+\bdgap+2*\igap,5+6*3-\bdgap) rectangle
    (6+6*1-\bdgap,6+6*3-\bdgap);
\draw[fill=col1]  (0+6*2+\bdgap,0+6*3+\bdgap) rectangle
    (1+6*2+\bdgap,6+6*3-\bdgap);
\draw[fill=col31] (1+6*2+\bdgap+\igap,5+6*3-\bdgap) rectangle
    (2+6*2+\bdgap+\igap,6+6*3-\bdgap);
\draw[fill=row3]  (2+6*2+\bdgap+2*\igap,5+6*3-\bdgap) rectangle
    (6+6*2-\bdgap,6+6*3-\bdgap);
\draw[fill=col1]  (0+6*3+\bdgap,0+6*3+\bdgap) rectangle
    (1+6*3+\bdgap,6+6*3-\bdgap);
\draw[fill=col41] (1+6*3+\bdgap+\igap,5+6*3-\bdgap) rectangle    
    (2+6*3+\bdgap+\igap,6+6*3-\bdgap);
\draw[fill=row4]  (2+6*3+\bdgap+2*\igap,5+6*3-\bdgap) rectangle
    (6+6*3-\bdgap,6+6*3-\bdgap);

\draw[step=6] (0,0) grid (24,24);
\draw [thick] (0,0) rectangle (24,24);

\draw (12,24) node[above=3pt,rectangle,fill=white] {$M$};

\end{tikzpicture}

%% file: fig-U.tex
\begin{tikzpicture}[scale = 0.12,baseline=-0.5ex]
\tikzset{every left delimiter/.style={xshift=-1ex},
    every right delimiter/.style={xshift=1ex},anchor=base}

\definecolor{col1}{RGB}{76,92,108}
\definecolor{col2}{RGB}{79,129,164}
\definecolor{col3}{RGB}{76,179,222}
\definecolor{col4}{RGB}{176,224,230}

\draw [xstep = 1,ystep = 6,dashed] (0,0) grid (4,24);

\draw[fill = col4] (3,0) rectangle (4,6);
\draw[fill = col3] (2,0+6*1) rectangle (3,6+6*1);
\draw[fill = col2] (1,0+6*2) rectangle (2,6+6*2);
\draw[fill = col1] (0,0+6*3) rectangle (1,6+6*3);

\draw [thick] (0,0) rectangle (4,24);

\draw (2,24) node[above=3pt,rectangle,fill=white] {$\widetilde{U}$};

\end{tikzpicture}

%% file: fig-C.tex
\begin{tikzpicture}[scale = 0.12,baseline=-0.5ex]
\tikzset{every left delimiter/.style={xshift=-1ex},
    every right delimiter/.style={xshift=1ex},anchor=base}

\definecolor{col11}{RGB}{128,0,0}
\definecolor{col21}{RGB}{244,164,96}
\definecolor{col22}{RGB}{160,82,45}
\definecolor{col31}{RGB}{255,222,173}
\definecolor{col32}{RGB}{255,165,0}
\definecolor{col33}{RGB}{139,69,19}
\definecolor{col41}{RGB}{255,250,205}
\definecolor{col42}{RGB}{238,232,170}
\definecolor{col43}{RGB}{255,215,0}
\definecolor{col44}{RGB}{184,134,11}

\draw[fill=col41] (0,0+20) rectangle (1,1+20);
\draw[fill=col42] (1,0+20) rectangle (2,1+20);
\draw[fill=col43] (2,0+20) rectangle (3,1+20);
\draw[fill=col44] (3,0+20) rectangle (4,1+20);

\draw[fill=col31] (0,1+20) rectangle (1,2+20);
\draw[fill=col32] (1,1+20) rectangle (2,2+20);
\draw[fill=col33] (2,1+20) rectangle (3,2+20);
\draw[fill=col43] (3,1+20) rectangle (4,2+20);

\draw[fill=col21] (0,2+20) rectangle (1,3+20);
\draw[fill=col22] (1,2+20) rectangle (2,3+20);
\draw[fill=col32] (2,2+20) rectangle (3,3+20);
\draw[fill=col42] (3,2+20) rectangle (4,3+20);

\draw[fill=col11] (0,3+20) rectangle (1,4+20);
\draw[fill=col21] (1,3+20) rectangle (2,4+20);
\draw[fill=col31] (2,3+20) rectangle (3,4+20);
\draw[fill=col41] (3,3+20) rectangle (4,4+20);

\draw [thick] (0,20) rectangle (4,24);

\draw (2,24) node[above=3pt,rectangle,fill=white] {$\widetilde{C}$};

\end{tikzpicture}

%% file: fig-VT.tex
\begin{tikzpicture}[scale = 0.12,baseline=-0.5ex]
\tikzset{every left delimiter/.style={xshift=-1ex},
    every right delimiter/.style={xshift=1ex},anchor=base}

\definecolor{col1}{RGB}{76,92,108}
\definecolor{col2}{RGB}{79,129,164}
\definecolor{col3}{RGB}{76,179,222}
\definecolor{col4}{RGB}{176,224,230}


\draw [xstep = 6,ystep = 1,dashed] (0,20) grid (24,24);

\draw[fill=col1] (0,3+20) rectangle (6,4+20);
\draw[fill=col2] (0+6*1,2+20) rectangle (6+6*1,3+20);
\draw[fill=col3] (0+6*2,1+20) rectangle (6+6*2,2+20);
\draw[fill=col4] (0+6*3,0+20) rectangle (6+6*3,1+20);

\draw [thick] (0,20) rectangle (24,24);

\draw (12,24) node[above=3pt,rectangle,fill=white] {$\widetilde{U}^\top$};

\end{tikzpicture}